\documentclass[12pt]{article}
\usepackage{amssymb}
\usepackage{amsfonts}
\usepackage{amsmath}
\usepackage{amsthm}
\usepackage{graphicx}
\usepackage{multirow}
\usepackage{caption}
\usepackage{subfig}
\usepackage[numbers]{natbib}
\usepackage[boxruled]{algorithm2e}

\usepackage[right=2.1cm,left=2.1cm, top=2.5cm, bottom=2.5cm]{geometry}
\usepackage{pstricks,pst-node,pst-plot}

\newcommand{\argmin}{\operatornamewithlimits{argmin}}
\newcommand{\argmax}{\operatornamewithlimits{argmax}}
\newcommand{\E}{\ensuremath{\mathrm{E}}}
\newcommand{\Var}{\ensuremath{\mathrm{Var}}}
\newcommand{\tr}[1]{\ensuremath{\mathrm{tr}\left(#1\right)}}
\newcommand{\df}{\ensuremath{\;\mathrm{d}}}

\numberwithin{equation}{section}
\numberwithin{figure}{section}
\numberwithin{table}{section}

\title{Bayesian Probabilistic Matrix Factorization: A User Frequency Analysis}
\author{Cody Severinski \\
Department of Statistical Sciences \\
University of Toronto \\
cody@utstat.utoronto.ca \\
\and Ruslan Salakhutdinov \\
Department of Statistical Sciences \\
\& Department of Computer Science \\
University of Toronto \\
rsalakhu@cs.toronto.edu}
\date{\today}

\begin{document}

\maketitle

\begin{abstract}
Matrix factorization (MF) has become a common approach to collaborative filtering, due to ease of implementation and scalability to large data sets.  Two existing drawbacks of the basic model is that it does not incorporate side information on either users or items, and assumes a common variance for all users.  We extend the work of constrained probabilistic matrix factorization by deriving the Gibbs updates for the side feature vectors for items \cite{ruslan:nips08}.  We show that this Bayesian treatment to the constrained PMF model outperforms simple MAP estimation.  We also consider extensions to heteroskedastic precision introduced in the literature \cite{laks2011rbmf}.  We show that this tends result in overfitting for deterministic approximation algorithms (ex: Variational inference) when the observed entries in the user / item matrix are distributed in an non-uniform manner.  In light of this, we propose a truncated precision model.  Our experimental results suggest that this model tends to delay overfitting.
\end{abstract}

\section{Introduction}
Matrix factorization (MF) techniques are commonly applied to model sparse data matrices.  The low-rank assumption of MF methods decomposes the data matrix as the product of two lower rank matrices.  By construction, each entry in the data matrix is an inner product of a vector from each of the two low rank matrices.  This has practical interpretations in recommender systems where the rows and columns  correspond to sets of objects, eg: Netflix (users and movies), Facebook (users and users), and genetics (genes and diseases).

Classical approaches (ex: SVD) are known to be inadequate, leading to the development of probabilistic approaches \cite{ruslan:nips08,ruslan:icml08}.  Here, the rating is assumed to be Gaussian conditional on vector latent features associated with the rows and columns.  Basic MAP estimation has been shown to be adequate, but prone to overfitting.  A fully Bayesian treatment places conjugate Gaussian-Wishart priors on the latent features and performs inference through Gibbs sampling \cite{ruslan:icml08}.

An existing limitation of most extensions is the assumption of homoskedastic variance for the prediction.  There have been proposals in the literature for modelling heteroskedastic variance using either multiplicative factors on the global precision or Student-$t$ priors on the features \cite{laks2011rbmf}.  Experimental results using variational inference suggested that the the inclusion of user / item rescaling factors on the precision improved model performance, but that Student-$t$ priors for the latent variables did not.  No direct comparison to Monte Carlo methods was given.

Similar users are known to rate similar sets of items, and also to rate these sets of items in a similar fashion.  This effect is known as correlational influence \cite{mjthesis}.  One approach to model correlational influence in matrix factorization models is to shift the user feature by additional features related to the items.  These additional features have been termed side features, and the model has been termed constrained probabilistic matrix factorization, or constrained PMF \cite{ruslan:nips08}.  We consider the Bayesian extension of this constrained model in this paper.  To our knowledge, this is the first time this extension has been considered.

In this paper, we compare the performance of these extensions using both a Gibbs sampler and a variational mean field approximation for inference.  We note how these various model extensions and inference algorithms perform for users of different frequency in the data.  Specifically, we

\begin{itemize}
\item Provide a comparison on the overall performance of Gibbs sampling and variational inference;
\item Highlight the tendency for variational inference to overfit, or get stuck in non-optimal modes, and indicate how this arises from the algorithm's objective of optimizing the variational lower bound;
\item Discuss the performance of these extensions with respect to the frequency of a user (i.e.: the number of ratings provided by the user);
\item Highlight the importance of side features.
\end{itemize}

\section{Probabilistic Matrix Factorization}
We review probabilistic matrix factorization in the context of users rating items, though it generalizes easily to other sets of objects.  Let $R \in \mathbb{R}^{N \times M}$ denote a rating matrix between $N$ users and $M$ items, where the $(i,j)$ entry is the rating given by user $i$ to item $j$.  The matrix $R$ presents two complications to modelling: sparsity and imbalance.  The sparsity is a consequence of most users rating a small subset of items, while the imbalance follows from difference in the popularity of items.

The goal of matrix factorization is to find a low-rank approximation to $R$ as $U^\top V$, where $U$ is a real-valued $d \times N$ matrix, and $V$ is a real-valued $d \times M$ matrix.  Each column of $U$ is a latent feature for the users, similarly for the matrix $V$.  Probabilistic Matrix Factorization  models this low rank approximation as

\begin{equation}
\begin{aligned}
\label{eq:pmf}
(r_{i,j} \;|\; U_i, V_j, I_{i,j} = 1) \sim& \mathcal{N}(r_{i,j} \;|\; U_i^\top V_j, \tau)\\
(U_i \;|\; \mu_U, \Lambda_U) \sim& \mathcal{N}(U_i \;|\; \mu_U, \Lambda_U)\\
(V_j \;|\; \mu_V, \Lambda_V) \sim& \mathcal{N}(V_j \;|\; \mu_V, \Lambda_V)
\end{aligned}
\end{equation}

Where $\mathcal{N}(x \;|\; \mu, \tau)$ denote the Gaussian distribution for $x$ with mean $\mu$ and precision $\tau$, and $I_{i,j} \in \{0,1\}$ is the indicator that user $i$ provided a rating for item $j$.  In our parametrization, $\mathcal{N}(x \;|\; \mu,\Lambda)$ is the normal distribution with mean vector $\mu$ and precision matrix $\Lambda$.

In practice, it is important to model the bias of each user and each item.  Let $\gamma_i$ denote the bias for user $i$, and $\eta_j$ the bias for item $j$.  The mean of the predicted rating in Equation~\eqref{eq:pmf} becomes

\begin{equation}
\begin{aligned}
\label{eq:rhat}
\mathrm{E}[r_{i,j} \;|\; \gamma_i, \eta_j, U_i, V_j] =& \gamma_i + \eta_j + U_i^\top V_j
\end{aligned}
\end{equation}

Letting $\hat{r}_{i,j} = \gamma_i + \eta_j + U_i^\top V_j$, the likelihood of the data $R$ given the parameters $\Theta_R = (\gamma_{1:N}, \eta_{1:M}, U_{1:N}, V_{1:M})$ is

\begin{equation}
\begin{aligned}
\label{eq:pmf:likelihood}
p(R \;|\; \Theta_R, \tau) =& \prod_{i=1}^{N} \prod_{j=1}^{M}\left[ \mathcal{N}(r_{i,j} \;|\; \hat{r}_{i,j}, \tau ) \right]^{I_{i,j}}
\end{aligned}
\end{equation}

\subsection{Constrained Probabilistic Matrix Factorization}
\label{sec:cbpmf}

\cite{ruslan:nips08} considers a constrained version of PMF with an offset on the user feature vectors depending on the movies the user watched.  For each movie $k \in \{1,\ldots, M\}$, introduce a $d$ dimensional latent vector $W_k$, and modify the expectation of the rating in equation \eqref{eq:pmf} as

\begin{equation}
\begin{aligned}
\label{eq:cpmf_mean}
\mathrm{E}[r_{i,j} | U_i, V_j, W_{1:M}] =& \left(\delta_{U}U_i + \delta_{W}\frac{\sum_{k=1}^{M}I_{i,k}W_k}{\sum_{k'=1}^{M}I_{i,k'}}\right)^\top V_j \\
  =& \left(\delta_{U}U_i + \frac{\delta_{W}}{n_i}\sum_{k=1}^{M}I_{i,k}W_k\right)^\top V_j \\
  =& S_i^\top V_j
\end{aligned}
\end{equation}

Where we have defined $n_i = \sum_{k'=1}^{M}I_{i,k'}$ the number of items observed by user $i$, and $S_i = \delta_{U}U_i + \delta_{W}\sum_{k=1}^{M}I_{i,k}W_k / n_i$ the combined user / side feature contribution by user $i$..  This ensures the user offset is on the same scale for all users, independent of the number of items observed.  Equation~\eqref{eq:cpmf_mean} explicitly includes delta functions $\delta_U, \delta_W \in \{0,1\}$ to emphasize this as an extension to the vanilla PMF model.  When $\delta_{W} = 0$, the model includes only user-specific latent features in the user offset, reducing to Equation~\eqref{eq:pmf}.  If $\delta_{U} = 0$, then the model does not consider user features.

To highlight the contribution of this extension, note that users with few or no ratings will have a posterior for $U_i$ that is close to the prior.  Hence, $U_i^\top V_j$ will be close to the overall average, with the inner product between the $W_k$ and $V_j$ shifting the rating based on the ratings of other users who also rated the same item.

To make this abstract, constrained PMF introduces a new set of features $W_k, k = 1, \ldots, M$ for each column of the rating matrix $R$.  The prediction for the value of entry $(i,u)$ in this matrix is the inner product of the feature associated with the row $U_i$ and the feature associated with the column $V_j$, plus the inner produce of the column feature $V_j$ and the average of these additional feature $W_k$ that are associated with observed entries in this row.  This abstraction highlights the symmetric nature of these latent side features.  The most reliable inference on these additional side features $W_k$ will be obtained when they are associated with the dimension of the matrix with less sparsity (row-wise or column-wise).

A similar constraint can be placed on the items by transposing the user-item matrix.  The model without side features is invariant to this transposition.  For the model with side features, the transposition modifies the expectation of the rating to be 

\begin{equation}
\begin{aligned}
\label{eq:cpmf:alt:mean}
\mathrm{E}[r_{i,j} \;|\; U_i, V_j, W_{1:M}] =& U_i^\top \left(\delta_{W}\frac{\sum_{\ell=1}^{N}I_{\ell,j}W_\ell}{\sum_{\ell'=1}^{N}I_{\ell',j}} +   V_j\right) \\
  =& U_i^\top \left(\frac{\delta_{W}}{m_j}\sum_{\ell=1}^{N}W_\ell + V_j\right) \\
\end{aligned}
\end{equation}

Where we have defined $m_j = \sum_{\ell=1}^{N}I_{\ell,j}$ to number of observed ratings for item $j$.

When the side features $W_k$ offset the user features $U_i$, the contribution from each user to the inner product is affected.  Similarly, when the side features $W_k$ offset the item features $V_j$ as in Equation~\eqref{eq:cpmf:alt:mean}, the contribution from each item to the inner product is affected.  Since both the user and item features $U_i, V_j$ can be close to the prior from lack of data in the respective rows / columns of the matrix, overall performance will be improved when the side features $W_k$ are introduced to shift either the user or item contribution away from the prior.

\subsection{Inference}

Learning for PMF is performed by maximizing the log-posterior over the parameters $(U,V,\gamma,\eta)$ given the data

\begin{equation}
\begin{aligned}
\log p(U,V,\gamma,\eta \;|\; R) =& \log p(R \;|\; U,V, \gamma, \eta) + \log p(U) + \log p(\gamma) + \log p(V) + \log p(\eta) + \log p(W)
\end{aligned}
\end{equation}

Maximizing this posterior with respect to the parameters is equivalent to minimizing the sum of squared error function with quadratic regularization terms on the parameters:

\begin{equation}
\begin{aligned}
\label{eq:energy}
E =& \frac{\tau}{2}\sum_{i=1}^{N}\sum_{j=1}^{M}I_{i,j}(r_{i,j} - \hat{r}_{i,j})^2 	
    + \frac{\lambda_U}{2}\sum_{i=1}^{N}\gamma_i^2
    + \frac{\lambda_V}{2}\sum_{j=1}^{M}\eta_j^2 \\
   & + \frac{\lambda_U}{2}\sum_{i=1}^{N}\|U_i\|_2^2 
    + \frac{\lambda_V}{2}\sum_{j=1}^{M}\|V_j\|_2^2
    + \frac{\lambda_K}{2}\sum_{k=1}^{M}\|W_k\|_2^2,
\end{aligned}
\end{equation}

In practice, we find learning is improved by learning the parameters in stages.  In the first stage, we fix the features $U,V$ to zero and learn the biases $\gamma, \eta$ by minimizing

\begin{equation}
\begin{aligned}
\label{eq:energy:bias}
E =& \frac{\tau}{2}\sum_{i=1}^{N}\sum_{j=1}^{M}I_{i,j}(r_{i,j} - (\gamma_i + \eta_j))^2 	
    + \frac{\lambda_U}{2}\sum_{i=1}^{N}\gamma_i^2
    + \frac{\lambda_V}{2}\sum_{j=1}^{M}\eta_j^2
\end{aligned}
\end{equation}

Once estimates for $\gamma_{1:N}, \eta_{1:M}$ are obtained, we then learn $U,V$ by minimizing

\begin{equation}
\begin{aligned}
\label{eq:energy:feature}
E =& \frac{\tau}{2}\sum_{i=1}^{N}\sum_{j=1}^{M}I_{i,j}(r_{i,j} - \hat{r}_{i,j})^2 	
   + \frac{\lambda_U}{2}\sum_{i=1}^{N}\|U_i\|_2^2 
    + \frac{\lambda_V}{2}\sum_{j=1}^{M}\|V_j\|_2^2
    + \frac{\lambda_K}{2}\sum_{k=1}^{M}\|W_k\|_2^2,
\end{aligned}
\end{equation}

Batch gradient descent with momentum was used to find an approximate MAP estimate for these parameters.  To reduce the number of tuning parameters, a common penalty and learning rate was assumed for all features.  The final MAP estimate was selected by stopping gradient descent when error on the validation set increased.

\section{Bayesian (C)PMF}

The likelihood for the ratings remains as in Equation~\eqref{eq:pmf}.  Conjugate Gaussian prior distributions are placed over the features and the biases:

\begin{equation}
\begin{aligned}
\label{eq:prior:feature_bias}
(U_i \;|\; \mu_U, \Lambda_U) 
  \sim& \mathcal{N}(U_i  \;|\; \mu_U, \Lambda_U) \\
(V_j \;|\; \mu_V, \Lambda_V) 
  \sim& \mathcal{N}(V_j \;|\; \mu_V, \Lambda_V) \\
(W_k \;|\; \mu_W, \Lambda_W) 
  \sim& \mathcal{N}(W_k \;|\; \mu_W, \Lambda_W) \\
(\gamma_i \;|\; \mu_\gamma, \lambda_\gamma) \sim& \mathcal{N}(\gamma_i \;|\; \mu_\gamma, \lambda_\gamma) \\
(\eta_j \;|\; \mu_\eta, \lambda_\eta) \sim& \mathcal{N}(\eta_j \;|\; \mu_\eta, \lambda_\eta)
\end{aligned}
\end{equation}

Following the literature, Gaussian-Wishart priors are placed on the feature hyper-parameters $\{\mu_U, \Lambda_U\}, \{\mu_V, \Lambda_V\}, \{\mu_W, \lambda_W\}$

\begin{equation}
\begin{aligned}
\label{eq:prior:hyper}
(\mu_U, \Lambda_U) 
  \sim& \mathcal{N}(\mu_U \;|\; \mu_0, \beta_0\Lambda_U) 
    \cdot \mathcal{W}(\Lambda_U \;|\; W_0, \nu_0) \\
(\mu_V, \Lambda_V) 
  \sim& \mathcal{N}(\mu_V \;|\; \mu_0, \beta_0\Lambda_V)
    \cdot \mathcal{W}(\Lambda_V \;|\; W_0, \nu_0)\\
(\mu_W, \Lambda_W) 
  \sim& \mathcal{N}(\mu_W \;|\; \mu_0, \beta_0\Lambda_W) 
    \cdot \mathcal{W}(\Lambda_W \;|\; W_0, \nu_0) \\
\end{aligned}
\end{equation}

Where $\mathcal{W}(\Lambda \;|\; W_0, \nu_0)$ is the Wishart distribution for a random variable $\Lambda$ with $\nu_0$ degrees of freedom and scale matrix $W_0$.

This is the model illustrated in Figure~\ref{fig:gm:bpmf} without the dashed lines.

\section{Scaled BPMF}

\cite{laks2011rbmf} has considered heteroskedastic extensions to BPMF.  One extension investigated incorporated rescaling factors $\alpha_i$ and $\beta_j$ specific to each row $i$ and column $j$ on the distribution of the rating.  This modified the likelihood of the data in Equation~\eqref{eq:pmf} to

\begin{equation}
\begin{aligned}
\label{eq:pmf_scaled}
(r_{i,j} \;|\; U_i, V_j) \sim& \left[\mathcal{N}(r_{i,j} \;|\; U_i^\top V_j, \alpha_i\beta_j\tau)\right]^{I_{i,j}}\\
\end{aligned}
\end{equation}

The Bayesian extension placed Gamma prior distributions for these precision factors:
\begin{equation}
\begin{aligned}
\label{eq:sbpmf}
(\alpha_i \;|\; a_U, b_U) \sim& \mathcal{G}(\alpha_i \;|\; a_U, b_U) \\
(\beta_j \;|\; a_V, b_V) \sim& \mathcal{G}(\beta_j \;|\; a_V, b_V) \\
(\tau \;|\; a_\tau, b_\tau) \sim& \mathcal{G}(\tau \;|\; a_\tau, b_\tau)
 \end{aligned}
\end{equation}

Where $\mathcal{G}(x \;|\; a, b)$ is the gamma distribution for $x$ parametrized with rate $b$, having density proportional to $x^{a-1} e^{-bx}$.

A second extension considered placed Student-$t$ priors on the feature vectors.  The distribution on the features $U_i, V_j$ were redefined as Gaussian scale mixtures

\begin{equation}
\begin{aligned}
(U_i, \alpha_i) \sim& \mathcal{N}(U_i \;|\; \mu_U, \alpha_i\Lambda_U) \mathcal{G}(\alpha_i) \\
(V_j, \beta_j) \sim& \mathcal{N}(V_j \;|\; \mu_V, \beta_j\Lambda_V) \mathcal{G}(\beta_j)
\end{aligned}
\end{equation}

Analytically integrating out the $\alpha_i, \beta_j$ produces Student-$t$ distributions for $U_i, V_j$.
\begin{equation}
\begin{aligned}
\label{eq:pmf_t}
(r_{i,j} \;|\; U_i, V_j) \sim& \left[\mathcal{N}(r_{i,j} \;|\; U_i^\top V_j, \tau)\right]^{I_{i,j}}\\
(U_i \;|\; \mu_U, \Lambda_U) \sim& \int \mathcal{N}(U_i \;|\; \mu_U, \alpha_i\Lambda_U) \mathcal{G}(\alpha_i) \df{\alpha_i}\\
(V_j \;|\; \mu_V, \Lambda_V) \sim& \int \mathcal{N}(V_j \;|\; \mu_V, \beta_j\Lambda_V) \mathcal{G}(\beta_j) \df{\beta_j}
\end{aligned}
\end{equation}

Experimental results \cite{laks2011rbmf} suggested that the the inclusion of user / item rescaling factors on the precision improved model performance, but that Student-$t$ priors for the latent variables did not.

\begin{figure}[htb]
\begin{center}
\caption{Bayesian Constrained Probabilistic Matrix Factorization with Gaussian-Wishart Priors over the latent user, item, and side feature vectors.  The user precision factors $\alpha_i$ and item precision factors $\beta_j$ allows for non-constant variance in the observed preference $r_{i,j}$.  The extension to scaled precision is obtained by including the dashed lines.}
\label{fig:gm:bpmf}
\includegraphics[]{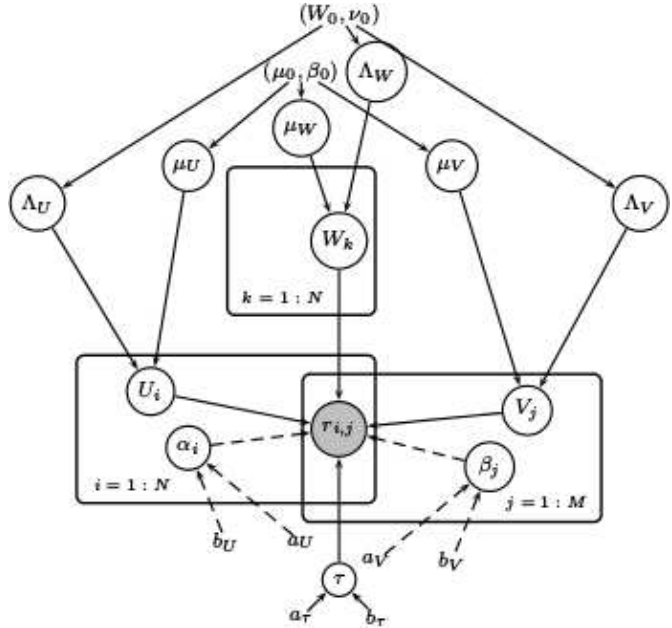}
\end{center}
\end{figure}

The model with Gamma prior distributions for the user and item precisions is illustrated in Figure~\ref{fig:gm:bpmf} by including the dashed lines.

\section{Truncated Precisions}
\label{sec:truncated}

The choice of the Gamma distribution for the precision factors is computationally convenient, but unrealistic in practice.  It is limited in that a distribution taking values arbitrarily small or large does not reflect the prior knowledge that the actual rating system is bounded.

In our experimental results, we noticed this posed problems when direct minimization of an error function was the objective.  A deterministic algorithm can arbitrarily shrink some user / item precisons to zero, while driving others arbitrarily large.  The result is a decrease in the overall error by optimizing for a subset of the user-item matrix.  This is analogous to a similar behaviour with finite mixture models.  In these models, variational inference may shrink some mixing weights to zeros.

A simple approach to this would be to bound the precisions to values suggested by the actual data.  Such truncation is commonly applied to different distributions, such as the Normal distribution.  They appear often enough in practice that software has been developed in \texttt{R} for it \cite{nadarajah:trunc}.  In general, an unbounded distribution with density $g_X(x)$ is truncated to $(\ell,u)$ by defining $f_X(x) \propto g_X(x) \mathbf{1}\{\ell < x < u\}$.  For the case of our Gamma$(\alpha,\beta)$ precisions, the density becomes

\begin{equation}
\begin{aligned}
f_X(x) \propto& x^{\alpha-1} e^{-\beta x} \mathbf{1}\{\ell < x < u\}
\end{aligned}
\end{equation}

\section{Inference}

The predictive distribution for the ratings $r_{i,j}$ is obtained by integrating out the features, the hyper-parameters, and the precisions.  This integral is computationally intractable, requiring the use of approximate methods.  We can either perform approximate inference on the true posterior, or approximate the posterior by a simpler distribution and perform exact inference.  The first choice (approximating the truth) is the realm of Monte Carlo methods.  The second (exact inference on an approximation) is the realm of Variational methods.

\subsection{Gibbs Sampling}
The choice of conjugate prior distributions yields tractable posterior distributions that are easy to sample from.  In particular, the user, item and side features are multivariate Gaussian distributions.  They are similar to the form derived in \cite{ruslan:icml08}, with two exceptions.  Our derivation includes the scaling from the precision factors $\alpha_i, \beta_j$, and also the shifting effect from the side features $W_k$.  The posterior for the user features is

\begin{equation}
\begin{aligned}
U_i \sim& \mathcal{N}(U_i \;|\; \mu_{U_i}, \Lambda_{U_i}, R, \mu_U, \Lambda_U), \\
  \text{where }\mu_{U_i} =& \Lambda_{U_i}^{-1}\bigg[\Lambda_U\mu_U + \delta_{U}\tau\alpha_i \sum_{j=1}^{M}I_{i,j}\beta_j V_j\left(r_{i,j} - \frac{\delta_{W}}{n_i}V_j^\top \left(\sum_{k=1}^{M}I_{i,k}W_k\right)\right)\bigg] \\
\Lambda_{U_i} =& \Lambda_U + \delta_{U}\tau\alpha_i\sum_{j=1}^{M}I_{i,j}\beta_jV_j V_j^\top
\end{aligned}
\end{equation}

The full derivation of these is given in Appendices~\ref{appendix:varDist}, \ref{appendix:gibbs}, and summarized in Tables~\ref{table:dist:summary1}--\ref{table:dist:summary2} in Appendix~\ref{appendix:summary}.

The conditional distribution for the feature hyper-parameters is unaffected by the addition of the precision factors and the side features.  The symmetry in the probabilistic formulation of the model means the form is identical.  The posterior for the user hyper-parameters is

\begin{equation}
\begin{aligned}
(\mu_U,\Lambda_U) \sim& \mathcal{N}(\mu_U \;|\; \tilde{\mu}_U, \tilde{\Lambda}_U) \cdot \mathcal{W}(\Lambda_U \;|\; \tilde{\nu}_U, \tilde{W}_U) \\
\text{where } \tilde{\mu}_U =& \frac{N\overline{U} + \beta_0\mu_0}{N + \beta_0},\\
\overline{U} =& \frac{1}{N}\sum_{i=1}^{N}U_i , \\
\tilde{\Lambda}_U =& (N + \beta_0)\Lambda_U, \\
\tilde{\nu}_U =& N + \nu_0 \\
\tilde{W}_U^{-1} =& W_0^{-1} + \frac{N\beta_0}{N + \beta_0}(\overline{U} - \mu_0)(\overline{U} - \mu_0)^\top 
  + \sum_{i=1}^{N}(U_i - \overline{U})(U_i - \overline{U})^\top
\end{aligned}
\end{equation}

The posterior for the hyper-parameters of the item and side features is analogous.

The precisions have the same Gamma conditional posterior as in \cite{laks2011rbmf},

\begin{equation}
\begin{aligned}
\alpha_i \sim& \mathcal{G}(\alpha_i \;|\; \tilde{a}_{U_i}, \tilde{b}_{U_i}) \\
\text{where } \tilde{a}_\tau =& a_\tau + \frac{1}{2}\sum_{i=1}^{N}\sum_{j=1}^{M} I_{i,j} \\
\tilde{b}_\tau =& b_\tau + \frac{\tau}{2}\sum_{i=1}^{N}\sum_{j=1}^{M} I_{i,j} \alpha_i\beta_j(r_{i,j} - \hat{r}_{i,j})^2
\end{aligned}
\end{equation}

In the case of truncated precisions, Gibbs sampling can still be achieved through the introduction of a latent variable \cite{damien2001sampling}.

Samples of the features $(U_i, V_j, W_k)$, the precisions $(\alpha_i, \beta_j, \tau)$, the hyper-parameters $(\mu_U, \Lambda_U)$, $(\mu_V, \Lambda_V)$, $(\mu_K, \Lambda_K)$, and the biases $(\gamma_i, \eta_{j})$ are obtained by running a Markov Chain with stationary distribution equal to the true posterior distribution over the parameters.  The Gibbs algorithm is Algorithm~\ref{algo:gibbs} in the Appendix.

After iteration $t$, we have samples for the features $(U^{(t)}_i, V^{(t)}_j, W^{(t)}_k)$, the precisions $(\alpha^{(t)}_i, \beta^{(t)}_j, \tau^{(t)})$, the hyper-parameters $(\mu^{(t)}_U, \Lambda^{(t)}_U)$, $(\mu^{(t)}_V, \Lambda^{(t)}_V)$, $(\mu^{(t)}_K, \Lambda^{(t)}_K)$, and the biases $(\gamma^{(t)}_i, \eta^{(t)}_{j})$.  We estimate the ratings via:

\begin{equation}
\begin{aligned}
\label{eq:gibbs:predict}
\hat{r}_{i,j} =& \gamma^{(t)}_i + \eta^{(t)}_j + {U^{(t)}_i}^\top V^{(t)}_j
\end{aligned}
\end{equation}

The prediction after $T$ runs of the Gibbs sampler is the average:

\begin{equation}
\begin{aligned}
\label{eq:gibbs:predict:average}
\hat{r}^{(T)}_{i,j} =& \frac{1}{T}\sum_{t=1}^{T}\gamma^{(t)}_i + \eta^{(t)}_j + {U^{(t)}_i}^\top V^{(t)}_j
\end{aligned}
\end{equation}

\subsection{Variational Inference}

\label{sec:varinf:intro}

Rather than attempting to make inference on the true posterior distribution $p(\theta \;|\; R)$, Variational methods makes dependence assumptions on the parameters $\theta$, formalizes them as an unknown joint distribution $Q(\theta)$, and infers the distribution and choice of parameters on $Q(\theta)$ to make it a good approximation to $p(\theta \;|\; R)$.  There are two decisions to be made: the form of the approximation, and how to measure ``close''.

To measure the distance between the approximation $Q$ and the truth $p$, we use Kullback-Leibler (KL) divergence,
\begin{equation}
\begin{aligned}
\label{eq:kl}
\mathrm{KL}(p\,\|\,Q) =& \int Q(\theta) \log \frac{Q(\theta)}{p(\theta\,|\,R)} \df{\theta}
\end{aligned}
\end{equation}

The values of $\theta$ are selected to minimize the KL divergence,

\begin{equation}
\begin{aligned}
\label{eq:kl:min}
\hat\theta_\text{VI} =& \argmin_{\theta} \text{KL}(Q\,\|\,p)
\end{aligned}
\end{equation}

The KL divergence in equation \eqref{eq:kl} can be rewritten as

\begin{equation}
\begin{aligned}
\label{eq:klfree}
\text{KL}(Q\|p) 
  =& \int Q(\theta) \log \frac{Q(\theta)}{p(\theta\;|\;R)} \df{\theta} \\
  =& \int Q(\theta) \log \frac{Q(\theta)}{p(\theta,R)} \df{\theta} + \log p(R) \\
\end{aligned}
\end{equation}

Rearranging, we express the log probability of the data, $\log p(R)$ in terms of two quantities involving the variational distribution: the KL divergence and the entropy $\mathcal{H}(Q) = \E_{Q}[\log p(Q)]$.

\begin{equation}
\begin{aligned}
\label{eq:logprobKL}
\log p(X) 
  =& KL(Q\|p) -  \int Q(\theta) \log \frac{Q(\theta)}{p(\theta,R)} \df{\theta} \\
  =& KL(Q\|p) +  \int Q(\theta) \log \frac{p(\theta,R)}{Q(\theta)} \df{\theta} \\
  =& KL(Q\|p) +  E_{Q}[\log P(\theta,R)] - \mathcal{H}(Q)
\end{aligned}
\end{equation}

Since $\log p(R)$ is fixed,  minimizing the KL is equivalent to jointly maximizing the second and third terms, corresponding to the expected complete log likelihood and the entropy.  This expression is known as the variational lower bound.  We derive this in Appendix \ref{appendix:varLowerBound}.
To make the optimization tractable, a mean field approximation is frequently chosen, where the parameters $\theta$ are independent in the approximation $Q$.  This leads to Equation~\eqref{eq:kl} factorizing with respect to $\theta$,

\begin{equation}
\begin{aligned}
\label{eql:kl:factorize}
\mathrm{KL}(p\,\|\,Q) 
  =& \int Q(\theta) \log \frac{Q(\theta)}{p(\theta\,|\,R)} \df{\theta} \\
  =& \int \prod_{i=1}^{N}Q(\theta_i) \log \frac{Q(\theta_i)}{p(\theta\,|\,R)} \df{\theta_i} \\
  =& \prod_{i=1}^{N} \int Q(\theta_i) \log \frac{Q(\theta_i)}{p(\theta\,|\,R)} \df{\theta_i}
\end{aligned}
\end{equation}

Optimization can then be performed sequentially over parameters until convergence is achieved, as measure by the variational lower bound.  Optimizing Equation~\eqref{eq:logprobKL}, or the factorized equivalent in Equation~\eqref{eql:kl:factorize}, with respect to $\theta$ imposes assumptions on the distribution of $\theta$

Our approximation follows the standard structured mean field approximation in the literature \cite{laks2011rbmf},

\begin{equation}
\begin{aligned}
\label{eq:meanFieldApprox}
& Q(U_{1:n}, V_{1:m}, W_{1:m}, \alpha_{1:n}, \beta_{1:m}, \tau, \mu_U, \Lambda_U, \mu_V, \lambda_V, \mu_W, \Lambda_W | R) \\
  =& Q(\tau) \,\cdot\, \prod_{i=1}^{N} \left[Q(U_i)Q(\alpha_i)\right] \,\cdot\,
    \prod_{j=1}^{M} \left[ Q(V_j)Q(\beta_j) \right]\,\cdot\,
    \prod_{k=1}^{M} Q(W_k) \\
    & 
    \,\cdot\, Q(\mu_U, \Lambda_U)\,\cdot\,
    Q(\mu_V, \Lambda_V)\,\cdot\,
    Q(\mu_W, \Lambda_W)
\end{aligned}
\end{equation}

This approximation assumes pairwise independence between the user, item, and side features, while allowing for structure in the latent feature hyper-parameters.

Under this mean field approximation, it can be shown that $Q(\alpha_i), Q(\beta_j), Q(\tau)$ are Gamma distributions, possibly truncated depending on the model chosen for $p(\theta, R)$ (see Appendix).  Inference for the optimal parameters follows immediately as the  MAP estimate is still available in a closed form as,

\begin{equation}
\begin{aligned}
\argmax_{x} f_{X}(x) =& \max\{\min\{\alpha/\beta, u\}, \ell\}.
\end{aligned}
\end{equation}

This is just the unbounded MAP estimate, with barriers at the truncation endpoints.

The variational algorithm is described in pseudo-code in Algorithm~\ref{algo:varinf} in the Appendix.  The prediction after $T$ full updates of the parameters under the variational algorithm is the prediction given the current variational approximation:

\begin{equation}
\begin{aligned}
\label{eq:varinf:predict}
\hat{r}^{(T)}_{i,j} =& \gamma^{(T)}_i + \eta^{(T)}_j + {U^{(T)}_i}^\top V^{(T)}_j
\end{aligned}
\end{equation}

\section{Experimental Setup}

The data set for model evaluation is the MovieLens 1M data set.  This consists of $1,000,209$ ordinal ratings on the scale $\{1, \ldots, 5\}$ by $N = 6,040$ users on $M = 3,952$ items.  To make a direct comparison to previously reported variational results \cite{laks2011rbmf}, we removed any movies rated less than three times, and ensured that each user and movie appeared in the training set once.  The data was split into a $70\%$ training, $30\%$ testing set for evaluation.  We report root-mean-square error (RMSE) on the test set for the models considered.

A second data set used was the Epinions data set.  This consists of 664,824 ordinal ratings on a $\{1, \dots, 5\}$ scale by $N = 49,290$ users who rated $M = 139,738$ items.  We ensured that each user and each item appeared at least once in the training set.  No other conditions were imposed on the train / test split.  The data was split into a $70\%$ training, $30\%$ testing set for evaluation.  We report root-mean-square error (RMSE) on the test set for the models considered.  Table \ref{table:data} provides summary information on the two data sets considered.

Section~\ref{sec:cbpmf} abstracted the notion of constrained PMF as an additional set of latent features associated with either rows or columns.  The optimal choice is to associate the additional set of latent features with the dimension of the matrix with less sparsity.  In MovieLens, the average user rates 

For Gibbs, experimentation with different number of samples was used to determine a point at which convergence occurred.  Burn-in was ignored.  An exploratory analysis of traceplots of the feature vectors suggested quick mixing, and the initial decline in the overall test error was rapid.  Combined, both of these suggest that allowing for burn-in would have minimal improvement.  Convergence of the variational algorithm was assessed using the variational lower bound where possible.  For further discussion, see Section \ref{sec:varinf}.

Unless otherwise noted, all simulations that followed used the following choices for the tuning parameters.  For the Gaussian-Wishart priors on the feature vectors, $(\mu_U, \Lambda_U, \beta_0, \nu_0) = (0_{d\times 1}, \mathbf{I}_{d}, 1, d+1)$.  The mean value was chosen to reflect that the features are mean zero after accounting for the biases, while the values for the scale matrix and degrees of freedom were selected to give a vague prior that was still proper.  All precisions were given shape and scale parameters of 2.  This yields a mean of 1, variance of 1/2, and contains approximately 95\% of the probability mass at values of five or less.

\begin{table}[ht]
\begin{center}
\caption{Summary data on the MovieLens 1M and the Epinions data set.}
\label{table:data}
\begin{tabular}{rr|rr}
\hline
 & MovieLens & Epinions \\
\hline
\multicolumn{2}{r}{Number of Ratings} & 1,000,209 & 664,824 \\
\multicolumn{2}{r}{Number of Users} & 6,040 & 49,290 \\
\multicolumn{2}{r}{Number of Items} & 3,952 & 139,738 \\
\multirow{6}{*}{Ratings per Item} & Min & 0 & 0\\
& 25$^{\mathrm{th}}$ & 23 & 1 \\
& 50$^{\mathrm{th}}$ & 104 & 1 \\
& Mean & 166 & 3 \\
& 75$^{\mathrm{th}}$ & 323 & 2 \\
& Max & 3,428 & 1408 \\
\multirow{6}{*}{Ratings per User}  & Min & 20 & 0 \\
& 25$^{\mathrm{th}}$ & 44 & 1 \\
& 50$^{\mathrm{th}}$ & 96 & 3 \\
& Mean & 253 & 9 \\
& 75$^{\mathrm{th}}$ & 208 & 9 \\
& Max & 2,314 & 724 \\
\multicolumn{2}{r}{Sparsity} & 4.19\% & 0.01\%\\
\hline
\end{tabular}
\end{center}
\end{table}

\section{Results}

Table~\ref{table:ml1mError} summarizes the test RMSE values obtained on the MoviLens 1M data set under the different models and inference algorithms considered.  The subsections that follow describe these results in detail.  To summarize our results, we find:

\begin{itemize}
\item The Variational algorithm tends to overfit,
\item The degree the Variational algorithm overfits is dependant on the choice of hyper-parameters, specifically the Wishart scale matrix $W_0$,
\item Modelling precisions can improve performance, though it may be necessary to bound the precisions for deterministic approximations,
\item The most significant gain in performance results from including side features to model correlational influence.  When side features are included, there is no gain from modelling precisions.
\end{itemize}

\begin{table}[t]
\begin{center}
\caption{Overall test error rates on the (a) MovieLens 1M and (b) Epinions data set under the precision and inference models considered. MAP estimate values are in parenthesis.}
\label{table:ml1mError}
\subfloat[][]{
\begin{tabular}{r|rccc|c}
\hline
& & Constant & Robust & Truncated ($n = 2$) & MAP\\
\hline 
\multirow{2}{*}{No Features}
  & Gibbs & \multicolumn{3}{c|}{\multirow{2}{*}{0.9101}} & \multirow{2}{*}{0.9210} \\
  & VI &  \\
  \hline
\multirow{2}{*}{No Side}
  & Gibbs & 0.8452 & 0.8448 & 0.8475& \multirow{2}{*}{0.8888} \\
  & VI\dag & 0.8546 & 0.8570 & 0.8521\\
\hline
\multirow{1}{*}{Side} 
  & Gibbs & 0.8407 & 0.8407 & &  \multirow{1}{*}{0.8805}\\
\hline
\end{tabular}
}\qquad
\subfloat[][]{
\begin{tabular}{r|rccc|c}
\hline
& & Constant & Robust & Truncated ($n = 2$) & MAP\\
\hline 
\multirow{2}{*}{No Features}
  & Gibbs & \multicolumn{3}{c|}{\multirow{2}{*}{1.0460}} & \multirow{2}{*}{1.1298} \\
  & VI &  \\
  \hline
\multirow{2}{*}{No Side}
  & Gibbs & 1.0455 & & & \multirow{2}{*}{1.1211} \\
  & VI & 1.0550 &  & \\
\hline
\multirow{1}{*}{Side} 
  & Gibbs$\ast$ & 1.0457 &  & &  \multirow{1}{*}{1.1134}\\
\hline
\end{tabular}
}
\\\dag These results are reported with an alternate choice of hyper-parameters, as discussed in the analysis below.
\\$\ast$ The sparsity of the Epinions data set limits the incremental benefit of side features for this data set.
\end{center}
\end{table}

\subsection{MAP Estimation}

A starting point for the Gibbs sampler and variational algorithm was obtained through a multi-step MAP estimation phase.  In the first step, we learned the user and item offsets, $\gamma_i, \eta_j$ through batch gradient descent on the sum-of-squares errors term with prediction $\hat{r}_{i,j} = \gamma_i + \eta_j$ with quadratic regularizers for the biases, 

\begin{equation}
\begin{aligned}
\frac{\tau}{2}\sum_{i=1}^{N}\sum_{j=1}^{M} I_{i,j}(r_{i,j} - (\gamma_i + \eta_j))^2
  + \frac{\lambda_\gamma}{2}\sum_{i=1}^{N}\gamma_i^2
    + \frac{\lambda_\eta}{2}\sum_{j=1}^{M}\eta_j^2
\end{aligned}
\end{equation}

By learning $\gamma_i, \eta_j$ prior to learning the user and item feature vectors, we ensure that the user and item biases are accounted for in these parameters, and not absorbed as constants in the feature vectors.  This is crucial as it prevents singularity issues with the variational algorithm.  Learning in this stage is terminated when the training prediction error converges.

With these offsets, we learn the user, item, and side feature vectors using batch gradient descent on the sum-of-squares errors term with quadratic regularizers for the features, 

\begin{equation}
\begin{aligned}
E =& \frac{\tau}{2}\sum_{i=1}^{N}\sum_{j=1}^{M}I_{i,j}(r_{i,j} - \hat{r}_{i,j})^2
    + \frac{\lambda_U}{2}\sum_{i=1}^{N}\|U_i\|_2^2 
    + \frac{\lambda_V}{2}\sum_{j=1}^{M}\|V_j\|_2^2
    + \frac{\lambda_K}{2}\sum_{k=1}^{M}\|W_k\|_2^2 \\
\end{aligned}
\end{equation}

Again, a common penalty and learning rate was assumed for all feature vectors to reduce the number of tuning parameters.  20-dimensional vectors were estimated to facilitate comparison with previous work in the literature \citep{laks2011rbmf}.

Finally, for the precision models, we set $\alpha_i, \beta_j, \tau$ to the maximum likelihood estimates based on the learned user / item parameters

\begin{equation}
\begin{aligned}
\label{map:precision}
\hat{\alpha}_i^{-1} =& 
  \frac{\hat{\tau}\sum_{j=1}^{M}I_{i,j}\hat{\beta}_j(r_{i,j} - \hat{r}_{i,j})^2}{\sum_{j=1}^{M}I_{i,j}} \\
\hat{\beta}_j^{-1} =& 
  \frac{\hat{\tau}\sum_{i=1}^{N}I_{i,j}\hat{\alpha}_i(r_{i,j} - \hat{r}_{i,j})^2}{\sum_{i=1}^{N}I_{i,j}} \\
\hat{\tau}^{-1} =&
  \frac{\hat{\tau}\hat{\alpha}_i\hat{\beta_j}\sum_{i=1}^{N}\sum_{j=1}^{M}I_{i,j}\hat{\beta}_j(r_{i,j} - \hat{r}_{i,j})^2}{\sum_{i=1}^{N}\sum_{j=1}^{M}I_{i,j}} \\
\end{aligned}
\end{equation}

These estimates are coupled, and we iterate through the set of precisions until convergence.  Note that learning the precisions after learning the features does not change the actual prediction.

\subsection{Variational Inference}
\label{sec:varinf}

\begin{figure}[t]
\centering
\subfloat[]{\includegraphics[scale=0.3]{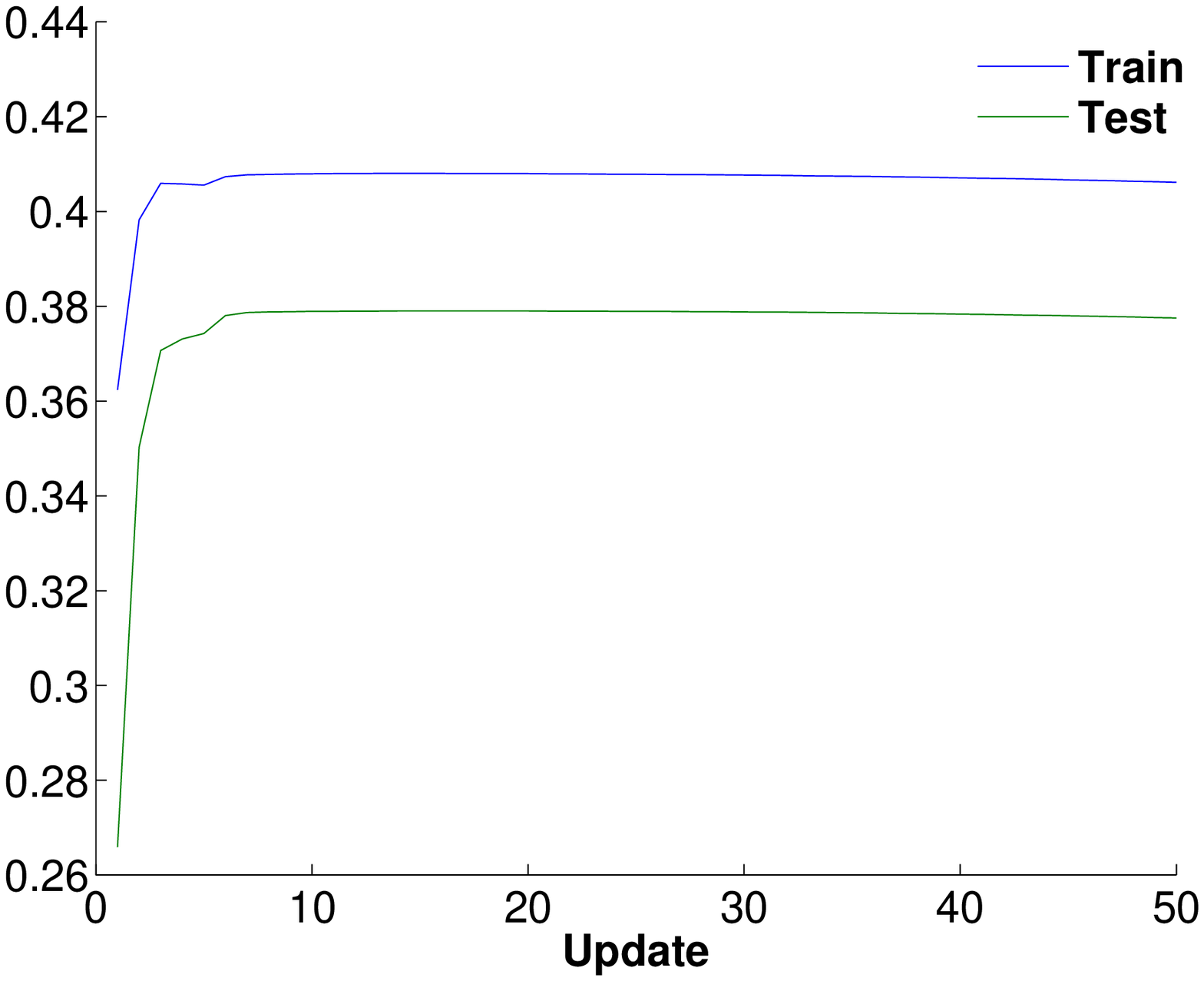}}
\subfloat[]{\includegraphics[scale=0.3]{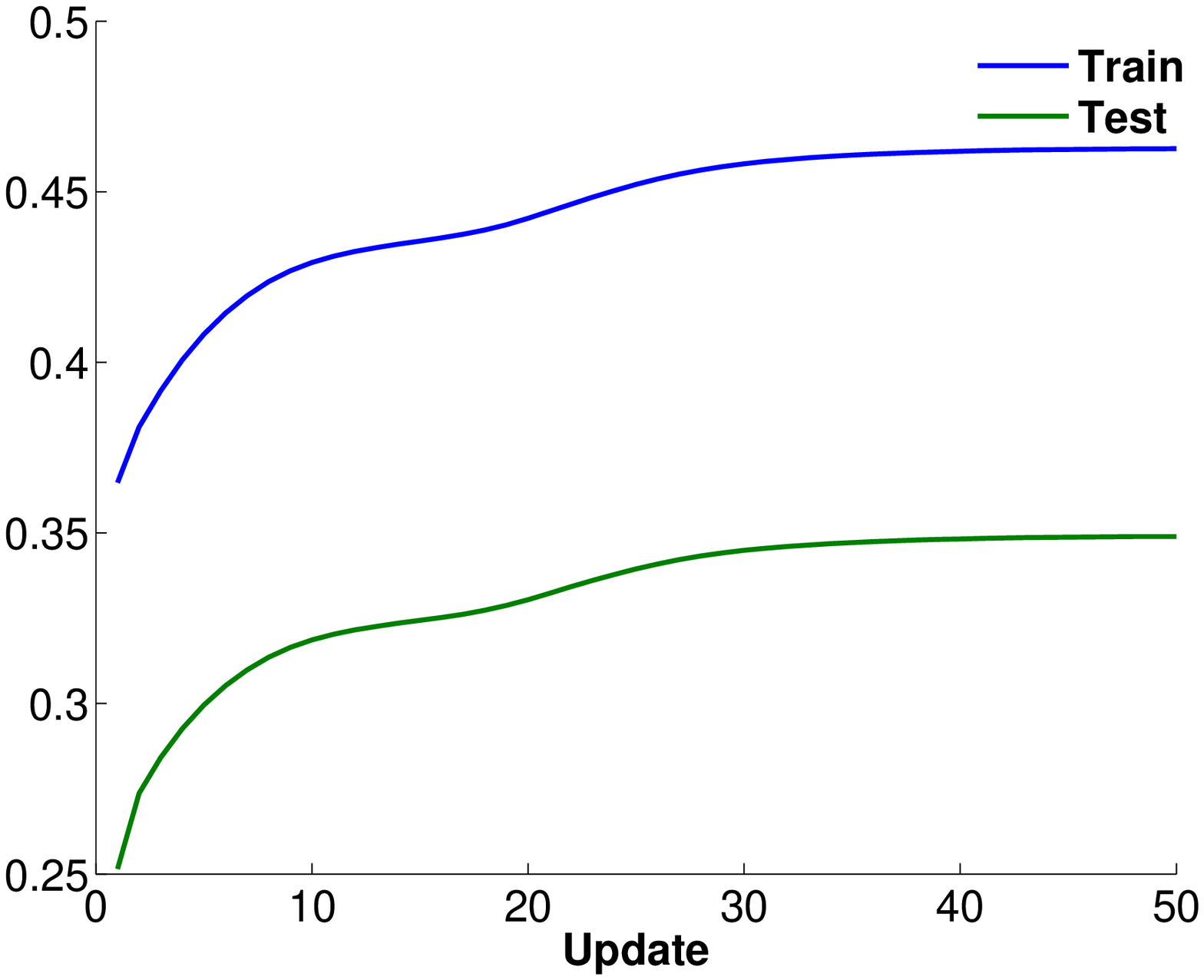}}
\subfloat[]{\includegraphics[scale=0.3]{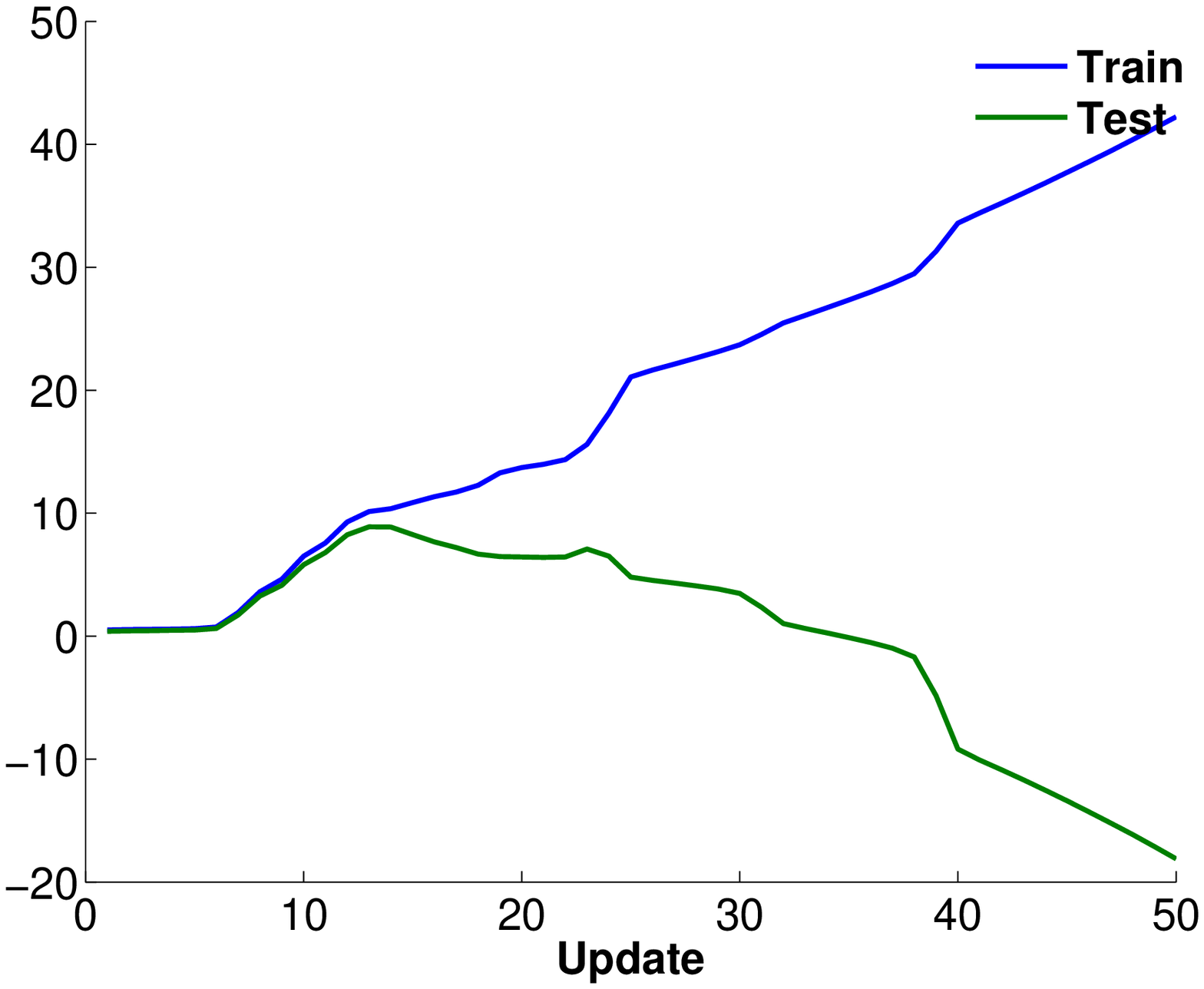}} \qquad
\subfloat[]{\includegraphics[scale=0.3]{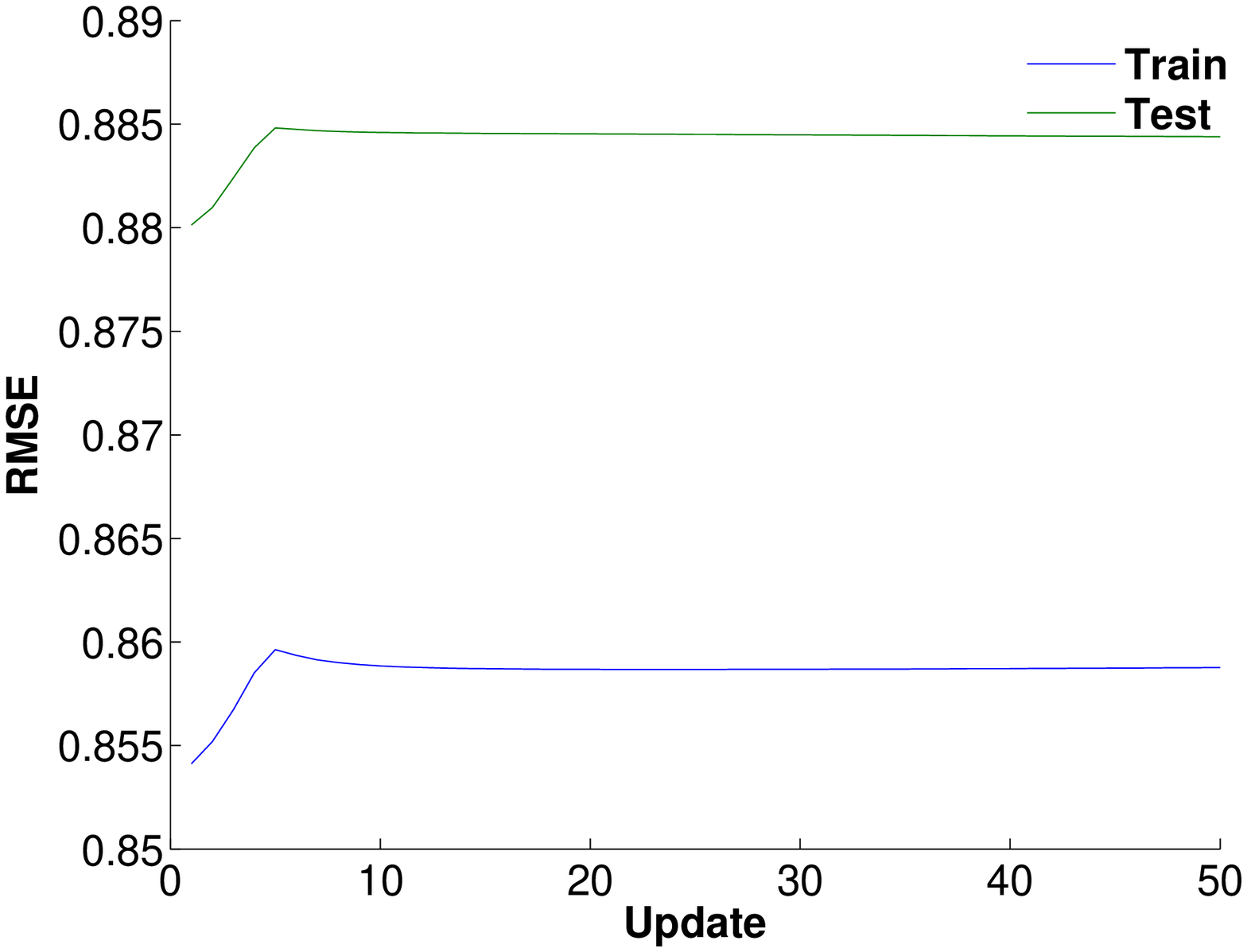}}
\subfloat[]{\includegraphics[scale=0.3]{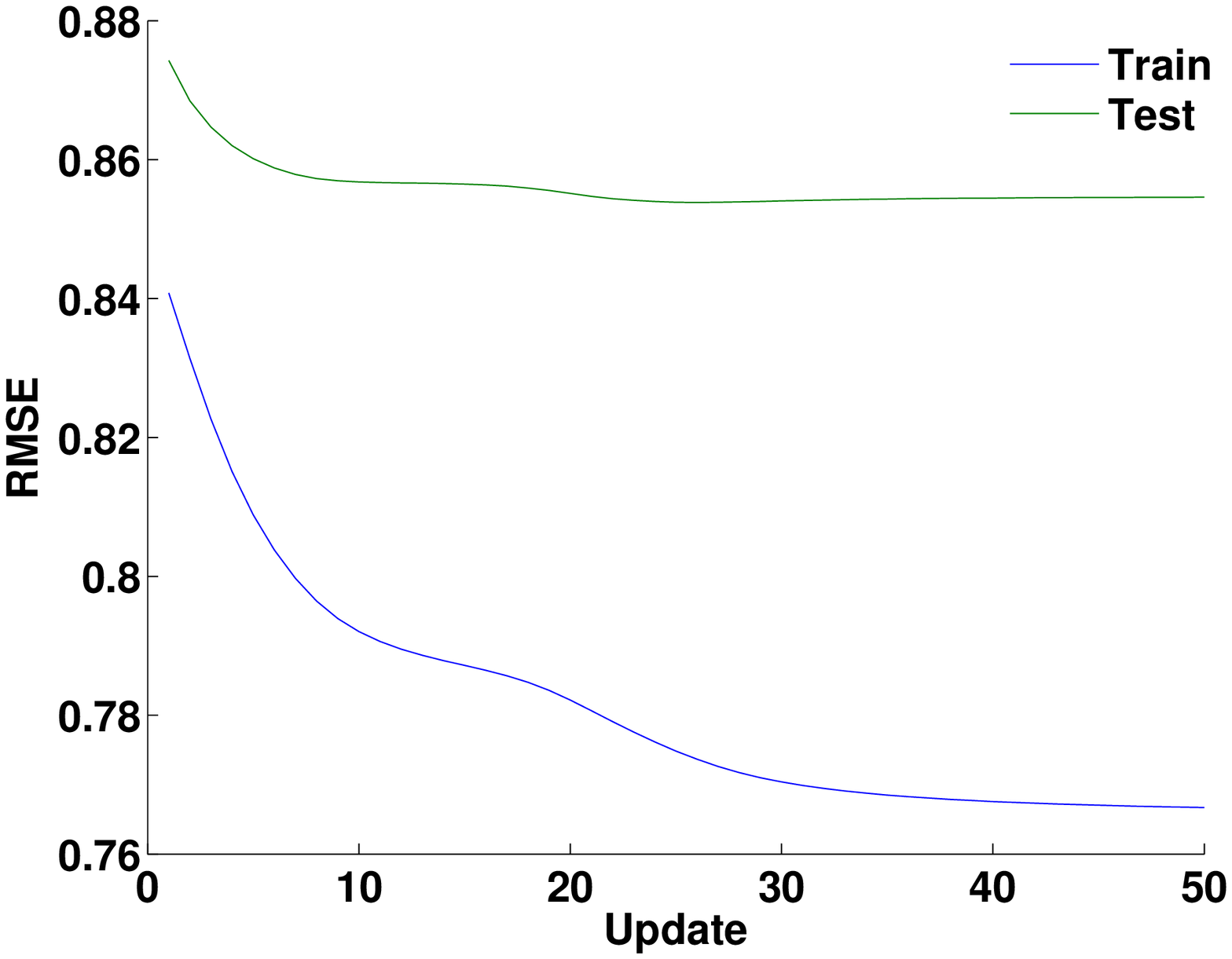}}
\subfloat[]{\includegraphics[scale=0.3]{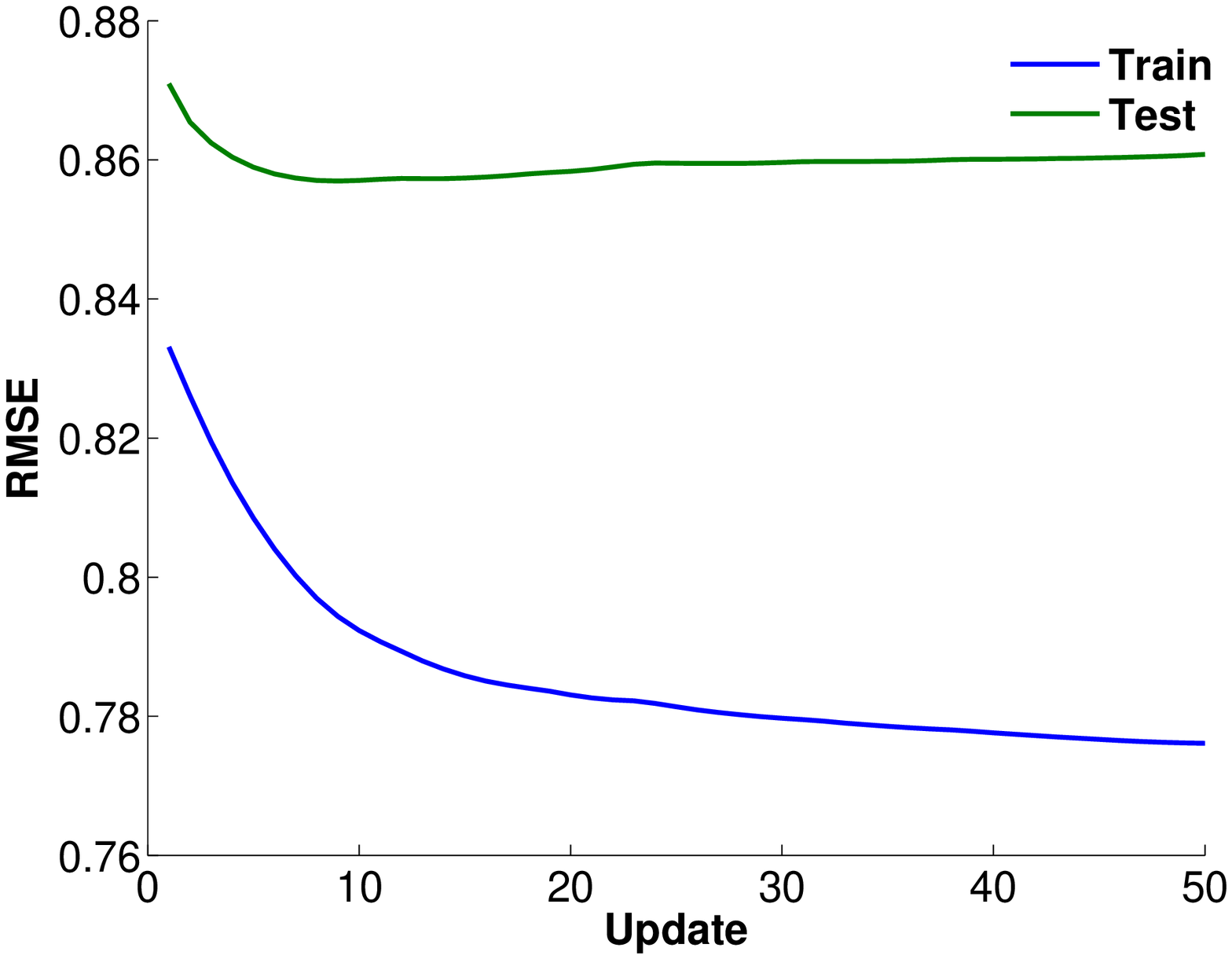}}
\caption{(top) Variational lower bound and (bottom) RMSE for the training and test sets for the (left) constant precision model with default hyper-parameters, (middle) constant precision model with alternative choice of hyper-parameters, and (right) robust precision model with alternative choice of hyper-parameters.}
\label{fig:var:lberror}
\end{figure}

We first consider the performance of the variational algorithm on the model with no side features, no precision factors, and with the default choice of hyper-parameters.  Under this setup, the variational lower bound monotonically increases and converges within the first 10 full updates of the parameter set.  Despite this, there is overfitting in both the training and test set.  The model is unable to improve upon the MAP estimate, see Figure \ref{fig:var:lberror} (left column).

Further investigation shows that the MAP estimate obtained is probabilistically unlikely under the prior selected, specifically the hyper-parameters of the Gaussian-Wishart priors.  The MAP values do not suggest a Wishart scale matrix set to the identity.  We rerun variational inference using a modified set of hyper-parameters.  Letting $U^{(0)}_i$ and $V^{(0)}_j$ denote the features obtained in the map estimate, we set

\begin{equation}
\begin{aligned}
W_0^{-1} = \frac{1}{2} \mathrm{diag}\left(\sum_{i=1}^{N} U^{(0)}_i {U^{(0)}_i}^\top\right) + \frac{1}{2} \mathrm{diag}\left(\sum_{J=1}^{M} V^{(0)}_j {V^{(0)}_j}^\top\right)
\end{aligned}
\end{equation}

For our map estimate, this creates a scale matrix $W_0$ with diagonal elements ranging from $38-82$.  We refer to this modified choice of prior as the ``MAP-driven'' prior.

This choice of hyper-parameters allows the variational algorithm to improve on the RMSE obtained by the MAP estimate, but is prone to overfitting on the test set prior to the convergence of the variational lower bound, see Figure \ref{fig:var:lberror} (middle column).  The training lower bound starts to converge after approximately 50 full updates of the variables, while the test error reaches a minimum of 0.8538 after 26 updates.  By the time the lower bound has converged, this has increased to 0.8546.

Assessing the convergence of the robust precision model, in practice, is more difficult.  The inclusion of user and item multiplicative precision factors allows the variational model to arbitrarily weight the contribution to the complete log-likelihood from different rows and columns in accordance with the predictive accuracy of the model.  We discuss this further when we compare the variational algorithm to the Gibbs sampler in Section~\ref{sec:precisions}.  The training lower bound increases almost linearly, with the training error smoothly dropping.  The test error reaches a minimum of 0.8570 after nine updates, while the lower bound on the test set continues to decrease for 13 updates, Figure \ref{fig:var:lberror} (right column).  This means that the lower bound on the test set cannot be used to assess convergence, as this happens after overfitting has started.

\subsection{Gibbs Sampling}
\label{sec:Gibbs}

\begin{figure}[htb]
\begin{center}
\subfloat[][]{\includegraphics[scale=0.3]{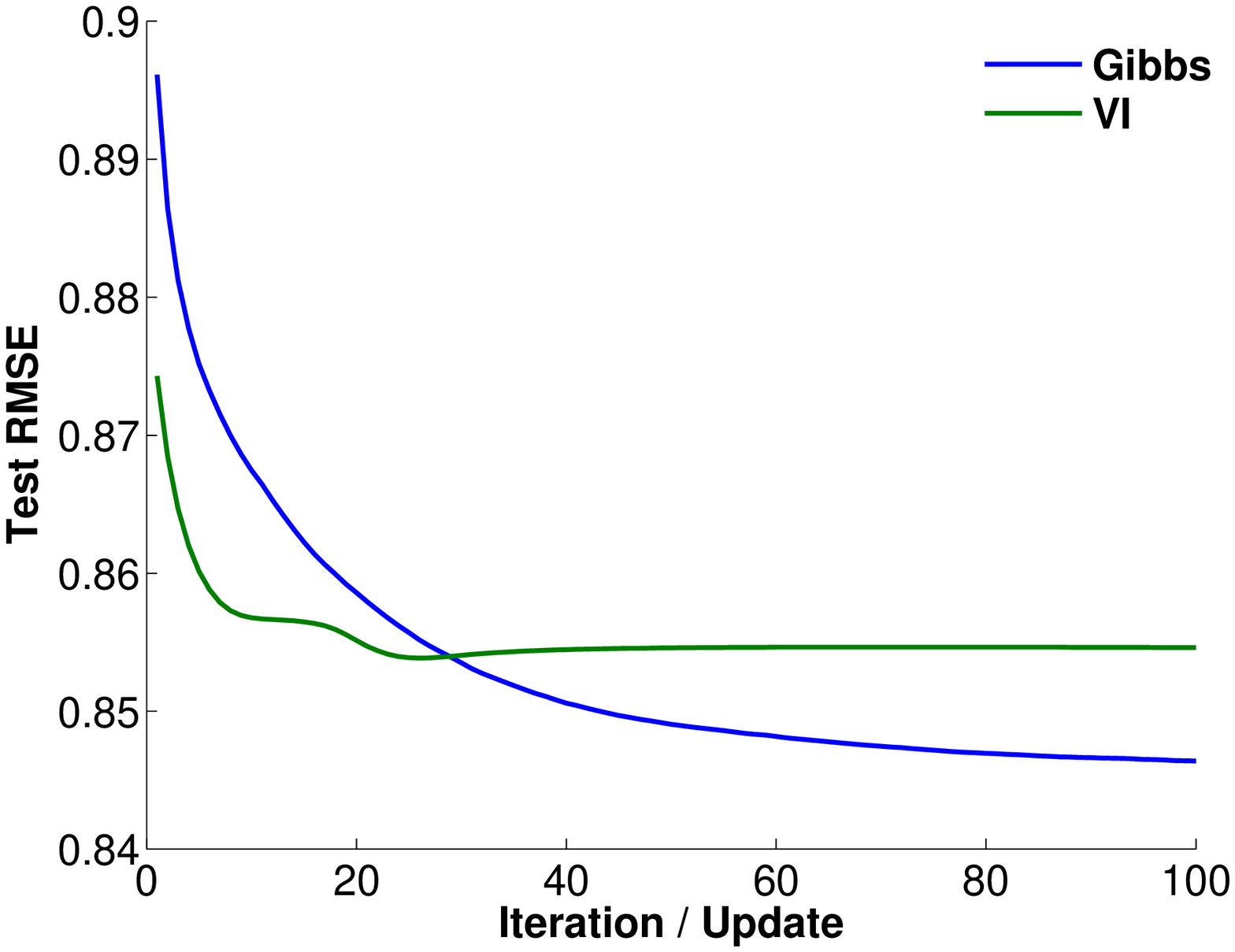}}
\subfloat[][]{\includegraphics[scale=0.3]{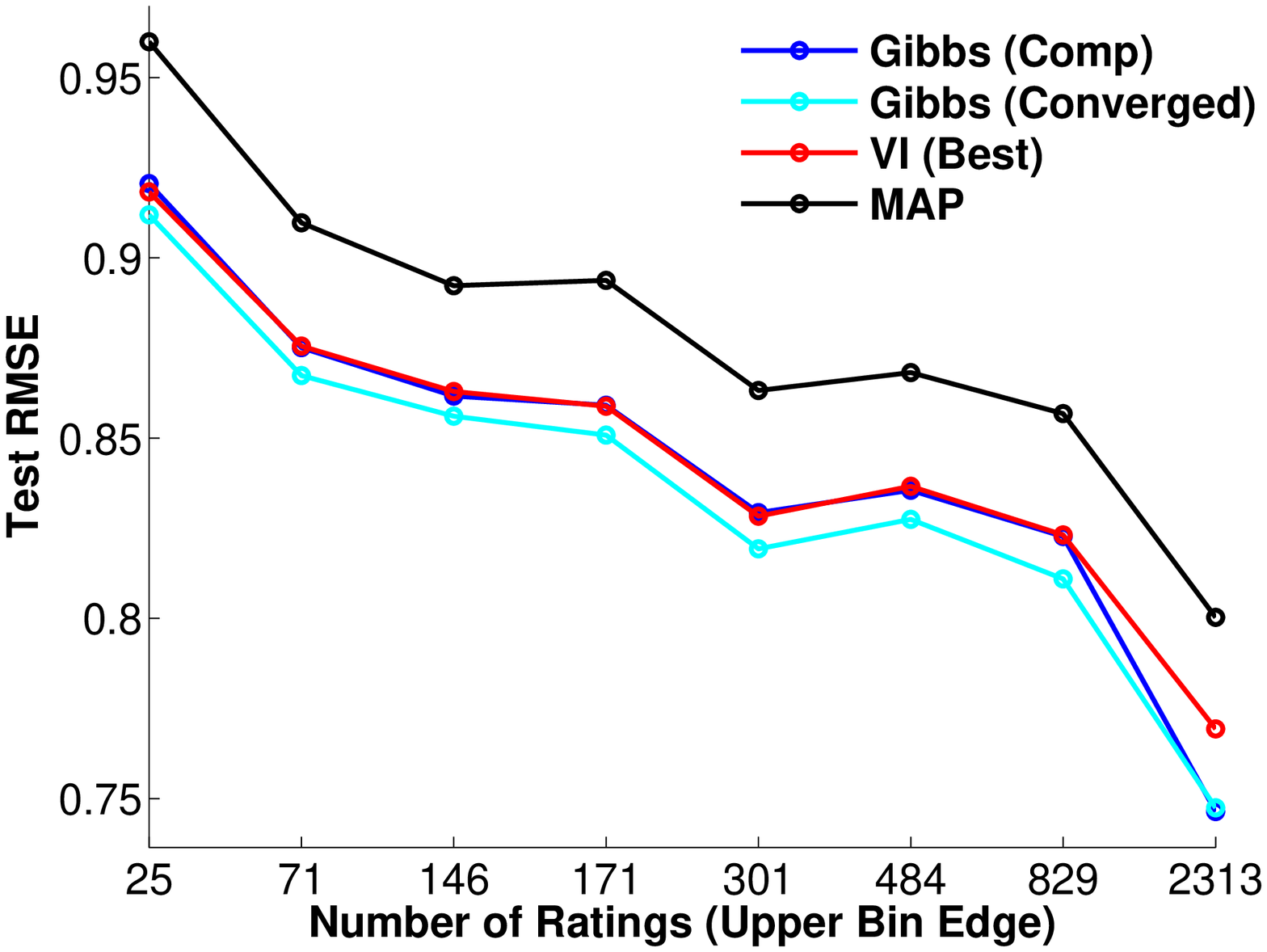}}
\subfloat[][]{\includegraphics[scale=0.3]{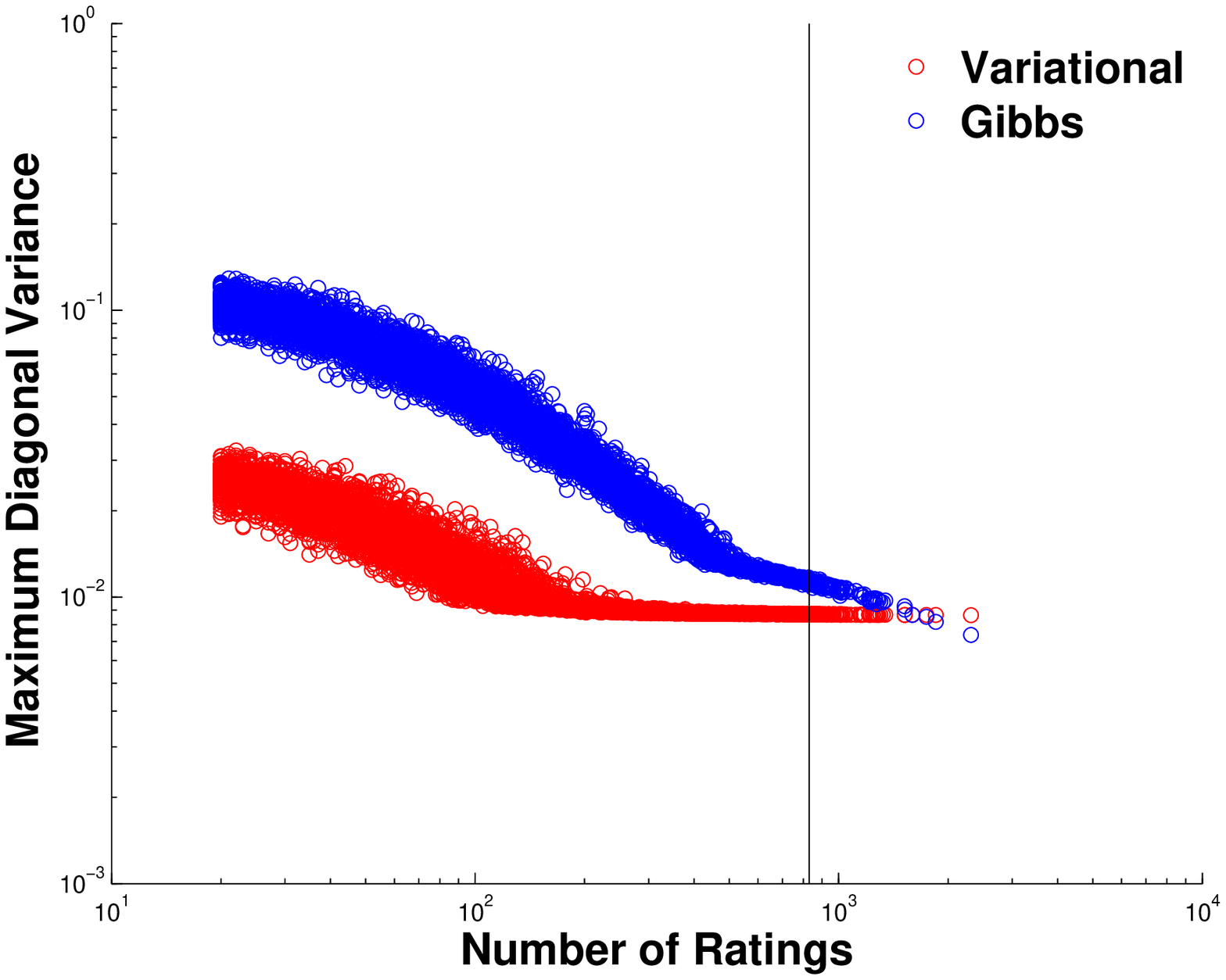}}
\caption{(a) Overall test error for the Gibbs sampler and variational algorithm for the constant precision models without side features, with the performance of the MAP estimate included for comparison.  (b) The test error with respect to user frequency.  The two are nearly identical for users of all frequency, with the exception of the most frequent users.  In this bin, the Gibbs sampler outperforms the variational approximation.  (c) Maximum variance for the user features plotted against the user frequency under both Gibbs sampling and the Variational algorithm after convergence.}
\label{fig:gibbsVarError}
\end{center}
\end{figure}

With a fixed choice model (defined by a set of latent features and precision), inference via Gibbs sampling outperforms the Variational mean field approximation.  This comparison is trivial when we consider the default choice for the Wishart scale matrix of $W_0 = \mathbf{I}_{d\times d}$.  For a more interesting comparison, we consider the performance of the Gibbs sampler with the default prior and the variational algorithm with the ``MAP-driven'' prior.  For simplicity, we focus our discussion on the basic model without precisions and without side features.  We select the Gibbs iteration and Variational update for which the overall test errors are near equal.  From Figure~\ref{fig:gibbsVarError} (a), this is the 29$^\mathrm{th}$ update of the parameters under variational inference, and the 29$^\mathrm{th}$ iteration of the Gibbs sampler.

With $W_0$ driven by the MAP estimate, the variational algorithm outperforms the Gibbs sampler in overall test error for approximately the first 30 iterations.  This performance gain is motivated by drops in the first five iterations.  After the first five iterations, the variational algorithm experiences diminishing returns.  However, the Gibbs sampler continues to drop at a similar rate beyond this point.

The two algorithms have approximately the same overall error rate on the test set, to within $0.0001$, after the 29th iteration / update.  Figure \ref{fig:gibbsVarError} (b) illustrates the error of the two inference methods at this point with respect to user frequency.  The difference in the performance of the two algorithms is on the order of $0.001$ or less, except for the most frequent bin.  This bin corresponds to the $10\%$ most frequent users, and the Gibbs sampler outperforms the variational algorithm.

The performance gap between the Gibbs sampler and the Variational approximation for the most frequent users suggests that the sampling distribution of the user feature vectors has noticeable variability.  Figure~\ref{fig:gibbsVarError} (c) illustrates this by plotting the maximum variance of the $d$-dimensional user feature against the number of ratings the user has in the training set.  These are representative values for both inference algorithms after convergence.  This space between the plot for the Gibbs sampler and the Variational approximation indicates that the variational approximation tends to produce smaller estimates of the variance than the Gibbs sampler.  The vertical line represents users with 829 ratings in the training set, which is the value above which users are included in the last bin in Figure~\ref{fig:gibbsVarError} (b).  The persistent significant difference between these two beyond this point means that there is still variability in the distribution of the user features that the Gibbs sampler is exploiting.

\subsection{Precisions}
\label{sec:precisions}

We hope to determine if modelling precisions tends to improve performance for users of a given frequency (cold start, rare, frequent, etc.).  To determine this, we examine the final converged error rate on a user frequency basis for a model with and without precisions, holding all else equal.    For this comparison, we select the model with side features, and look at the error under Gibbs sampling.  The numerical results are in Table \ref{table:userTestFreq}.  

While there are some minor departures from equality for moderately frequent users (the third and fourth bin), they are in the fourth decimal place of the error, representing less than a 1\% relative change in predictive performance.  The largest difference occurs for the most frequent users.  This is a relative gain of $0.42\%$.  However, these users have predictions that are already well calibrated relative to the rest of the population.  Based on these results, we conclude that modelling precisions does not tend to significantly favour users of any given frequency in the test set.

\begin{figure}[htb]
\begin{center}
\subfloat[]{\includegraphics[scale=0.3]{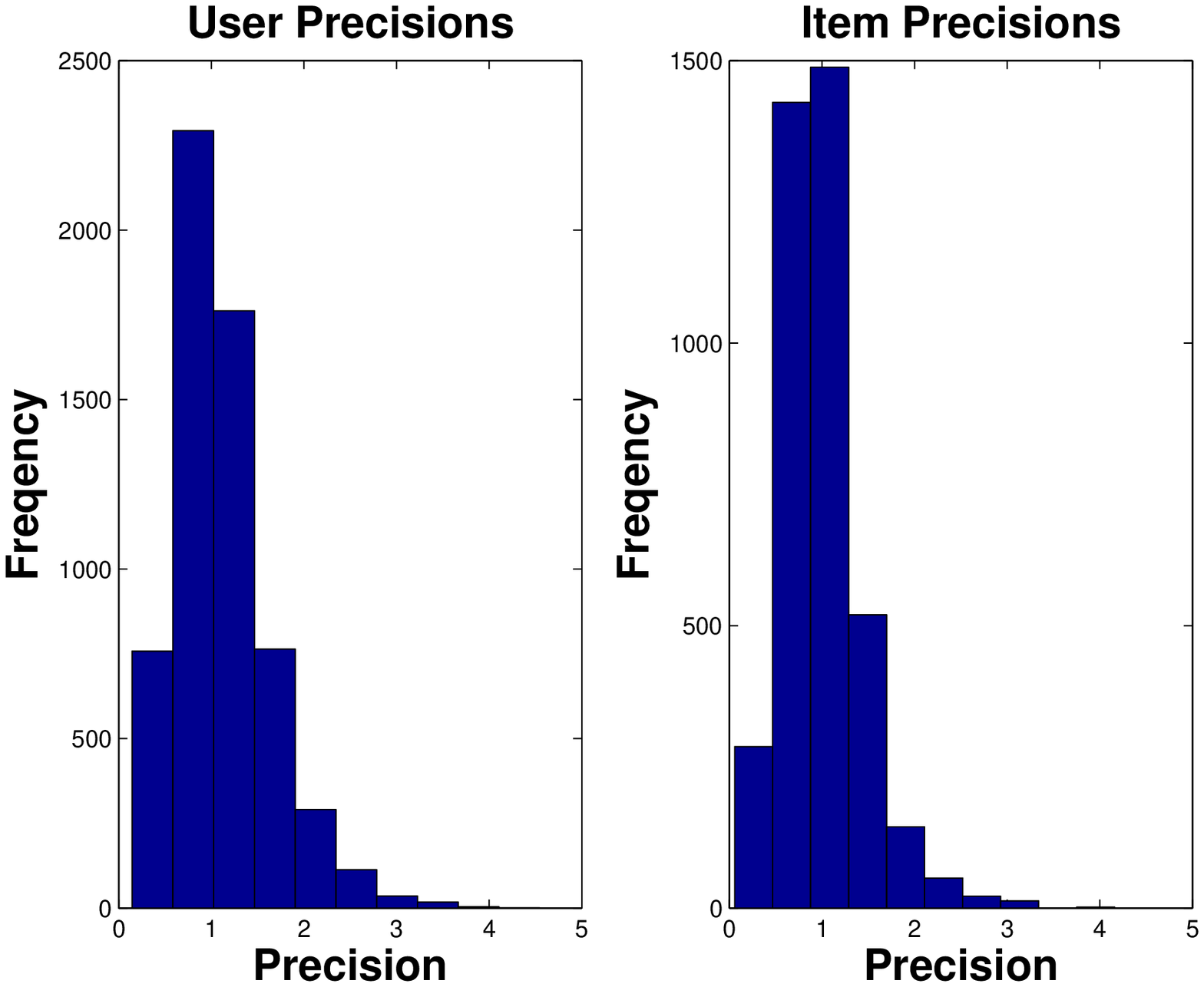}}
\subfloat[]{\includegraphics[scale=0.3]{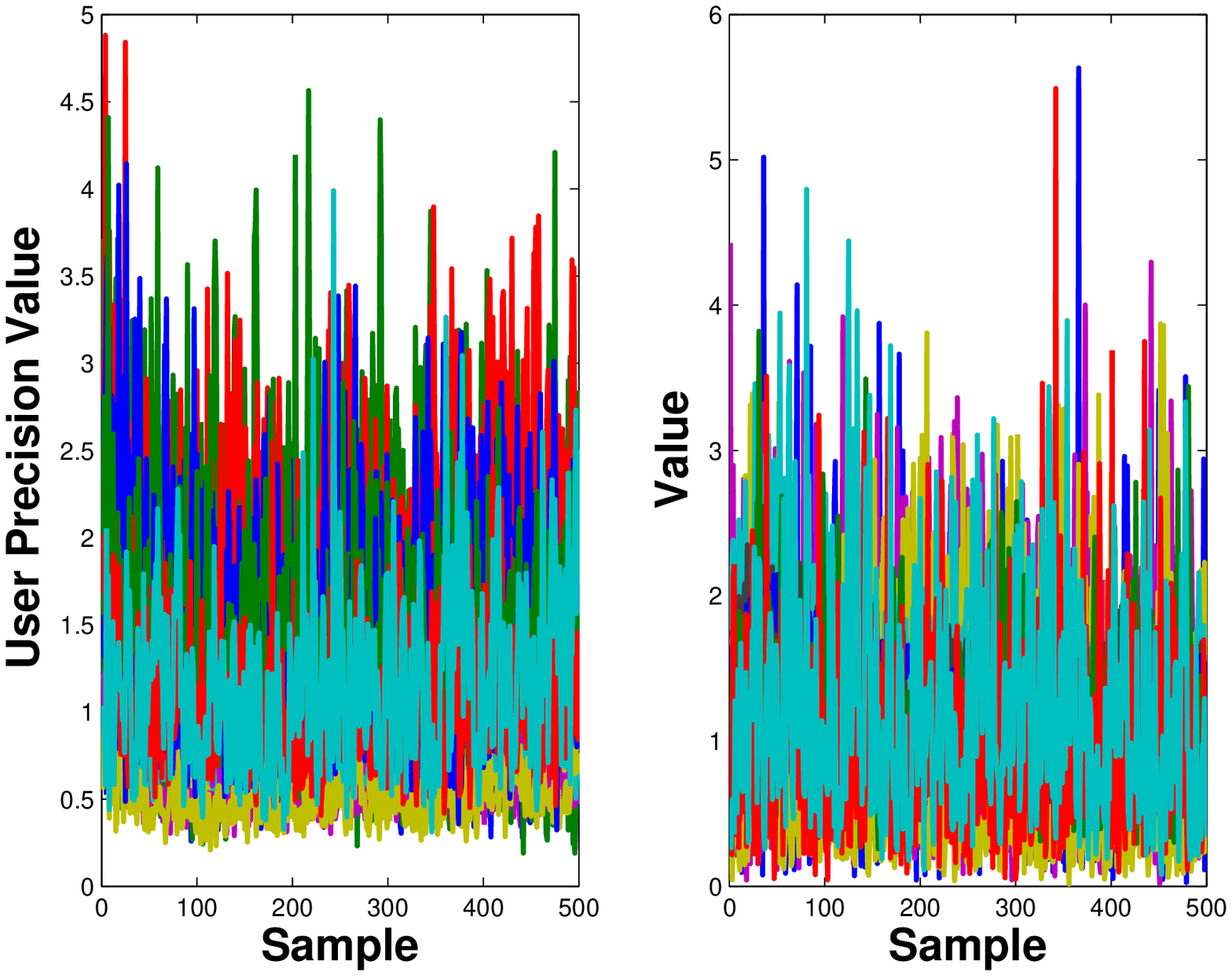}}
\subfloat[]{\includegraphics[scale=0.3]{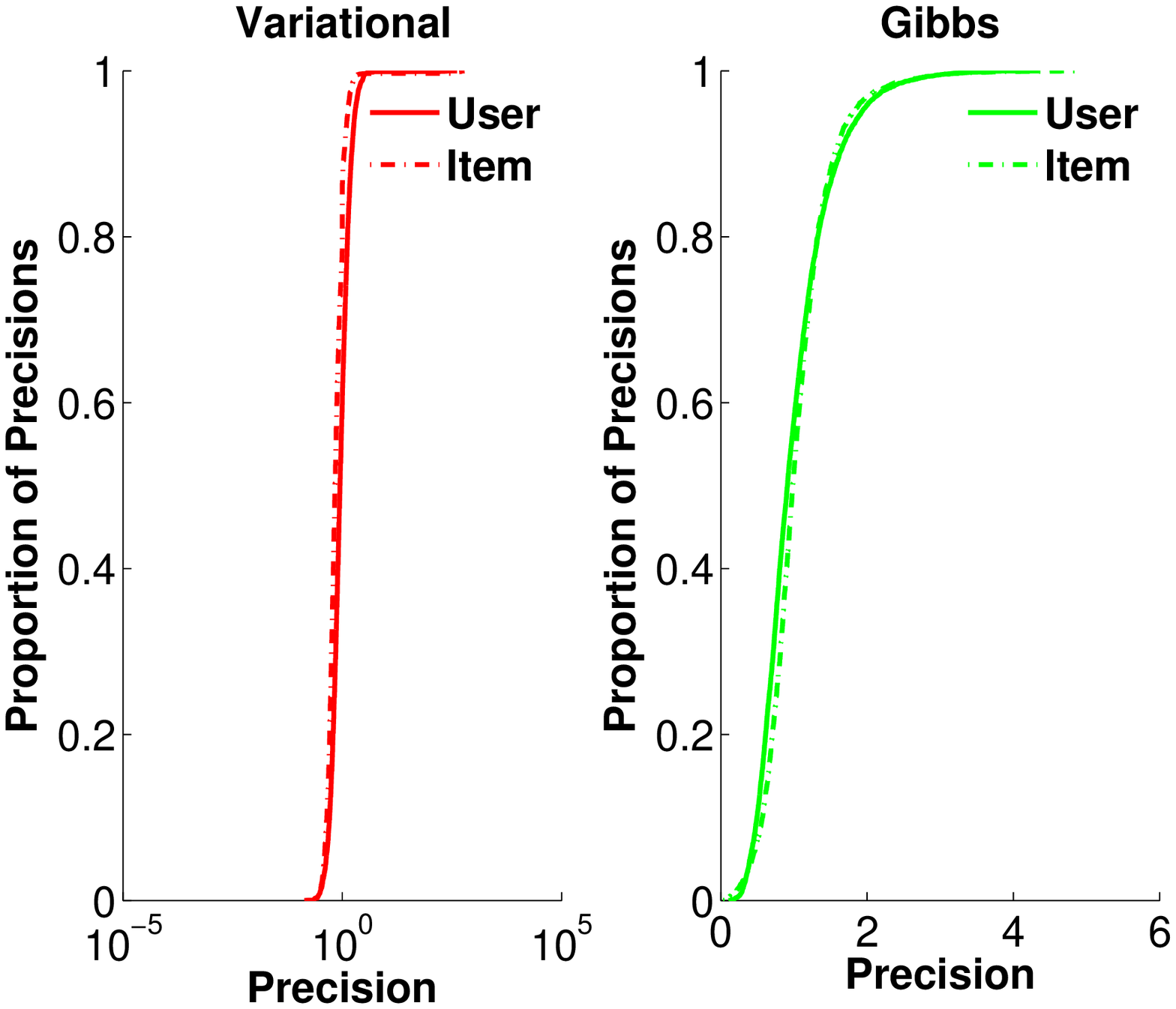}}
\caption{(a) Histograms of the user and item precisions from a Gibbs sample after convergence. (b) Traceplots of the user and item precisions (c) CDF curves of the converged user / item precisions for the model under (left) Variational and (right) Gibbs. Note that the curves have similar shape, but that the variational cdf is plotted on a log-scale for comparison.}
\label{fig:precisions}
\end{center}
\end{figure}

The near equality of both the constant and robust precision models could be a result of a near-constant posterior for the precisions.  To check this, we examine traceplots for the user and item precisions, as well as the histogram of the precisions for a sample at convergence.  The distribution of both the user and item precisions was clearly non-constant and right skewed. Figure \ref{fig:precisions} (a) displays histograms of the user and item precisions from a Gibbs sampler after convergence, indicating this skewness.  Figure~\ref{fig:precisions} (b) displays traceplots for a sample of the user and item precisions, showing the sampler mixed well over a range of values.  Both of these indicate that the sampler was exploiting the robust precision model.  Therefore, the near equality in predictive performance is not a result of model degeneracy.

It has been noted in Section~\ref{sec:varinf} that the introduction of precisions prompts the variational algorithm to drive some precisions arbitrarily small and arbitrarily large.  The histograms in Figure~\ref{fig:precisions} (a) and traceplots in Figure~\ref{fig:precisions} (b) indicates that the Gibbs sampler does not suffer from this limitation.  For comparison, empirical CDF curves are plotted in Figure~\ref{fig:precisions} (c) for the user and item precisions under the variational algorithm (left panel) and the Gibbs sampler (right panel) after convergence.  The two panels are similar, though the CDFs for the Variational algorithm have been plotted on a log scale.  In other words, the converged values of the precisions under Variational inference are exponentially larger.

\begin{table}[tb]
\begin{center}
\caption{Test RMSE broken down by user frequency under Gibbs sampling for the models with and without user / item precisions when side features are included.}
\label{table:userTestFreq}
\begin{tabular}{rr|rrr}
&     & \multicolumn{2}{c}{Error} & Relative \\
Quantile & Number of Ratings & Constant & Robust & Change (\%)\\
    \hline
(0.01) & $<=25$ & 0.9035 & \textbf{0.9031} &  $+0.05$\\
(0.10) & $26-71$     & \textbf{0.8608} & 0.8615 &  $-0.08$\\ 
(0.25) & $72-146$    & 0.8494 & \textbf{0.8478} & $+0.19$\\
(0.30) & $147-171$    & 0.8459 & \textbf{0.8449} & $+0.11$\\
(0.50) & $172-301$    & \textbf{0.8155} & 0.8165 & $-0.12$\\
(0.70) & $302-484$    & \textbf{0.8243} & 0.8250 & $-0.08$\\
(0.90) & $485-829$    & 0.8107 & \textbf{0.8098} & $+0.10$\\
(1) & $830-2,313$    & 0.7474 & \textbf{0.7443} & $+0.42$\\
\hline
\end{tabular}
\end{center}
\end{table}

\subsection{Truncated Precisions}

It was noted in Section~\ref{sec:varinf} that the introduction of precision lead to pathological results with variational inference.  We explored if bounding the precisions would alleviate this issue.  Using the truncated approach discussed in Section~\ref{sec:truncated}, we ran experiments bounding the precisions to values sensible for a scale constrained to the interval $[1,5]$.  Some of the initial precision values from the converged values of Equations~\eqref{map:precision} are outside this region.  As is appropriate based on our discussion of the MAP estimate for the truncated model, we set such values to the closest boundary point.

We found that bounds of $(1/2, 2)$ produced results for MovieLens that outperformed both the constant and robust precision model.  These values also delayed the overfitting in the variational approximation for several updates.  Overfitting for this model only starts after 20 full updates of the parameters.

\begin{figure}[t]
\centering
\subfloat[][]{\includegraphics[scale=0.3]{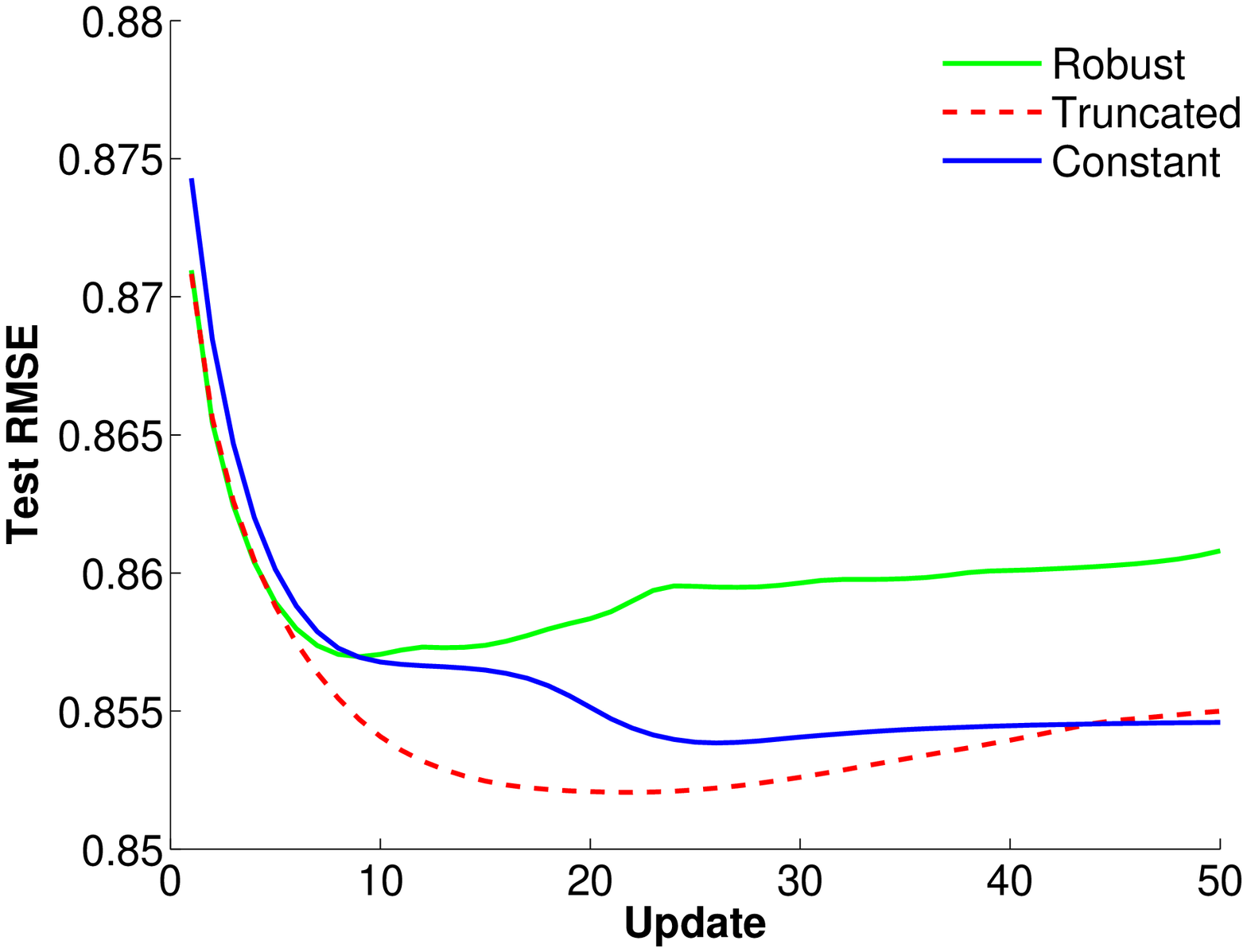}}
\subfloat[][]{\includegraphics[scale=0.3]{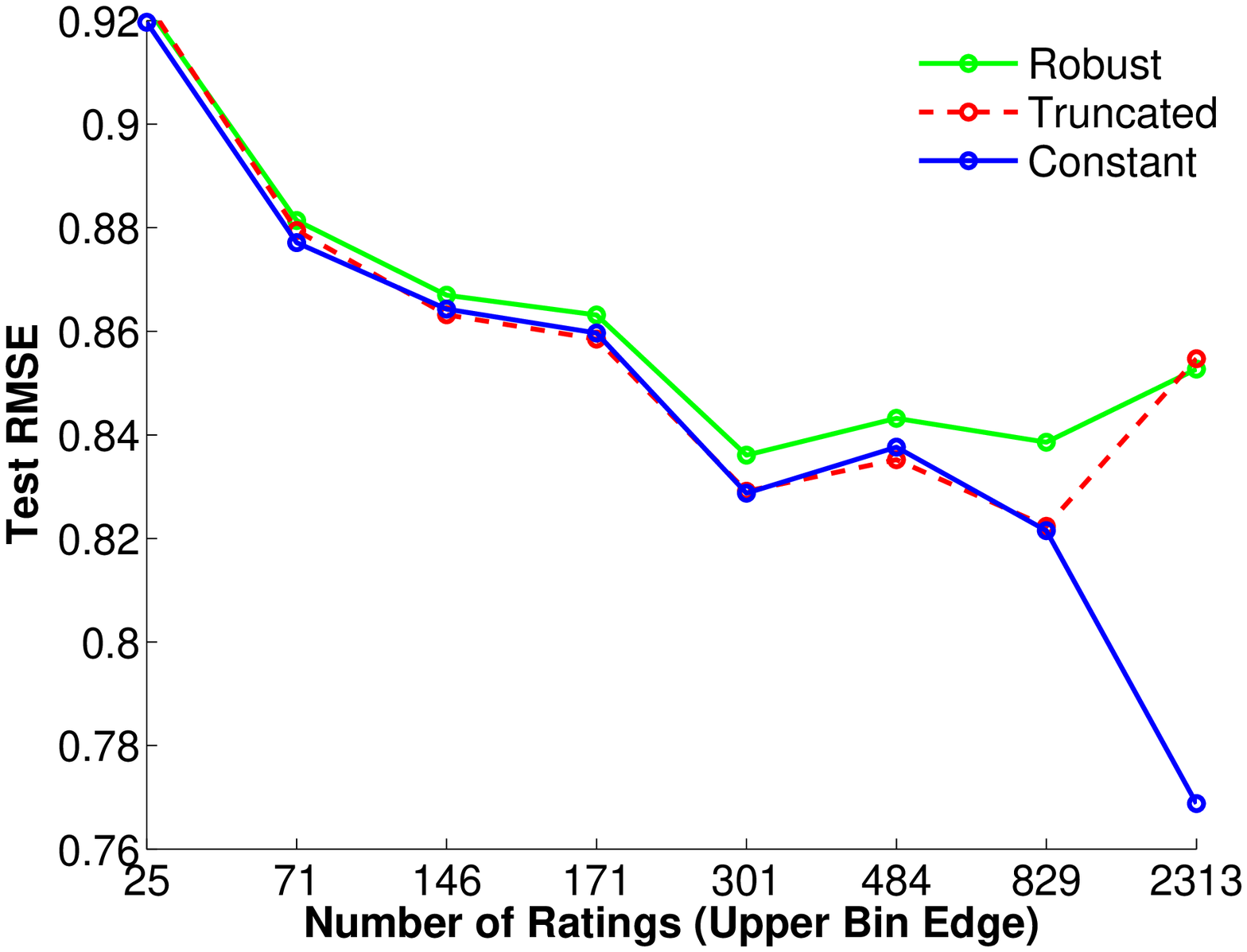}}
\subfloat[][]{\includegraphics[scale=0.3]{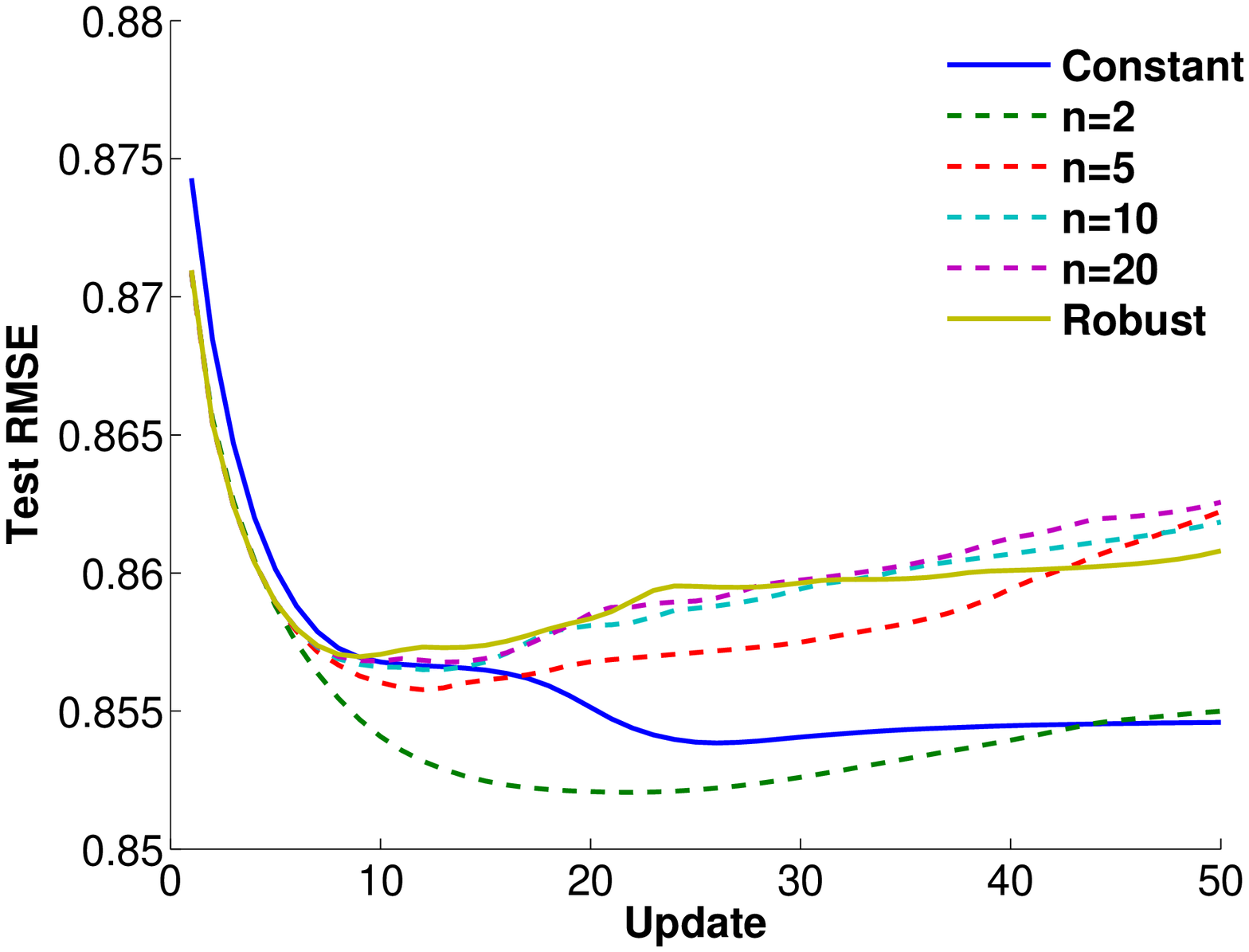}}
\caption{(a) Overall test error and (b) test error binned by user frequency for the variational algorithm under the robust and truncated precision model (with precision bounds $(1/2,2)$).  (c) Overall test error for the variational algorithm under the robust, constant, and truncated precision models for different choices of bounds.}
\label{fig:truncPrec}
\end{figure}

Figure~\ref{fig:truncPrec} (a) shows the overall test error of the variational algorithm under the constant, robust, and truncated precision model with the bounds of $(1/2,2)$.  These three models have similar behaviour in the initial set of parameter updates.  Differences start to appear after the eighth update.  At this point, the algorithm overfits on the robust precision model, and continues to drop on the truncated precision model for 3-4 additional iterations.  The rate of increase for the two is approximately the same until nearly the 40th parameter update, at which point the truncated precision model tends to increase at a faster rates.

Figure~\ref{fig:truncPrec} (b) shows the test error by user frequency for the algorithm under the two models after 50 full parameter updates.  This is the point that the algorithm has begun to overfit in the truncated and the robust model, and has appeared to stabilize for the constant precision model.  This graph shows the most significant difference is in the most frequent users.  The constant precision model outperforms  either heteroskedastic model by a difference of at least 0.1 in test RMSE, a relative improvement in RMSE of $12\%$.  The error rates are approximately the same in other user bins, with the constant precision model performing slightly worse for moderately frequent users.

With bounds of the form $(\ell,u) = (1/n,n)$, the truncated model has limiting cases of the constant model as $n\rightarrow 1$ and the robust model as $n \rightarrow \infty$.  This leads to the question of how inference on the truncated precision model perform as $n$ changes?  Figure \ref{fig:truncPrec} (c) plots the test error of the variational algorithm for several values of $n$ along with the constant and robust precision model.  As expected, larger values of $n$ are similar to the robust precision curve, while smaller values are similar to the constant precision curve.  What is surprising is the curve for $n = 2$, corresponding to the precision bounds $(1/2, 2)$.  The variational model obtains a significantly lower error rate under these bounds than the others choices of $n$, or the constant precision model.  The consistent tendency for the truncated model to overfit early in learning for larger values of $n$ suggests that the truncation value has little influence on performance after a certain point.  However, the improvement for the $n=2$ case over the constant precision model does indicate there is value in allowing for heteroskedastic precision among different users and different items.

The updates for the precision parameters are inversely proportional to the error for that user or item, scaled by the other values of the precisions.  In particular, recall that

\begin{equation}
\begin{aligned}
\label{eq:precusererror}
\hat{\alpha}^{-1}_i \propto& \sum_{j=1}^{M}I_{i,j}\tau\beta_j(r_{i,j}-\hat{r}_{i,j})^2
\end{aligned}
\end{equation}

\begin{figure}[t]
\centering
\subfloat[][]{\includegraphics[scale=0.4]{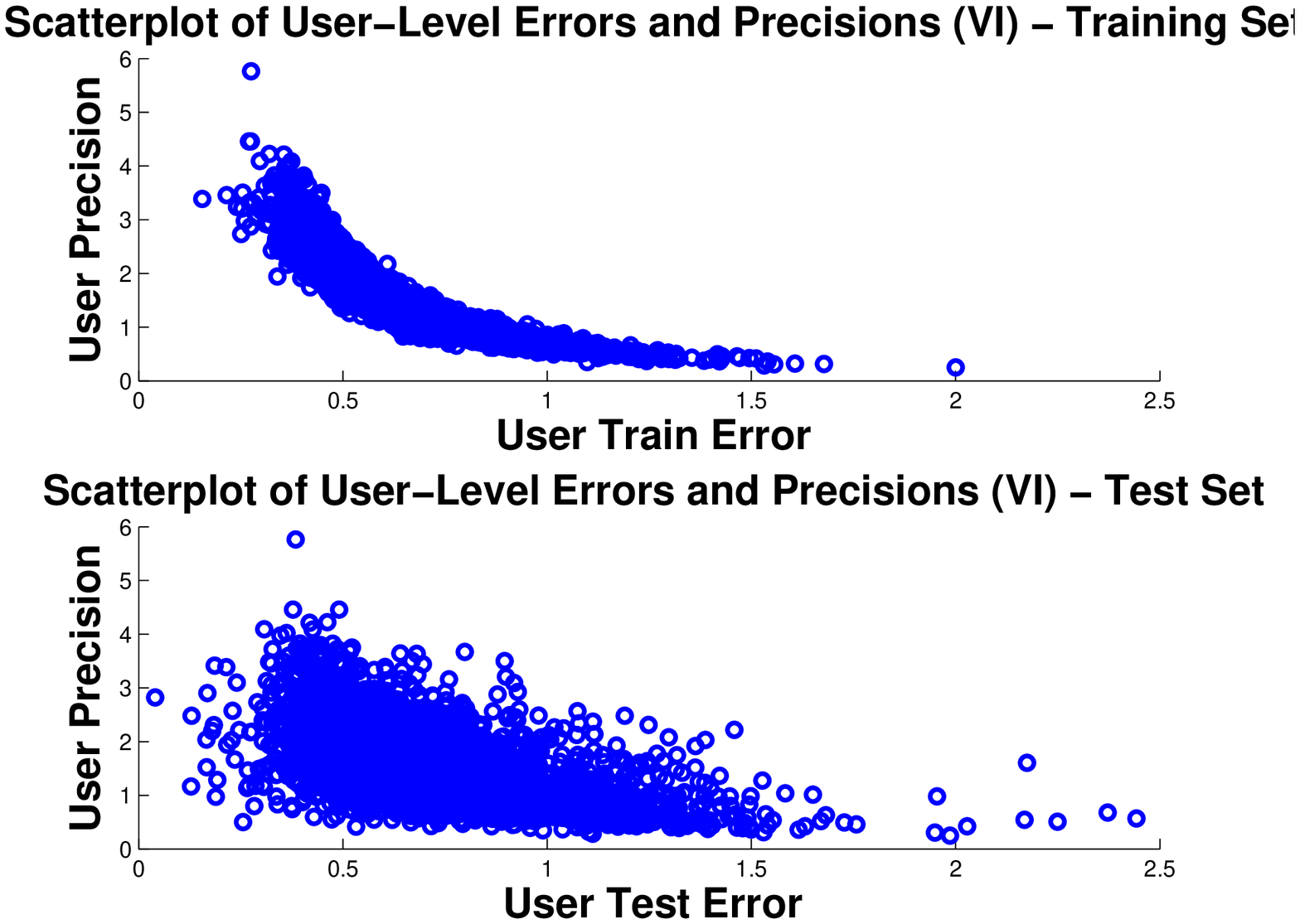}}
\subfloat[][]{\includegraphics[scale=0.4]{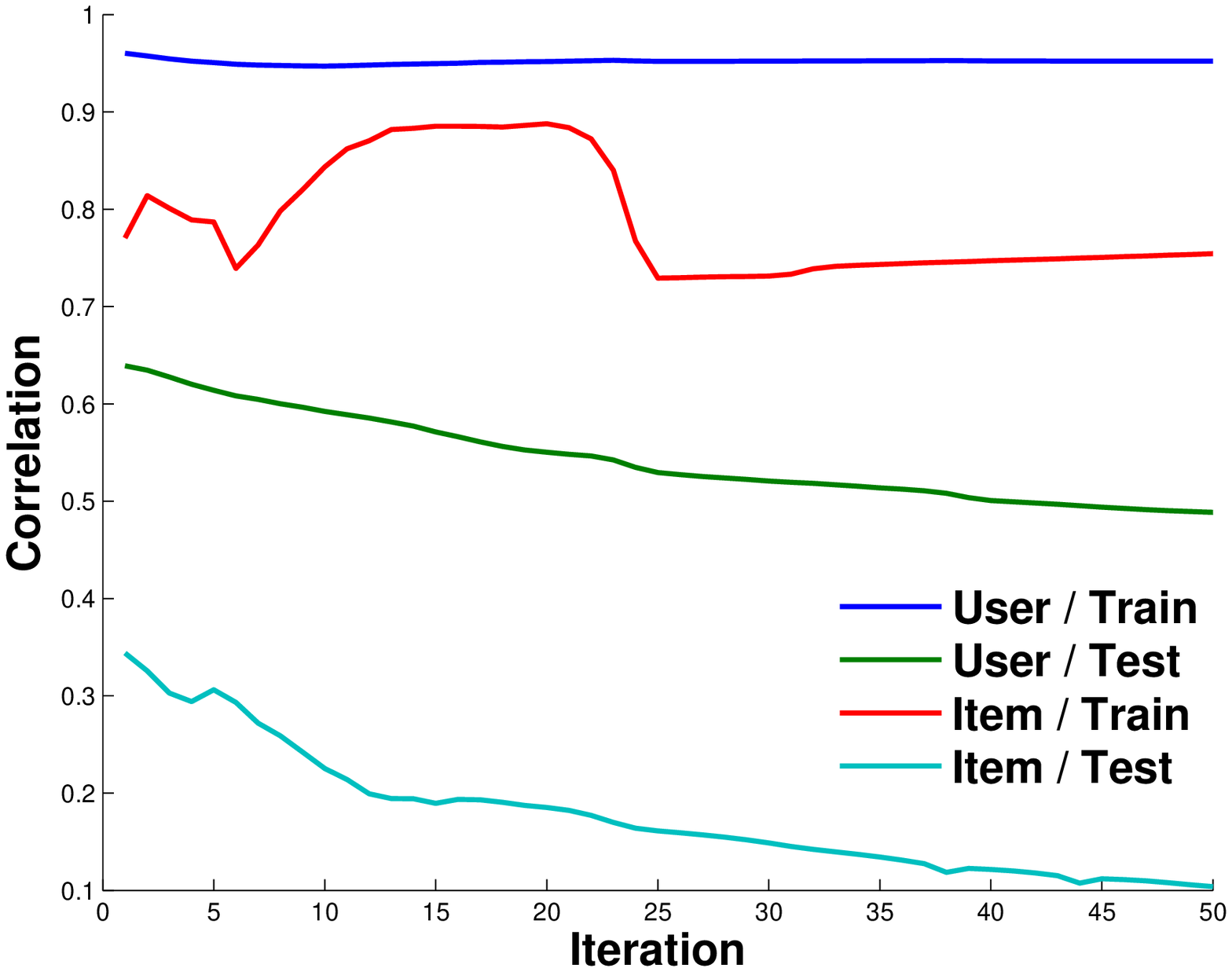}}
\caption{(a) Scatterplot of user/item level errors and precisions after the third full parameter update in the variational algorithm. (b) Correlation between (log transformed) user/item level errors and precisions over parameter updates.}
\label{fig:robustPrec:corr}
\end{figure}

A scatterplot of user precisions versus user training error will show this inverse relationship, similar for items.  A good test of generalization is to see if this inverse relationship holds on the test set.  Figure~\ref{fig:robustPrec:corr} (a) shows scatterplots of the user and item level errors and precisions for the training (top) and test (bottom) after the third full parameter update.  The inverse relationship is clear in the training set, but weak in the test set.

Log transforming Equation~\eqref{eq:precusererror} yields:

\begin{equation}
\begin{aligned}
\label{eq:precusererror:log}
-\log(\hat{\alpha}_{i}) \propto& \log\left(\sum_{j=1}^{M}I_{i,j}\tau\beta_j(r_{i,j}-\hat{r}_{i,j})^2\right),
\end{aligned}
\end{equation}

Equation~\eqref{eq:precusererror:log} suggests there should be strong linear correlation between the set of log precisions and log error rates on a user / item level basis for both the training and test set.  We compute these for the set of users and items in both the training and test sets, and plot these correlations over iterations in Figure~\ref{fig:robustPrec:corr} (b).  The correlations computed on the training set are typically large and stable over iterations.  The user correlation is consistently above $0.94$, while the item correlation is consistently above $0.70$.  They are not exactly 1 since the updates are sequential, while the correlations are computed after a full parameter update.

When the same values are computed for the user and item errors in the test set, significantly smaller values are obtained, and they decrease  monotonically over parameter updates.  By the the point the model overfits in the ninth full parameter update, the correlation in the test set for the items has dropped from $0.3440$ to $0.2422$, while the correlation in the test set for the users has dropped from $0.6391$ to $0.5965$.  The large difference between the training and test set, both in initial values and in the magnitude of the drop over iterations, shows that the robust model is overfitting and not generalizing to the test set.

We ran similar experiments with truncated precisions with the Gibbs sampler.  We did not find that inference with the Gibbs sampler was consistently improved by using truncated precisions.  This is not unexpected given that the histograms and traceplots of the precisions in Figure~\ref{fig:precisions} (a)-(b) indicate the precisions remain at sensible values under the Gibbs sample.

\subsection{Side Features}

When side features are included in the model, the incremental gain from using a robust precision model is lost, see Table \ref{table:ml1mError}.  The Gibbs sampler converges to the same RMSE value for both the constant and robust precision model.

\begin{figure}[t]
\centering
\subfloat[][]{\includegraphics[scale=0.4]{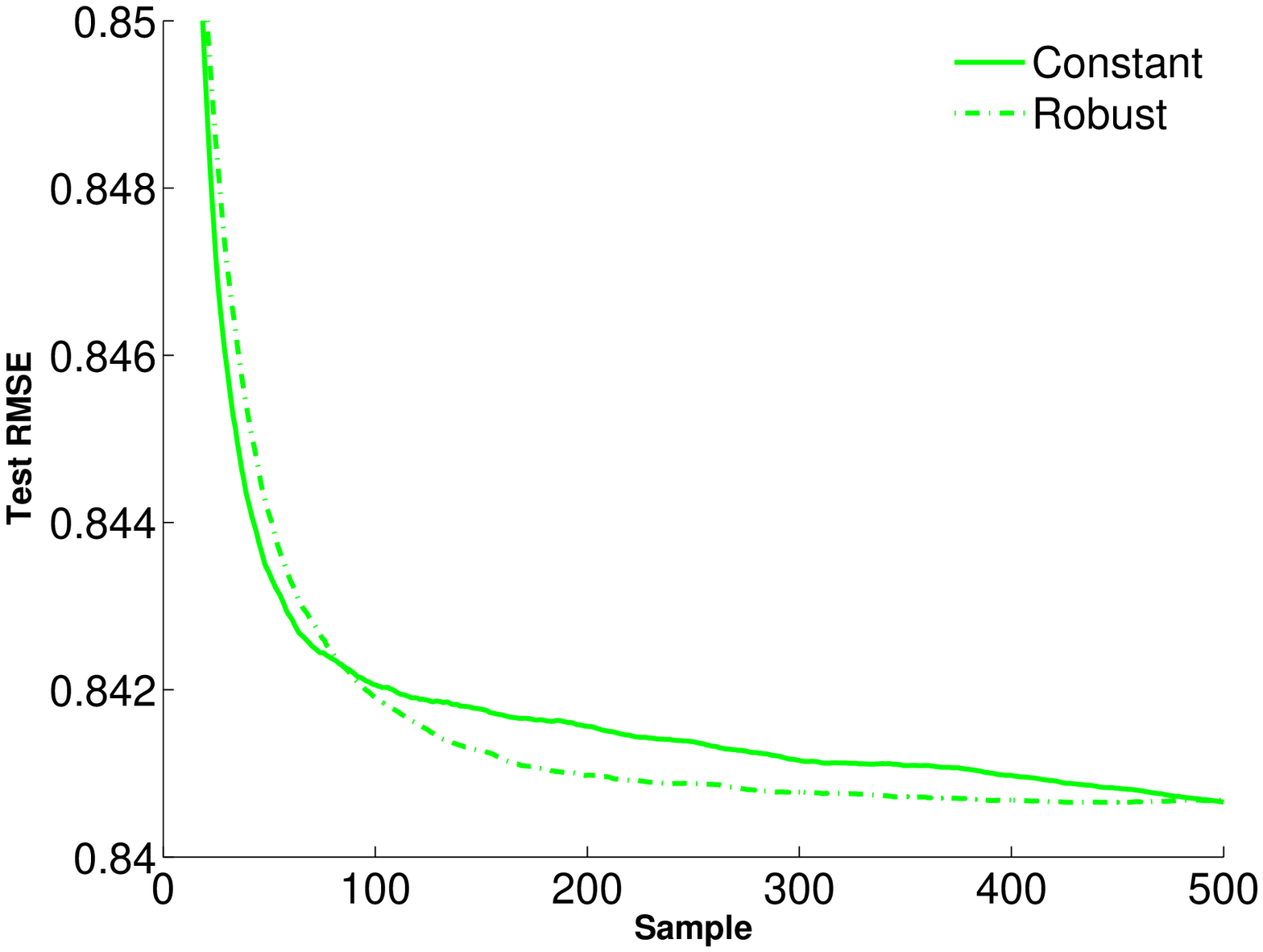}}
\subfloat[][]{\includegraphics[scale=0.4]{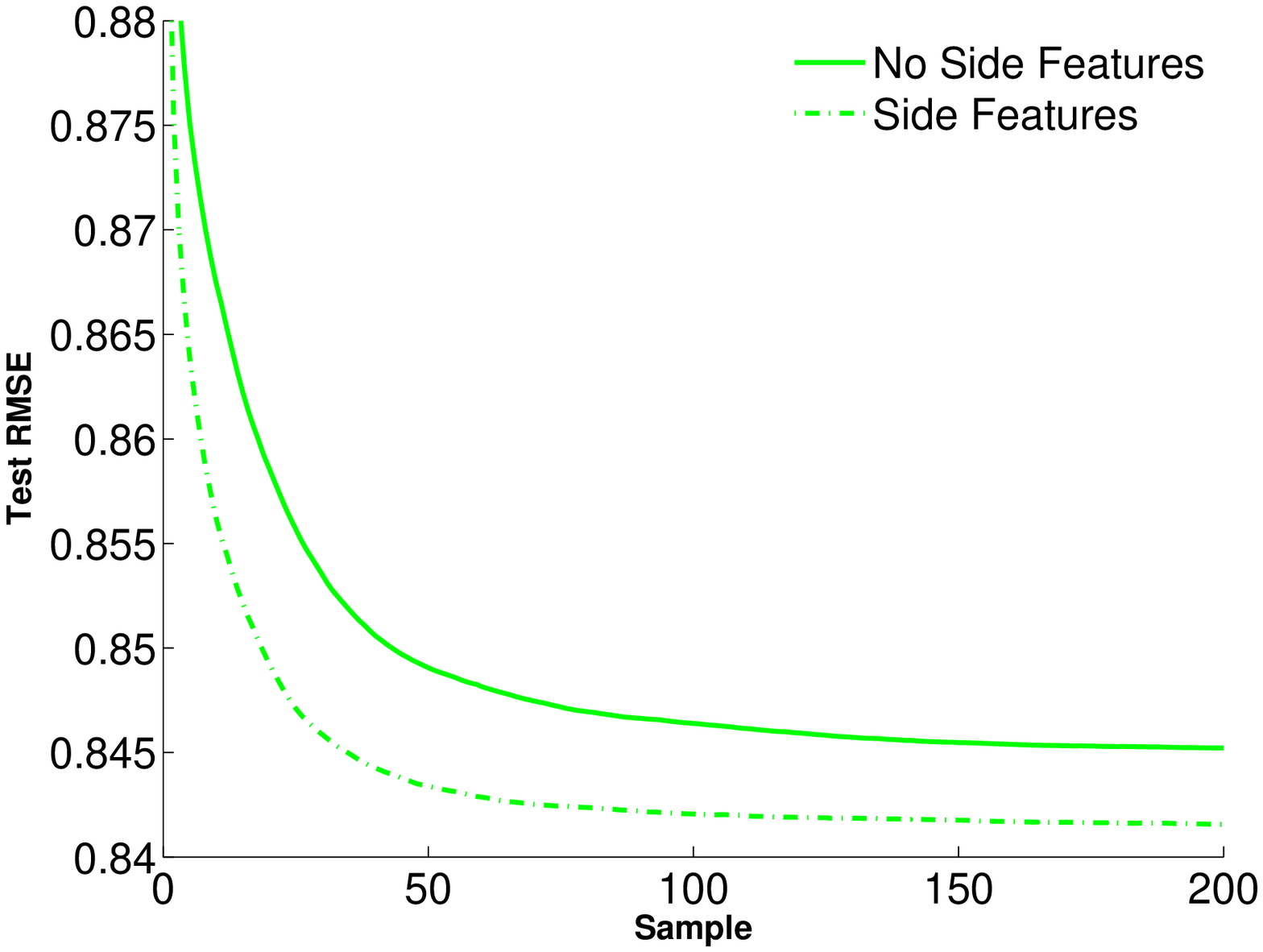}}
\caption{Test RMSE for the Gibbs sampler for the models with (a) side features under the constant and robust precision model, and (b) constant precision with and without side features.}
\label{fig:errorGibbsSide}
\end{figure}

A closer examination of the test error over iterations show subtle differences in how the common converged value is obtained.  In Figure~\ref{fig:errorGibbsSide}, we observe the constant model sees a more substantial drop in the first 50 iterations, after which the incremental gain is minor.  The robust model takes longer to converge, outperforming the constant model after approximately 80 iterations.

Finally, we consider the effect on predictive gain when including the side features in the model.  Table \ref{table:mlGibbsSide}  tabulates the converged prediction error for the sampler in the model with only user and item features (``No Side'') and the model with user, item, and side features (``Side'') with respect to user frequency.  For extremely common users (over 800 ratings), there is no predictive gain.  As expected, there is a predictive gain for the least common users (with number or ratings on the order of 20-30).  However, it is interesting to note that there is still a noticeable gain in test performance for moderately frequent users, those with several hundred ratings.  This gain can be attributed to the side features modelling the correlational influence in the rating structure \cite{mjthesis}.

Comparing to Table \ref{table:userTestFreq}, we see that the largest relative improvement of $0.42\%$ by the precisions for the top $10\%$ of users is comparable to the gains made by the inclusion of side features for the first five bins, corresponding to half of the MovieLens test set.  This highlights the importance of a model to make accurate predictions for rare users.  Significant gains overall may be the result of gains for a small selection of users, as is the case for the precision model. 

\begin{table}[tb]
\begin{center}
\caption{Test RMSE broken down by user frequency under Gibbs sampling for the models with and without side features.}
\label{table:mlGibbsSide}
\begin{tabular}{rr|rrr}
&     & \multicolumn{2}{c}{Error} & Relative\\
& Number of Ratings & No Side & Side & Change (\%)\\
    \hline
(0.01) & $<=25$ & 0.9120 & \textbf{0.9035} & $+0.92$\\
(0.10) & $26-71$ & 0.8674 & \textbf{0.8608} & $+0.76$\\
(0.25) & $72-146$ & 0.8561 & \textbf{0.8494} & $+0.78$\\
(0.30) & $147-171$ & 0.8508 & \textbf{0.8459} & $+0.58$\\
(0.50) & $172-301$ & 0.8193 & \textbf{0.8155} & $+0.46$\\
(0.70) & $302-484$ & 0.8275 & \textbf{0.8243} & $+0.38$\\
(0.90) & $485-829$ & 0.8109 & \textbf{0.8107} & $+0.03$\\
(1) & $830-2,313$ & 0.7475 & \textbf{0.7474} & $+0.01$\\
\hline
\end{tabular}
\end{center}
\end{table}

\section{Conclusion}

This report looked at extensions to matrix factorization models for collaborative filtering.  We provided a comparison between Gibbs sampling and variational inference, noting that variational inference requires precise tuning of the Gaussian-Wishart priors for optimal performance and is also prone to overfitting.  Based on this, we advocate the further user of Monte Carlo methods for prediction in these models.

We further noted that the gain from modelling user and item level precision is not significant when we move to a model class that includes side features.  The same predictive performance is obtained overall.  In addition, there is near-equality in predictive performance within sets of users of different frequency.

The variational algorithm exhibited pathological behaviour with respect to user and item precisions.  In optimizing the variational lower bound, the algorithm drove a subset of precision to arbitrarily small values, and another subset to arbitrarily large values.  Based on this, we investigated if bounding the precisions had any influence on predictive performance.  We replaced the Gamma priors by truncated Gamma priors, and compared the performance of the variational algorithm for different bounds.  In changing the precision bounds monotonically, a non-monotonic change in the performance of the truncated models was observed over the constant model.  It was noted that some bounds do outperform both the constant and the robust precision models.  Further work could investigate automated ways to select the precision bounds.

An analysis of the performance of the Gibbs sampler with respect to user frequency demonstrated that the inclusion of side features offers predictive gains for even moderately common users, those with several hundred ratings.  This highlights the importance of modelling correlational influence in the rating patterns.

It is worth noting, however, that the computation time required for sampling the side features is substantial.  To sample a single side feature $W_k$, it is necessary to consider the subset of the entire rating matrix consisting of all users who rated a given item.  That is, one must consider all users $u_i$ for which $I_{u_i,k} = 1$, and all items each of these users rated.  For globally popular items, this can be a substantial proportion of the original data set.  However, the correlational influence that the side features model is likely to be small when considering such items.

\bibliographystyle{chicago}
\bibliography{bpmfAnalysis_report}

\clearpage

\appendix

\section{Appendix}

In this section, we provide a series of derivations and ancillary results needed to derive the given results.  Primarily, they are needed to obtain the variational lower bound and the conditionals for the variables of interest

\section{Algorithms}

We provide pseudo-code versions of the Gibbs sampler as Algorithm~\ref{algo:gibbs} and the Variational algorithm as Algorithm~\ref{algo:varinf}.  As discussed previously, the primary difference between the two is with respect the to the feature hyper-parameters.  The mean and precision matrix are marginally updated in the Gibbs sampler, but jointly updated in the Variational approximation.

\begin{algorithm}[ht]
\For{Sample $t \geq 1$}{
Sample hyper-parameters\\
$\Lambda_U^{(t)} \sim \mathcal{W}(\Lambda_U | \tilde{\nu}_U, \tilde{W}_U)$ defined by Equation~\eqref{eq:dist:userHyperParam}\\
$(\mu_U^{(t)} | \Lambda_U^{(t)}) \sim \mathcal{N}(\mu_U | \tilde{\mu}_U \tilde{\beta}_U \Lambda_U^{(t)})$ defined by Equation~\eqref{eq:dist:userHyperParam}\\
\hspace*{1em}\\
$\Lambda_V^{(t)} \sim \mathcal{W}(\Lambda_V | \tilde{\nu}_V, \tilde{W}_V)$ defined analogously to Equation~\eqref{eq:dist:userHyperParam}\\
$(\mu_V^{(t)} | \Lambda_V^{(t)}) \sim \mathcal{N}(\mu_V | \tilde{\mu}_V \tilde{\beta}_V \Lambda_V^{(t)})$ defined analogously to Equation~\eqref{eq:dist:userHyperParam}\\
\hspace*{1em}\\
$\Lambda_W^{(t)} \sim \mathcal{W}(\Lambda_W | \tilde{\nu}_W, \tilde{W}_W)$ defined analogously to Equation~\eqref{eq:dist:userHyperParam}\\
$(\mu_W^{(t)} | \Lambda_W^{(t)}) \sim \mathcal{N}(\mu_W | \tilde{\mu}_W \tilde{\beta}_W \Lambda_W^{(t)})$ defined analogously to Equation~\eqref{eq:dist:userHyperParam}\\
\hspace*{1em}\\
  \For{Each user $i=1:N$}{
    Sample user feature $U_i^{(t)} \sim \mathcal{N}(U_i^{(t)} | \mu_{U_i}, \Lambda_{U_i})$ defined by Equation~\eqref{eq:dist:userFeature}\\
    Sample user bias $\gamma_i^{(t)} \sim \mathcal{N}(\gamma_i^{(t)} | \mu_{\gamma_i}, \lambda_{\gamma_i})$ defined by Equation~\eqref{eq:dist:userOffset}\\
    Sample user precision $\alpha_i^{(t)} \sim \mathcal{G}(\alpha_i^{(t)} | a_{U_i}, b_{U_i})$ defined by Equation~\eqref{eq:dist:userPrecision}\\
  }\hspace*{1em}\\
  \For{Each item $j=1:M$}{
    Sample item feature $V_j^{(t)} \sim \mathcal{N}(V_j^{(t)} | \mu_{V_j}, \Lambda_{V_j})$ defined by Equation~\eqref{eq:dist:itemFeature}\\
    Sample item bias $\eta_j^{(t)} \sim \mathcal{N}(\eta_j^{(t)} | \mu_{\eta_j}, \lambda_{\eta_j})$ defined analogously to Equation~\eqref{eq:dist:userOffset}\\
    Sample item precision $\beta_j^{(t)} \sim \mathcal{G}(\beta_j^{(t)} | a_{V_j}, b_{V_j})$ defined analogously to Equation~\eqref{eq:dist:userPrecision}\\
  }\hspace*{1em}\\
  \For{Each item $k=1:M$}{
    Sample side feature $W_k^{(t)} \sim \mathcal{N}(W_k^{(t)} | \mu_{W_k}, \Lambda_{W_k})$ defined by Equation~\eqref{eq:dist:sideFeature}\\
  }\hspace*{1em}\\
  Sample global precision $\tau^{(t)} \sim \mathcal{G}(\tau^{(t)} | a_{\tau}, b_{\tau})$ defined analogously to Equation~\eqref{eq:dist:userPrecision}\\
  Predict test set ratings $\hat{r}^{(t)}_{i,j}$ according to Equation~\eqref{eq:gibbs:predict}\\
}
\caption{Gibbs sampler for Probabilistic Matrix Factorization and the extensions discussed.}
\label{algo:gibbs}
\end{algorithm}

\begin{algorithm}[ht]
\For{Update $t \geq 1$}{
Jointly update hyper-parameters\\
$(\mu_U^{(t)} , \Lambda_U^{(t)}) \sim \mathcal{N}(\mu_U | \tilde{\mu}_U \tilde{\beta}_U \Lambda_U) \cdot \mathcal{W}(\Lambda_U | \tilde{\nu}_U, \tilde{W}_U)$ defined by Equation~\eqref{eq:dist:userHyperParam}\\
\hspace*{1em}\\
$(\mu_V^{(t)} , \Lambda_V^{(t)}) \sim \mathcal{N}(\mu_V | \tilde{\mu}_V \tilde{\beta}_V \Lambda_V) \cdot \mathcal{W}(\Lambda_V | \tilde{\nu}_V, \tilde{W}_V)$ defined analogously to Equation~\eqref{eq:dist:userHyperParam}\\
\hspace*{1em}\\
$(\mu_W^{(t)} , \Lambda_W^{(t)}) \sim \mathcal{N}(\mu_W | \tilde{\mu}_W \tilde{\beta}_W \Lambda_W) \cdot \mathcal{W}(\Lambda_W | \tilde{\nu}_W, \tilde{W}_W)$ defined analogously to Equation~\eqref{eq:dist:userHyperParam}\\
\hspace*{1em}\\
  \For{Each user $i=1:N$}{
    Update user feature $U_i^{(t)} \sim \mathcal{N}(U_i^{(t)} | \mu_{U_i}, \Lambda_{U_i})$ defined by Equation~\eqref{eq:dist:userFeature}\\
    Update user bias $\gamma_i^{(t)} \sim \mathcal{N}(\gamma_i^{(t)} | \mu_{\gamma_i}, \lambda_{\gamma_i})$ defined by Equation~\eqref{eq:dist:userOffset}\\
    Update user precision $\alpha_i^{(t)} \sim \mathcal{G}(\alpha_i^{(t)} | a_{U_i}, b_{U_i})$ defined by Equation~\eqref{eq:dist:userPrecision}\\
  }\hspace*{1em}\\
  \For{Each item $j=1:M$}{
    Update item feature $V_j^{(t)} \sim \mathcal{N}(V_j^{(t)} | \mu_{V_j}, \Lambda_{V_j})$ defined by Equation~\eqref{eq:dist:itemFeature}\\
    Update item bias $\eta_j^{(t)} \sim \mathcal{N}(\eta_j^{(t)} | \mu_{\eta_j}, \lambda_{\eta_j})$ defined analogously to Equation~\eqref{eq:dist:userOffset}\\
    Update item precision $\beta_j^{(t)} \sim \mathcal{G}(\beta_j^{(t)} | a_{V_j}, b_{V_j})$ defined analogously to Equation~\eqref{eq:dist:userPrecision}\\
  }\hspace*{1em}\\
  \For{Each item $k=1:M$}{
    Update side feature $W_k^{(t)} \sim \mathcal{N}(W_k^{(t)} | \mu_{W_k}, \Lambda_{W_k})$ defined by Equation~\eqref{eq:dist:sideFeature}\\
  }\hspace*{1em}\\
  Update global precision $\tau^{(t)} \sim \mathcal{G}(\tau^{(t)} | a_{\tau}, b_{\tau})$ defined analogously to Equation~\eqref{eq:dist:userPrecision}\\
  Predict test set ratings $\hat{r}^{(t)}_{i,j}$ according to Equation~\eqref{eq:varinf:predict}\\
}
\caption{Variational Algorithm for Probabilistic Matrix Factorization and the extensions discussed.}
\label{algo:varinf}
\end{algorithm}

\section{Squared Error Term}

In the derivation of the conditionals of the feature vectors, it was necessary to expand the squared error term $(r_{i,j} - \hat{r}_{i,j})^2$ and rewrite as constants plus a quadratic in terms of $U_i, V_j$, and $W_k$.  We give these three derivations here.  For notational convenience, we suppress the bias terms, absorbing both $\gamma_i$ and $\eta_j$ into $r_{i,j}$.

\subsubsection{Quadratic with Respect to User Features}

In terms of the user feature vectors
\begin{equation}
\begin{aligned}
 & (r_{i,j} - \hat{r}_{i,j})^2 \\
  =& \left[r_{i,j} - \left(\delta_{U} U_i + \frac{\delta_{W}}{n_i}\sum_{k=1}^{M}I_{i,k}W_k\right)^\top V_j\right]^\top 
    \cdot \left[r_{i,j} - \left(\delta_{U} U_i + \frac{\delta_{W}}{n_i}\sum_{k=1}^{M}I_{i,k}W_k\right)^\top V_j \right] \\
  =& \left[\left(r_{i,j} - \frac{\delta_{W}}{n_i}\sum_{k=1}^{M}I_{i,k}W_k^\top V_j\right) - \delta_{U} U_i^\top V_j\right]^\top
    \cdot \left[\left(r_{i,j} - \frac{\delta_{W}}{n_i}\sum_{k=1}^{M}I_{i,k}W_k^\top V_j\right) - \delta_{U} U_i^\top V_j\right] \\
  =& \left[\left(r_{i,j} - \frac{\delta_{W}}{n_i}\sum_{k=1}^{M}I_{i,k}W_k^\top V_j\right)^\top - \delta_{U}V_j^\top U_i\right] 
  \cdot \left[\left(r_{i,j} - \frac{\delta_{W}}{n_i}\sum_{k=1}^{M}I_{i,k}W_k^\top V_j\right) - \delta_{U} U_i^\top V_j\right] \\
  =& \left(r_{i,j} - \frac{\delta_{W}}{n_i}\sum_{k=1}^{M}I_{i,k}W_k^\top V_j\right)^\top \left(r_{i,j} - \frac{\delta_{W}}{n_i}\sum_{k=1}^{M}I_{i,k}W_k^\top V_j\right) \\
  & - 2\delta_{U}\left(r_{i,j} - \frac{\delta_{W}}{n_i}\sum_{k=1}^{M}I_{i,k}W_k^\top V_j\right)^\top V_j^\top U_i
  + \delta_{U}^2U_i^\top V_j V_j^\top U_i
\end{aligned}
\end{equation}

Where the last line follows as $V_j^\top U_i U_i^\top V_j  = U_i^\top  V_jV_j^\top U_i$.

\subsubsection{Quadratic with Respect to Item Features}

In terms of the item feature vectors
\begin{equation}
\begin{aligned}
& (r_{i,j} - \hat{r}_{i,j})^2 \\
  =& \left[r_{i,j} - \left(\delta_{U} U_i + \frac{\delta_{W}}{n_i}\sum_{k=1}^{M}I_{i,k}W_k\right)^\top V_j\right]^\top 
    \cdot \left[r_{i,j} - \left(\delta_{U} U_i + \frac{\delta_{W}}{n_i}\sum_{k=1}^{M}I_{i,k}W_k\right)^\top V_j \right] \\
  =& r_{i,j}^2 - 2r_{i,j}\left(\delta_{U} U_i + \frac{\delta_{W}}{n_i}\sum_{k=1}^{M}I_{i,k}W_k\right)^\top V_j \\
  & + V_j^\top \left(\delta_{U} U_i + \frac{\delta_{W}}{n_i}\sum_{k=1}^{M}I_{i,k}W_k\right)\left(\delta_{U} U_i + \frac{\delta_{W}}{n_i}\sum_{k=1}^{M}I_{i,k}W_k\right)^\top V_j
\end{aligned}
\end{equation}

\subsubsection{Quadratic with Respect to Side Features}

Finally, in terms of the side feature vector $W_k$, we have

\begin{equation}
\begin{aligned}
& (r_{i,j} - \hat{r}_{i,j})^2 \\
  =& \left[r_{i,j} - \left(\delta_{U} U_i + \frac{\delta_{W}}{n_i}\sum_{k=1}^{M}I_{i,k}W_k\right)^\top V_j\right]^\top 
    \cdot \left[r_{i,j} - \left(\delta_{U} U_i + \frac{\delta_{W}}{n_i}\sum_{k=1}^{M}I_{i,k}W_k\right)^\top V_j \right] \\
  =& \left[\left(r_{i,j} - \left(\delta_{U}U_i + \frac{\delta_{W}}{n_i}\sum_{k\neq m}I_{i,k}W_k\right)^\top V_j\right) - \frac{\delta_{W}}{n_i}I_{i,m}W_m^\top V_j\right]^\top \\
  & \cdot \left[\left(r_{i,j} - \left(\delta_{U}U_i + \frac{\delta_{W}}{n_i}\sum_{k\neq m}I_{i,k}W_k\right)^\top V_j\right) - \frac{\delta_{W}}{n_i}I_{i,m}W_m^\top V_j\right] \\
\end{aligned}
\end{equation}

Let $\hat{r}_{i,j,-W_m} = \left(\delta_{U}U_i + \frac{\delta_{W}}{n_i}\sum_{k\neq m}I_{i,k}W_k\right)^\top V_j$ denote the prediction made without $W_m$.  Then,

\begin{equation}
\begin{aligned}
 & (r_{i,j} - \hat{r}_{i,j})^2 \\
  =& \left[\left(r_{i,j} - \hat{r}_{i,j,-W_m}\right) - \frac{\delta_{W}}{n_i}I_{i,m}W_m^\top V_j\right]^\top 
   \cdot \left[\left(r_{i,j} - \hat{r}_{i,j,-W_m}\right) - \frac{\delta_{W}}{n_i}I_{i,m}W_m^\top V_j\right] \\
  =& \left(r_{i,j} - \hat{r}_{i,j,-W_m}\right)^\top\left(r_{i,j} - \hat{r}_{i,j,-W_m}\right) 
  - 2 \frac{\delta_{W}}{n_i}I_{i,m}\left(r_{i,j} - \hat{r}_{i,j,-W_m}\right)V_j^\top W_m \\
  & + \frac{\delta_{W}^2}{n_i^2} W_m^\top V_jV_j^\top W_m
\end{aligned}
\end{equation}

\section{Expectation of Forms}
\subsection{Expectation of Quadratic Forms}
\label{sec:expQuadForm}

Let $x$ be a random vector with mean $\mu$ and covariance matrix $\Sigma$, and let $\Lambda$ be a symmetric matrix.  Then

\begin{equation}
\begin{aligned}
\E[x^\top\Lambda x] =& \tr{\Lambda\Sigma} + \mu^\top\Lambda\mu
\end{aligned}
\end{equation}

Combined with iterated expectation, this is used to find some expectations in the variational lower bound.  An alternative is to expand the quadratic, which we give an example of below using the user feature quadratic form.

\subsection{User Quadratic Form}

In computing the variational lower bound, we need to consider the expectation of quadratic forms such as

\begin{equation}
\begin{aligned}
   & \E_{Q}[(U_i - \mu_U)^\top\Lambda_U(U_i - \mu_U)] \\
  =& \E_{Q}[U_i^\top\Lambda_U U_i] - 2\E_{Q}[U_i^\top \Lambda_U\mu_u] + \E_{Q}[\mu_U^\top\Lambda_U\mu_U]
\end{aligned}
\end{equation}

Which appear from the prior placed on the user, item, and side features.  We compute the expectation term by term.

For the first term

\begin{equation*}
\begin{aligned}
\E_{Q}[U_i^\top\Lambda_U U_i] 
  =& \E_{Q}[ \E_{Q}[U_i^\top\Lambda_U U_i | \Lambda_U] ] \\
  =& \E_{Q}[\tr{\Lambda_{U}\Lambda_{U_i}^{-1}} + \mu_{U_i}^\top\Lambda_U\mu_{U_i}] \\
  =& \tr{\E_{Q}[\Lambda_{U}]\Lambda_{U_i}^{-1}} + \mu_{U_i}^\top\E_{Q}[\Lambda_U]\mu_{U_i} \\
  =& \tilde{\nu}_U\tr{\tilde{W}_U\Lambda_{U_i}^{-1}} + \tilde{\nu}_U\mu_{U_i}^\top\tilde{W}_U\mu_{U_i}
\end{aligned}
\end{equation*}

For the second term
\begin{equation*}
\begin{aligned}
\E_{Q}[U_i^\top\Lambda_U\mu_U]
  =& \E_{Q}[\E_{Q}[ U_i^\top\Lambda_U\mu_U| \Lambda_U, \mu_U]] \\
  =& \E_{Q}[\E_{Q}[U_i]^\top\Lambda_U\mu_U] \\
  =& \mu_{U_i}^\top\E_{Q}[\E_{Q}[\Lambda_U\mu_U |\Lambda_U]] \\
  =& \mu_{U_i}^\top\E_{Q}[\Lambda_U]\tilde{\mu}_U \\
  =& \tilde{\nu}_U\mu_{U_i}^\top\tilde{W}_U\tilde{\mu}_U \\
\end{aligned}
\end{equation*}

For the final term,

\begin{equation*}
\begin{aligned}
\E_{Q}[\mu_U^\top\Lambda_U\mu_U] \\
  =& \E_{Q}[\E_{Q}[\mu_U^\top\Lambda_U\mu_U] | \Lambda_U]]\\
  =& \E_{Q}[\tr{\Lambda_{U}\tilde{\Lambda}_{U}^{-1}} + \tilde{\mu}_{U}^\top\Lambda_U\tilde{\mu}_{U}] \\
  =& \tr{\E_{Q}[\Lambda_{U}]\tilde{\Lambda}_{U}^{-1}} + \tilde{\mu}_{U}^\top\E_{Q}[\Lambda_U]\tilde{\mu}_{U} \\
  =& \tilde{\nu}_U\tr{\tilde{W}_U\tilde{\Lambda}_U^{-1}} + \tilde{\nu}_U\tilde{\mu}_{U}^\top\tilde{W}_U\tilde{\mu}_{U} \\
\end{aligned}
\end{equation*}

Together, the three terms give,

\begin{equation}
\begin{aligned}
& \E_{Q}[(U_i - \mu_U)^\top\Lambda_U(U_i - \mu_U)] \\
  =& \tilde{\nu}_U\tr{\tilde{W}_U\Lambda_{U_i}^{-1}} + \tilde{\nu}_U\mu_{U_i}^\top\tilde{W}_U\mu_{U_i} \\
   & - 2\tilde{\nu}_U\mu_{U_i}^\top\tilde{W}_U\tilde{\mu}_U  \\
   & + \tilde{\nu}_U\tr{\tilde{W}_U\tilde{\Lambda}_U^{-1}} + \tilde{\nu}_U\tilde{\mu}_{U}^\top\tilde{W}_U\tilde{\mu}_{U} \\
  =& \tilde{\nu}_U\left[(\mu_{U_i} - \tilde{\mu}_U)^\top\tilde{W}_U(\mu_{U_i} - \tilde{\mu}_U) + \tr{\tilde{W}_U(\Lambda_{U_i}^{-1} + \tilde{\Lambda}_U^{-1})}\right]
\end{aligned}
\end{equation}

Similar expressions hold for the items and the side features.

\subsection{Gamma Random Variable Expectation}
\label{sec:appendix:ExpGamma}

If $X \sim \mathcal{G}(\alpha, \beta)$, with probability density function $f_{X}(x | \alpha, \beta) \propto x^{\alpha-1} e^{-\beta x}$.

\begin{equation*}
\begin{aligned}
\label{eq:ExpLogGamma}
\E[\log X] =& -\log(\beta) + \psi(\alpha)
\end{aligned}
\end{equation*}

Where $\psi(\cdot) = \frac{d}{d\cdot} \log\Gamma(\cdot)$.  This result important in computing the contribution to the variational lower bound from the user, item, and global precisions.

\subsection{Wishart Random Variable Expectation}

If $\mathbf{X} \sim \mathcal{W}(n,\mathbf{V})$, with probability density function $f_{\mathbf{X}}(\mathbf{X} | n, \mathbf{V}) \propto |\mathbf{X}|^{(n-p-1)/2} e^{-\tr{\mathbf{V}^{-1}\mathbf{X}} / 2}$

\begin{equation*}
\begin{aligned}
\E[\log |X|] =& \sum_{i=1}^{p} \psi\left(\frac{n+1-i}{2}\right) + 2\log2 + \log|\mathbf{V}|
\end{aligned}
\end{equation*}

Like the last result, this is necessary to compute the variational lower bound, as it appears from the conjugate Normal-Wishart priors.

\section{Constrained PMF}
\label{appendix:cpmf}

In this section, we derive the conditional distribution of the features given the observed rating data with the presence of side information.  The inclusion of side information into the model shifts the mean of the user features, and is also re-derived.  The conditional for the item features follows by substituting the combination of user and side features for the user features in the original derivation from \cite{ruslan:icml08}.

For notational convenience, we suppress the offsets, absorbing $\gamma_i$ and $\eta_j$ into $r_{i,j}$.

\subsection{Conditional Posterior for Side Feature}
\label{appendix:cpmf:sideDeriv}

The inclusion of the the side information $W_m$ complicates the log likelihood contribution to the log posterior.  The square in the exponent of the Gaussian for $r_{i,j}$ becomes:

\begin{equation}
\begin{aligned}
 \log p(r_{i,j} | \cdots) =& \sum_{i=1}^{N}\sum_{j=1}^{M}-\frac{I_{i,j}\alpha_i \beta_j\tau}{2} 
    [r_{i,j} - (\delta_{U} U_i + \delta_{W} \frac{1}{n_i}\sum_{k=1}^{m}I_{i,k}W_k)^\top V_j]^\top \\
    &\times[r_{i,j} - (\delta_{U} U_i + \delta_{W} \frac{1}{n_i}\sum_{k=1}^{m}I_{i,k}W_k)^\top V_j],
\end{aligned}
\end{equation}

Where $n_i = \sum_{k=1}^{M}I_{i,k}$.  Using the properties of the transpose and expanding the square yields

\begin{equation}
\begin{aligned}
&\sum_{i=1}^{N}\sum_{j=1}^{M}-\frac{I_{i,j}\alpha_i \beta_j\tau}{2} 
    [r_{i,j} - V_j^\top(\delta_{U} U_i + \delta_{W} \frac{1}{n_i}\sum_{k=1}^{m}I_{i,k}W_k)] \\
    & \times [r_{i,j} - (\delta_{U} U_i + \delta_{W} \frac{1}{n_i}\sum_{k=1}^{m}I_{i,k}W_k)^\top V_j]\\
=&\sum_{i=1}^{N}\sum_{j=1}^{M}-\frac{I_{i,j}\alpha_i \beta_j\tau}{2} 
      [r_{i,j}^2 - 2r_{i,j}V_j^\top(\delta_{U}U_i + \delta_{W}\frac{1}{n_i}\sum_{k=1}^{m}I_{i,k}W_k) \\
    & + V_j^\top(\delta_{U}U_i + \frac{\delta_{W}}{n_i}\sum_{k=1}^{m}I_{i,k}W_k) (\delta_{U}U_i + \frac{\delta_{W}}{n_i}\sum_{k=1}^{m}I_{i,k}W_k)^\top V_j ]
\end{aligned}
\end{equation}

Expanding the quadratic in the final term and dropping terms independent of $W_m$, we obtain,

\begin{equation}
\begin{aligned}
& \sum_{i=1}^{N}\sum_{j=1}^{M}-\frac{I_{i,j}\alpha_i \beta_j\tau}{2} 
    [-2r_{i,j}V_j^\top\delta_{W}\frac{I_{i,m}}{n_i}W_m + 2\delta_{U}\delta_{W}V_j^\top(U_i\frac{I_{i,m}}{n_i}W_m^\top)V_j\\
   & + \delta_{W}^2V_j^\top(\frac{I_{i,m}W_m}{n_i} + \frac{\sum_{k\neq m}^{}I_{i,k}W_k}{n_i})(\frac{I_{i,m}W_m}{n_i} + \frac{\sum_{k\neq m}^{}I_{i,k}W_k}{n_i})^\top)V_j
\end{aligned}
\end{equation}

Note the sum over $W_k$ has been separated into the term involving $W_m$ and the sum over the other $W_k, k \neq m$.

Rearranging vectors to place it in the form $\mu_w^\top\Lambda_w W_m$,

\begin{equation}
\begin{aligned}
& \sum_{i=1}^{N}\sum_{j=1}^{M}-\frac{I_{i,j}\alpha_i \beta_j\tau}{2} 
    [-2\delta_{W}r_{i,j}V_j^\top \frac{I_{i,m}}{n_i}W_m + 2\delta_{U}\delta_{W}U_i^\top(V_j V_j^\top) \frac{I_{i,m}}{n_i}W_m \\
  & + \delta_{W}^2W_m^\top(\frac{I_{i,m}}{n_i})^2V_jV_j^\top W_m + 2\delta_{W}^2\frac{I_{i,m}}{n_i}(\frac{\sum_{k\neq m}I_{i,k}W_k}{n_i})^\top V_jV_j^\top W_m
\end{aligned}
\end{equation}

Adding the log prior $(W_m - \mu_w)^\top\Lambda_w(W_m - \mu_w) / 2$ and grouping terms linear in $W_m$ and quadratic in $W_m$, we obtain the system

\begin{equation}
\begin{aligned}
\Lambda_{W_m} =& \Lambda_w + \delta_{W}^2\tau\sum_{i=1}^{N}\sum_{j=1}^{M}\frac{I_{i,j}I_{i,m}\alpha_i\beta_j}{n_i^2}V_jV_j^\top \\
\mu_{W_m} =& \Lambda_{W_m}^{-1} \bigg[ \Lambda_w\mu_w \\
  & + \tau\sum_{i=1}^{N}\sum_{j=1}^{M}I_{i,j}\alpha_i\beta_j \bigg( r_{i,j}V_j \delta_{W} \frac{I_{i,m}}{n_i} - \delta_{U}\delta_{W}(V_jV_j^\top)U_i \frac{I_{i,m}}{n_i} \\
  & - \delta_{W}^2 V_jV_j^\top\frac{I_{i,m}}{n_i^2}(\sum_{k \neq m}I_{i,k}W_k)\bigg)\bigg]
\end{aligned}
\end{equation}

Rewriting,

\begin{equation}
\begin{aligned}
\mu_{W_m} =& \Lambda_{W_m}^{-1} \bigg[ \Lambda_w\mu_w
    + (1-u)\tau\sum_{\substack{(i,j) : \\I_{i,j}I_{i,m}=1 }} \frac{\alpha_i\beta_j}{n_i}V_j\bigg((r_{i,j} - \delta_{U}V_j^\top U_i) - \delta_{W}V_j^\top\frac{\sum_{k \neq m} I_{i,k}W_k}{n_i}\bigg)\bigg]
\end{aligned}
\end{equation}

This can be re-expressed in a more compact form by defining the prediction made without $W_m$ as

\begin{equation*}
\begin{aligned}
\hat{r}_{i,j,-W_m} = \left(\delta_{U}U_i + \frac{\delta_{W}}{n_i}\sum_{k\neq m}I_{i,k}W_k\right)^\top V_j
\end{aligned}
\end{equation*}

\begin{equation}
\begin{aligned}
\mu_{W_m} =& \Lambda_{W_m}^{-1} \left[ \Lambda_w\mu_w
    + \delta_{W}\tau\sum_{\substack{(i,j) : \\I_{i,j}I_{i,m}=1 }} \frac{\alpha_i\beta_j}{n_i}V_j \left(r_{i,j} - \hat{r}_{i,j,-W_m}\right)) \right]
\end{aligned}
\end{equation}

Which can be interpreted as the inner product between the $j^\text{th}$ feature vector and the error made by all feature vectors with the $m^\text{th}$ side feature omitted.

\subsection{Conditional Posterior for User Feature}
\label{appendix:cpmf:userDeriv}

With the inclusion of side features $W_k$, the log posterior for $U_i$ becomes

\begin{equation}
\begin{aligned}
\label{eq:posterior:userFeature1}
\log p(U_i | \cdots) 
  =& \frac{\tau\alpha_i}{2}\sum_{j=1}^{M}I_{i,j}\beta_j(r_{i,j} - \hat{r}_{i,j})^2 + (U_i - \mu_U)\Lambda_U(U_i - \mu_U)
\end{aligned}
\end{equation}

Expanding the squared term yields,

\begin{equation}
\begin{aligned}
  & (r_{i,j} - \hat{r}_{i,j})^2 \\
  =& \left[r_{i,j} - \left(\delta_{U} U_i + \frac{\delta_{W}}{n_i}\sum_{k=1}^{M}I_{i,k}W_k\right)^\top V_j\right]^\top 
    \cdot \left[r_{i,j} - \left(\delta_{U} U_i + \frac{\delta_{W}}{n_i}\sum_{k=1}^{M}I_{i,k}W_k\right)^\top V_j \right] \\
  =& \left[\left(r_{i,j} - \frac{\delta_{W}}{n_i}\sum_{k=1}^{M}I_{i,k}W_k^\top V_j\right) - \delta_{U} U_i^\top V_j\right]^\top
    \cdot \left[\left(r_{i,j} - \frac{\delta_{W}}{n_i}\sum_{k=1}^{M}I_{i,k}W_k^\top V_j\right) - \delta_{U} U_i^\top V_j\right] \\
  =& \left[\left(r_{i,j} - \frac{\delta_{W}}{n_i}\sum_{k=1}^{M}I_{i,k}W_k^\top V_j\right)^\top - \delta_{U}V_j^\top U_i\right] 
  \cdot \left[\left(r_{i,j} - \frac{\delta_{W}}{n_i}\sum_{k=1}^{M}I_{i,k}W_k^\top V_j\right) - \delta_{U} U_i^\top V_j\right] \\
  =& \left(r_{i,j} - \frac{\delta_{W}}{n_i}\sum_{k=1}^{M}I_{i,k}W_k^\top V_j\right)^\top \left(r_{i,j} - \frac{\delta_{W}}{n_i}\sum_{k=1}^{M}I_{i,k}W_k^\top V_j\right) \\
  & - 2u\left(r_{i,j} - \frac{\delta_{W}}{n_i}\sum_{k=1}^{M}I_{i,k}W_k^\top V_j\right)^\top V_j^\top U_i
  + \delta_{U}^2U_i^\top V_j V_j^\top U_i
\end{aligned}
\end{equation}

Plugging into equation \ref{eq:posterior:userFeature1} and dropping terms not involving $U_i$ yields,

\begin{equation}
\begin{aligned}
\log p(U_i | \cdots) 
  =& \frac{\tau\alpha_i}{2}\sum_{j=1}^{M}I_{i,j}\beta_j\left[- 2u\left(r_{i,j} - \frac{\delta_{W}}{n_i}\sum_{k=1}^{M}I_{i,k}W_k^\top V_j\right)^\top V_j^\top U_i
  + \delta_{U}^2U_i^\top V_j V_j^\top U_i\right] \\
  & + (U_i - \mu_U)\Lambda_U(U_i - \mu_U)
\end{aligned}
\end{equation}

Which shows that the conditional posterior for $U_i$ is Gaussian with parameters
\begin{equation}
\begin{aligned}
\Lambda_{U_i} =& \Lambda_U + \delta_{U}\tau\alpha_i\sum_{j=1}^{M}I_{i,j}\beta_jV_j V_j^\top \\
\mu_{U_i} =& \Lambda_{U_i}^{-1}\left[\Lambda_U\mu_U + \delta_{U}\tau\alpha_i \sum_{j=1}^{M}I_{i,j}\beta_j V_j\left(r_{i,j} - \delta_{W}V_j^\top \left(\frac{\sum_{k=1}^{M}I_{i,k}W_k}{n_i}\right)\right)\right]
\end{aligned}
\end{equation}

Note that the inclusion of side information affects only the mean, not the precision.

\section{Optimal Variational Distributions}
\label{appendix:varDist}

In this section, we derive the optimal variational distribution under the mean field approximation of equation \eqref{eq:meanFieldApprox}.  The subsections are as follows:

\begin{itemize}
\item In Subsection \ref{appendix:varDist:userFeatures}, we derive the optimal variational distribution for the user features;
\item In Subsection \ref{appendix:varDist:itemFeatures}, we derive the optimal variational distribution for the item features;
\item In Subsection \ref{appendix:varDist:sideFeatures}, we derive the optimal variational distribution for the side features;
\item In Subsection \ref{appendix:varDist:precision}, we derive the optimal variational distribution for the user, item, and global precisions;
\item In Subsection \ref{appendix:varDist:hyperparameters}, we derive the optimal variational distribution for the user hyper-parameters.  By symmetry, the results for item and side hyper-parameters follow immediately.
\end{itemize}

\subsection{User Feature Vectors}
\label{appendix:varDist:userFeatures}

For the user feature vectors, the terms involving $U_i$ are the conditional expectation for the rating $r_{i,j}$ and the prior for the feature vector $U_i$.  We then have

\begin{equation}
\begin{aligned}
& \sum_{j=1}^{M} I_{i,j} \log p(r_{i,j} | U_i, V_j ,W_{1:m}, \alpha_i, \beta_j, \tau)
    + \log p(U_i | \mu_U, \Lambda_U) \\
=& -\frac{I_{i,j}}{2}\tau\alpha_i\sum_{j=1}^{M} \beta_j(r_{i,j} - \hat{r}_{i,j})^2 + \frac{1}{2}\log|\Lambda_U| - \frac{1}{2}(U_i - \mu_U)^\top\Lambda_U(U_i - \mu_U) \\
=& -\frac{I_{i,j}}{2}\tau\alpha_i\sum_{j=1}^{M}\beta_j \left(- 2\delta_{U}\left(r_{i,j} - \frac{\delta_{W}}{n_i}\sum_{k=1}^{M}I_{i,k}W_k^\top V_j\right) V_j^\top U_i
  + \delta_{U}U_i^\top V_j V_j^\top U_i\right) \\
  & - \frac{1}{2}(U_i - \mu_U)^\top\Lambda_U(U_i - \mu_U)
\end{aligned}
\end{equation}

Which shows the variational distribution for $U_i$ is Gaussian, with parameters
\begin{equation}
\begin{aligned}
  \mu_{U_i} =& \Lambda_{U_i}^{-1}\left[\Lambda_U\mu_U + \delta_{U}\tau\alpha_i \sum_{j=1}^{M}I_{i,j}\beta_j V_j\left(r_{i,j} - \delta_{W}V_j^\top \left(\frac{\sum_{k=1}^{M}I_{i,k}W_k}{n_i}\right)\right)\right] \\
\Lambda_{U_i} =& \Lambda_U + \delta_{U}\tau\alpha_i\sum_{j=1}^{M}I_{i,j}\beta_jV_j V_j^\top
\end{aligned}
\end{equation}

\subsection{User Offset}

If we include a user offset $\gamma_i$, then the relevant terms are 

\begin{equation}
\begin{aligned}
\label{eq:varDist:userOffset}
   & \sum_{j=1}^{M}I_{i,j} \log p(r_{i,j} | \cdots) + \log p(\gamma_i | \mu_\gamma, \lambda_\gamma) \\
  & \sum_{j=1}^{M}\frac{I_{i,j}}{2}\log|\Lambda_U| - \frac{I_{i,j}}{2}\tau\alpha_i\sum_{j=1}^{M}(r_{i,j} - \hat{r}_{i,j})^2 + \frac{1}{2}\log\lambda_\gamma - \frac{\lambda_\gamma}{2}(\gamma_i - \mu_\gamma)^2
\end{aligned}
\end{equation}

The quadratic $(r_{i,j} - \hat{r}_{i,j})^2$ can be rewritten as

\begin{equation*}
\begin{aligned}
(r_{i,j} - \hat{r}_{i,j})^2 
  =& (r_{i,j} - \gamma_i - \eta_j - S_i^\top V_j)^2 \\
  =& \gamma_i^2 - 2\gamma_i(r_{i,j} - \eta_j - S_i^\top V_j) + (r_{i,j} - \eta_j - S_i^\top V_j)^2
\end{aligned}
\end{equation*}

Inserting into Equation \ref{eq:varDist:userOffset}, taking expectations, and retaining terms involving $\gamma_i$ only, we get that the optimal distribution for $\gamma_i$ is univariate Gaussian, with parameters

\begin{equation}
\begin{aligned}
\tilde{\lambda}_{\gamma_{i}} =& \tau\alpha_i\sum_{j=1}^{M} I_{i,j}\beta_j + \lambda_{\gamma} \\
\tilde{\mu}_{\gamma_{i}} =& \tilde{\lambda}_{\gamma_{i}}^{-1} \left(\lambda_\gamma\mu_\gamma + \tau\alpha_i\sum_{j=1}^{M}I_{i,j}\beta_j (r_{i,j} - \eta_j - S_i^\top V_j) \right)
\end{aligned}
\end{equation}

\subsection{Item Feature Vectors}
\label{appendix:varDist:itemFeatures}

By symmetry, the terms involving $V_j$ are

\begin{equation}
\begin{aligned}
& \sum_{i=1}^{N} I_{i,j} \log p(r_{i,j} | U_i, V_j ,W_{1:m}, \alpha_i, \beta_j, \tau)
    + \log p(V_j | \mu_V, \Lambda_V) \\
=& -\frac{I_{i,j}}{2}\beta_j\tau\sum_{i=1}^{N} \alpha_i(r_{i,j} - \hat{r}_{i,j})^2 + \frac{1}{2}\log|\Lambda_V| - \frac{1}{2}(V_j - \mu_V)^\top\Lambda_V(V_j - \mu_V) \\
=& -\frac{I_{i,j}}{2}\beta_j\tau\sum_{i=1}^{N} \alpha_i\left(- 2r_{i,j}\left(\delta_{U} U_i + \frac{\delta_{W}}{n_i}\sum_{k=1}^{M}I_{i,k}W_k\right)^\top V_j\right) \\
  & + V_j^\top \left(\delta_{U} U_i + \frac{\delta_{W}}{n_i}\sum_{k=1}^{M}I_{i,k}W_k\right)\left(\delta_{U} U_i + \frac{\delta_{W}}{n_i}\sum_{k=1}^{M}I_{i,k}W_k\right)^\top V_j \\
  & - \frac{1}{2}(V_j - \mu_V)^\top\Lambda_V(V_j - \mu_V) \\
\end{aligned}
\end{equation}

Which shows the variational distribution for $V_j$ is Gaussian, with parameters

\begin{equation}
\begin{aligned}
  \mu_{V_j} =& \Lambda_{V_j}^{-1}\left[\Lambda_V\mu_V + \tau\beta_j \sum_{i=1}^{N}I_{i,j}\alpha_i (r_{i,j} - (\delta_{U}U_i + \frac{\delta_{W}}{n_i}\sum_{k=1}^{M}I_{i,k}W_k)^\top V_j)\right] \\
  \Lambda_{V_j} =& \Lambda_V + \tau\beta_j\sum_{i=1}^{N}I_{i,j}\alpha_i (\delta_{U}U_i + \frac{\delta_{W}}{n_i}\sum_{k=1}^{M}I_{i,k}W_k) (\delta_{U}U_i + \frac{\delta_{W}}{n_i}\sum_{k=1}^{M}I_{i,k}W_k)^\top
\end{aligned}
\end{equation}

\subsection{Side Feature Vectors}
\label{appendix:varDist:sideFeatures}

The terms involving $W_m$ are

\begin{equation}
\begin{aligned}
& \sum_{i=1}^{N}\sum_{j=1}^{M} I_{i,j} \log p(r_{i,j} | U_i, V_j ,W_{1:m}, \alpha_i, \beta_j, \tau)
    + \log p(W_m | \mu_W, \Lambda_W) \\
  =& \frac{\tau}{2}\sum_{i=1}^{N}\sum_{j=1}^{M} I_{i,j} \alpha_i\beta_j(r_{i,j} - \hat{r}_{i,j})^2 - \frac{1}{2}(W_m - \mu_W)^\top\Lambda_W(W_m - \mu_W) \\
  =& -\frac{\tau}{2}\sum_{i=1}^{N}\sum_{j=1}^{M}I_{i,j}\alpha_i\beta_j \left(- 2 \frac{\delta_{W}}{n_i}I_{i,m}(r_{i,j} - \hat{r}_{i,j,-W_m})V_j^\top W_m + \frac{\delta_{W}}{n_i^2} W_m^\top V_jV_j^\top W_m\right) \\
  & - \frac{1}{2}(W_m - \mu_W)^\top\Lambda_W(W_m - \mu_W)
\end{aligned}
\end{equation}

Where $\hat{r}_{i,j,-W_m} = \left(\delta_{U}U_i + \frac{\delta_{W}}{n_i}\sum_{k\neq m}I_{i,k}W_k\right)^\top V_j$ denotes the prediction made without $W_m$.  This shows the variational distribution for $W_m$ is Gaussian with parameters

\begin{equation}
\begin{aligned}
\mu_{W_m} =& \Lambda_{W_m}^{-1} \left[ \Lambda_w\mu_w
    + \delta_{W}\tau\sum_{\substack{(i,j) : \\I_{i,j}I_{i,m}=1 }} \frac{\alpha_i\beta_j}{n_i}V_j \left(r_{i,j} - \hat{r}_{i,j,-W_m}\right)\right]\\
\Lambda_{W_m} =& \Lambda_W + \delta_{W}\tau\sum_{\substack{(i,j) : \\I_{i,j}I_{i,m}=1 }}\frac{\alpha_i\beta_j}{n_i^2}V_jV_j^\top \\
\end{aligned}
\end{equation}

Note the product of the two indicators $I_{i,j}I_{i,m}$.  The side information sum only considers those users who rated this item, and then considers those items these users rated.

\subsection{Precisions}
\label{appendix:varDist:precision}

The terms involving $\alpha_i$ are

\begin{equation}
\begin{aligned}
& \sum_{j=1}^{M} I_{i,j} \log p(r_{i,j} | U_i, V_j ,W_{1:m}, \alpha_i, \beta_j, \tau)
    + \log p(\alpha_i | a_U, b_U) \\ 
  =& \sum_{j=1}^{M}\frac{I_{i,j}}{2}\log\alpha_i - \frac{\tau}{2}\sum_{j=1}^{M} I_{i,j} \beta_j(r_{i,j} - \hat{r}_{i,j})^2 + (a_U-1)\log\alpha_i - b_U\alpha_i
\end{aligned}
\end{equation}

Which shows the variational distribution for $\alpha_i$ is Gamma, with parameters

\begin{equation}
\begin{aligned}
\tilde{a}_{U_i} =& a_U + \frac{1}{2}\sum_{j=1}^{M} I_{i,j} \\
\tilde{b}_{U_i} =& b_U + \frac{\tau}{2}\sum_{j=1}^{M} I_{i,j} \beta_j(r_{i,j} - \hat{r}_{i,j})^2
\end{aligned}
\end{equation}

Identical derivations show the variational distributions for $\beta_j$ and $\tau$ are Gamma, with parameters

\begin{equation}
\begin{aligned}
\tilde{a}_{V_j} =& a_V + \frac{1}{2}\sum_{i=1}^{N} I_{i,j} \\
\tilde{b}_{V_j} =& b_V + \frac{\tau}{2}\sum_{i=1}^{N} I_{i,j} \alpha_i(r_{i,j} - \hat{r}_{i,j})^2 \\
 & \\
\tilde{a}_\tau =& a_\tau + \frac{1}{2}\sum_{i=1}^{N}\sum_{j=1}^{M} I_{i,j} \\
\tilde{b}_\tau =& b_\tau + \frac{1}{2}\sum_{i=1}^{N}\sum_{j=1}^{M} I_{i,j} \alpha_i\beta_j(r_{i,j} - \hat{r}_{i,j})^2 \\
\end{aligned}
\end{equation}

\subsection{User / Item / Side Feature Hyper-parameters}
\label{appendix:varDist:hyperparameters}

The terms involving the user hyper-parameters $\mu_U, \Lambda_U$ are

\begin{equation}
\begin{aligned}
& \sum_{i=1}^{N} \log p(U_i | \mu_U, \Lambda_U) + \log p(\mu_U | \mu_0, \beta_0\Lambda_U) + \log(\Lambda_U | \nu_0, W_0) \\
  =& \frac{N}{2} \log|\Lambda_U| - \frac{1}{2}\sum_{i=1}^{N}(U_i - \mu_U)^\top\Lambda_U(U_i - \mu_U) \\
  & + \frac{1}{2}\log\Lambda_U - \frac{\beta_0}{2}(\mu_U - \mu_0)^\top\Lambda_U(\mu_U - \mu_0) \\
  & + \frac{\nu_0 - d - 1}{2}\log|\Lambda_U| - \frac{1}{2}\tr{W_0^{-1}\Lambda_U} \\
\end{aligned}
\end{equation}

Using derivations involving the completion of the square, see ex: \cite{fraley200501}, the quadratic terms can be rearranged as

\begin{equation}
\begin{aligned}
 & \sum_{i=1}^{N}(U_i - \mu_U)^\top \Lambda_U (U_i - \mu_U) + \beta_0(\mu_U - \mu_0)^\top \Lambda_U(\mu_U - \mu_0) \\
  =& \tr{\left[\sum_{i=1}^{N}(U_i - \mu_U)(U_i - \mu_U)^\top + \beta_0(\mu_U - \mu_0)(\mu_U - \mu_0)^\top \right]\Lambda_U} \\
  =& \tr{\left[N(\overline{U} - \mu_U)(\overline{U} - \mu_U)^\top + \sum_{i=1}^{N}(U_i - \overline{U})(U_i - \overline{U})^\top + \beta_0(\mu_U - \mu_0)(\mu_U - \mu_0)\right]\Lambda_U} \\
  =& \tr{\left[(N+\beta_0)(\mu_U - \tilde{\mu}_U)(\mu_U - \tilde{\mu}_U)^\top + \frac{N\beta_0}{N + \beta_0} (\overline{U} - \mu_0)(\overline{U} - \mu_0)^\top + \sum_{i=1}^{N}(U_i - \overline{U})(U_i - \overline{U})^\top\right]\Lambda_U}
\end{aligned}
\end{equation}

Where we have defined 

\begin{equation}
\begin{aligned}
\tilde{\mu}_U =& \frac{N\overline{U} + \beta_0\mu_0}{N + \beta_0} \\
\end{aligned}
\end{equation}

We can now write the $\mu_U, \Lambda_U$ terms as 

\begin{equation}
\begin{aligned}
& \sum_{i=1}^{N} \log p(U_i | \mu_U, \Lambda_U) + \log p(\mu_U | \mu_0, \beta_0\Lambda_U) + \log(\Lambda_U | \nu_0, W_0) \\
  =& \frac{1}{2}\log|\Lambda_U| - \frac{1}{2}(\mu_U - \tilde{\mu}_U)^\top[(N+\beta_0)\Lambda_U](\mu_U - \tilde{\mu}_U) \\
  & + \frac{N + \nu_0 - d - 1}{2}\log|\Lambda_U| \\
  & - \frac{1}{2}\tr{W_0^{-1} + \left[\frac{N\beta_0}{N + \beta_0}(\overline{U} - \mu_0)(\overline{U} - \mu_0)^\top + \sum_{i=1}^{N}(U_i - \overline{U})(U_i - \overline{U})^\top\right]\Lambda_U}
\end{aligned}
\end{equation}

Which shows the variational distribution for $(\mu_U, \Lambda_U)$ is a Normal-Wishart with parameters

\begin{equation}
\begin{aligned}
\label{eq:userhyperparam}
\tilde{\mu}_U =& \frac{N\overline{U} + \beta_0\mu_0}{N + \beta_0} \\
\tilde{\Lambda}_U =& (N + \beta_0)\Lambda_U \\
\tilde{\nu}_U =& N + \nu_0 \\
\tilde{W}_U^{-1} =& W_0^{-1} + \frac{N\beta_0}{N + \beta_0}(\overline{U} - \mu_0)(\overline{U} - \mu_0)^\top + \sum_{i=1}^{N}(U_i - \overline{U})(U_i - \overline{U})^\top
\end{aligned}
\end{equation}

Analogous statements (with the appropriate sample size, feature averages, etc.) hold for the item and side feature vectors.

\section{Gibbs Distributions}
\label{appendix:gibbs}

Many of the conditional posteriors for all the parameters of interest match those derived for variational inference by the independence assumptions.  The exception is for the feature hyper-parameters, as the variational approximation was a joint distribution over these.  Here, we derive the conditional posterior for the user feature hyper-parameters.  The others follow by symmetry.

The log posterior distribution for $(\mu_U, \Lambda_U)$ has been shown to the take the form of a Gaussian Wishart Distribution, with parameters given in Equation \ref{eq:userhyperparam}.  The conditional for $\Lambda_U$ takes the form

\begin{equation}
\begin{aligned}
\log p(\Lambda_U| \mu_U, \cdots) 
  =& +\frac{1}{2}\log |\Lambda_U| + \frac{\tilde{\nu}_U - d - 1}{2}\log|\Lambda_U| - \frac{1}{2}\tr{\tilde{W}_U^{-1}\Lambda_U} \\
  =& \frac{\tilde{\nu} - d}{2}\log|\Lambda_U| - \frac{1}{2}\tr{\tilde{W}_U^{-1}\Lambda_U}
\end{aligned}
\end{equation}

So the Gibbs distribution for $\Lambda_U$ is Wishart with $\tilde{\nu} + 1 = \nu_0 + N + 1$ degrees of freedom.  The scale matrix is unchanged.

For the mean $\mu_U$, it follows immediately that,

\begin{equation}
\begin{aligned}
\log p(\mu_U | \Lambda_U, \cdots) 
  =& -\frac{1}{2}(\mu_U - \tilde{\mu}_U)^\top [(N + \beta_0)\Lambda_U] (\mu_U - \tilde{\mu}_U)
\end{aligned}
\end{equation}

So the Gibbs distribution for $\mu_U$ is the Gaussian with mean $\tilde{\mu}_U$ and precision $(N+\beta_0)\Lambda_U$.

\section{Summary of Derived Distribution}
\label{appendix:summary}

The tables below summarize the distributions for the 

\newsavebox\userFeatureParam
\begin{lrbox}{\userFeatureParam}
\begin{minipage}{\textwidth}
\begin{equation}
\begin{aligned}
\label{eq:dist:userFeature}
Q(U_i) \sim& \mathcal{N}(U_i | \mu_{U_i}, \Lambda_{U_i}) \\
  \mu_{U_i} =& \Lambda_{U_i}^{-1}\bigg[\Lambda_U\mu_U  + \delta_{U}\tau\alpha_i \sum_{j=1}^{M}I_{i,j}\beta_j V_j\left(r_{i,j} - \frac{\delta_{W}}{n_i}V_j^\top \left(\sum_{k=1}^{M}I_{i,k}W_k\right)\right)\bigg] \\
\Lambda_{U_i} =& \Lambda_U + \delta_{U}\tau\alpha_i\sum_{j=1}^{M}I_{i,j}\beta_jV_j V_j^\top \\
\end{aligned}
\end{equation}
\text{$Q(U_i)$ for the variational algorithm agrees with $p(U_i | \cdots)$ for the Gibbs sampler}\\
\end{minipage}
\end{lrbox}

\newsavebox\itemFeatureParam
\begin{lrbox}{\itemFeatureParam}
\begin{minipage}{\textwidth}
\begin{equation}
\begin{aligned}
\label{eq:dist:itemFeature}
Q(V_j) \sim& \mathcal{N}(V_j | \mu_{V_j}, \Lambda_{V_j}) \\
  \mu_{V_j} =& \Lambda_{V_j}^{-1}\left[\Lambda_V\mu_V + \tau\beta_j \sum_{i=1}^{N}I_{i,j}\alpha_i r_{i,j}S_i\right] \\
  \Lambda_{V_j} =& \Lambda_V + \tau\beta_j\sum_{i=1}^{N}I_{i,j}\alpha_i S_iS_i^\top\\
  S_i =& \delta_{U}U_i + \frac{\delta_{W}}{n_i}\sum_{k=1}^{M}I_{i,k}W_k \\
  n_i =& \sum_{j=1}^{M} I_{i,j}
\end{aligned}
\end{equation}
\text{$Q(V_j)$ for the variational algorithm agrees with $p(V_j | \cdots)$ for the Gibbs sampler}\\
\end{minipage}
\end{lrbox}

\newsavebox\sideFeatureParam
\begin{lrbox}{\sideFeatureParam}
\begin{minipage}{\textwidth}
\begin{equation}
\begin{aligned}
\label{eq:dist:sideFeature}
Q(W_k) \sim& \mathcal{N}(W_k | \mu_{W_k}, \Lambda_{W_k})\\
\mu_{W_k} =& \Lambda_{W_k}^{-1} \bigg[ \Lambda_w\mu_w \\
     & + \delta_{W}\tau\sum_{\substack{(i,j) : \\I_{i,j}I_{i,m}=1 }} \frac{\alpha_i\beta_j}{n_i}V_j \left(r_{i,j} - \hat{r}_{i,j,-W_k}\right)\bigg]\\
\Lambda_{W_k} =& \Lambda_W + \delta_{W}\tau\sum_{\substack{(i,j) : \\I_{i,j}I_{i,k}=1 }}\frac{\alpha_i\beta_j}{n_i^2}V_jV_j^\top \\
\end{aligned}
\end{equation}
\text{$Q(W_k)$ for the variational algorithm agrees with $p(W_k | \cdots)$ for the Gibbs sampler}\\
\end{minipage}
\end{lrbox}

\newsavebox\userPrecision
\begin{lrbox}{\userPrecision}
\begin{minipage}{\textwidth}
\begin{equation}
\begin{aligned}
\label{eq:dist:userPrecision}
Q(\alpha_i) \sim& \mathcal{G}(\alpha_i | \tilde{a}_{U_i}, \tilde{b}_{U_i}) \\
\tilde{a}_{U_i} =& a_U + \frac{1}{2}\sum_{j=1}^{M} I_{i,j} \\
\tilde{b}_{U_i} =& b_U + \frac{\tau}{2}\sum_{j=1}^{M} I_{i,j} \beta_j(r_{i,j} - \hat{r}_{i,j})^2
\end{aligned}
\end{equation}
\text{$Q(\alpha_i)$ for the variational algorithm agrees with $p(\alpha_i | \cdots)$ for the Gibbs sampler}\\
\end{minipage}
\end{lrbox}

\newsavebox\itemPrecision
\begin{lrbox}{\itemPrecision}
\begin{minipage}{\textwidth}
\begin{equation}
\begin{aligned}
\label{eq:dist:itemPrecision}
Q(\beta_j) \sim& \mathcal{G}(\beta_j | \tilde{a}_{V_j}, \tilde{b}_{V_j}) \\
\tilde{a}_{V_j} =& a_V + \frac{1}{2}\sum_{i=1}^{N} I_{i,j} \\
\tilde{b}_{V_j} =& b_V + \frac{\tau}{2}\sum_{i=1}^{N} I_{i,j} \alpha_i(r_{i,j} - \hat{r}_{i,j})^2
\end{aligned}
\end{equation}
\end{minipage}
\text{$Q(\beta_j)$ for the variational algorithm agrees with $p(\beta_j | \cdots)$ for the Gibbs sampler}\\
\end{lrbox}

\newsavebox\globalPrecision
\begin{lrbox}{\globalPrecision}
\begin{minipage}{\textwidth}
\begin{equation}
\begin{aligned}
\label{eq:dist:globalPrecision}
Q(\tau) \sim& \mathcal{G}(\tau | \tilde{a}_{\tau}, \tilde{b}_{\tau}) \\
\tilde{a}_\tau =& a_\tau + \frac{1}{2}\sum_{i=1}^{N}\sum_{j=1}^{M} I_{i,j} \\
\tilde{b}_\tau =& b_\tau + \frac{\tau}{2}\sum_{i=1}^{N}\sum_{j=1}^{M} I_{i,j} \alpha_i\beta_j(r_{i,j} - \hat{r}_{i,j})^2
\end{aligned}
\end{equation}
\text{$Q(\tau)$ for the variational algorithm agrees with $p(\tau | \cdots)$ for the Gibbs sampler}\\
\end{minipage}
\end{lrbox}

\newsavebox\userOffset
\begin{lrbox}{\userOffset}
\begin{minipage}{\textwidth}
\begin{equation}
\begin{aligned}
\label{eq:dist:userOffset}
(\gamma_i) \sim& \mathcal{N}(\gamma_i| \tilde{\mu}_{\gamma_{i}}, \tilde{\lambda}_{\gamma_{i}})\\
\tilde{\lambda}_{\gamma_{i}} =& \tau\alpha_i\sum_{j=1}^{M} I_{i,j}\beta_j + \lambda_{\gamma} \\
\tilde{\mu}_{\gamma_{i}} =& \tilde{\lambda}_{\gamma_{i}}^{-1} \left(\lambda_\gamma\mu_\gamma + \tau\alpha_i\sum_{j=1}^{M}I_{i,j}\beta_j (r_{i,j} - \eta_j - S_i^\top V_j) \right)
\end{aligned}
\end{equation}
\text{$Q(\gamma_i)$ for the variational algorithm agrees with $p(\gamma_i | \cdots)$ for the Gibbs sampler}\\
\end{minipage}
\end{lrbox}

\newsavebox\userHyperParam
\begin{lrbox}{\userHyperParam}
\begin{minipage}{\textwidth}
\begin{equation}
\begin{aligned}
\label{eq:dist:userHyperParam}
Q(\mu_U,\Lambda_U) \sim& \mathcal{N}(\mu_U | \tilde{\mu}_U, \tilde{\Lambda}_U) \times \mathcal{W}(\Lambda_U | \tilde{\nu}_U, \tilde{W}_U) \\
\vspace{1em}
\\
\vspace{1em}
&\text{For the Gibbs sampler, the conditional posteriors take the form} \\
\Lambda_U \sim& \mathcal{W}(\Lambda_U | \tilde{\nu}_U, \tilde{W}_U)\\
(\mu_U | \Lambda_U) \sim& \mathcal{N}(\mu_U | \tilde{\mu}_U, (N + \beta_0)\Lambda_U)
\\
\tilde{\mu}_U =& \frac{N\overline{U} + \beta_0\mu_0}{N + \beta_0}\\
\overline{U} =& \frac{1}{N}\sum_{i=1}^{N}U_i \\
\tilde{\Lambda}_U =& (N + \beta_0)\Lambda_U, \\  
\tilde{\nu}_U =& N + \nu_0 \\
\tilde{W}_U^{-1} =& W_0^{-1} + \frac{N\beta_0}{N + \beta_0}(\overline{U} - \mu_0)(\overline{U} - \mu_0)^\top + \sum_{i=1}^{N}(U_i - \overline{U})(U_i - \overline{U})^\top
\end{aligned}
\end{equation}
\end{minipage}
\end{lrbox}

\begin{table}[htbp]
\begin{center}
\caption{Form of the variational distributions for the feature vectors.}
\label{table:dist:summary1}
\begin{tabular}{c}
\hline
 \usebox{\userFeatureParam} \\
\hline
\usebox{\itemFeatureParam} \\
\hline
\usebox{\sideFeatureParam} \\
\end{tabular}
\end{center}
\end{table}

\begin{table}[htbp]
\begin{center}
\caption{Form of the variational distributions for the biases, precisions, and hyper-parameters.  Note these agree completely with the Gibbs samplers, except as noted for the hyper-parameters.}
\label{table:dist:summary2}
\begin{tabular}{c}
\hline
\usebox{\userPrecision} \\
\hline
\usebox{\userOffset} \\
\hline
\usebox{\userHyperParam} \\
\hline
\end{tabular}
\end{center}
\end{table}

\section{Variational Lower Bound}
\label{appendix:varLowerBound}

In this section, we derive the variational lower bound for scaled  BPMF.  From Section \ref{sec:varinf:intro}, the lower bound takes the form

\begin{equation}
\begin{aligned}
\label{eq:vlb} 
    & \int Q(\theta) \log \frac{P(\theta,R)}{Q(\theta)} \df{\theta} \\
   =& \E_{Q}[\log p(\theta,R)] - \E_{Q}[\log Q(\theta)]
\end{aligned}
\end{equation}

the expected complete log likelihood of the model and entropy of the variational distribution.

In Section \ref{appendix:varLowerBound:completeLL}, we derive the expected complete log likelihood term.  The entropy is done in Section \ref{appendix:varLowerBound:entropy}.

\subsection{Expected Complete Log Likelihood}
\label{appendix:varLowerBound:completeLL}

The expected complete log likelihood is:

\begin{equation}
\begin{aligned}
\label{eq:vlb:compll}
 & \E_{Q}\left[\log P(R,\theta)\right] \\
  =& \sum_{i=1}^{N}\sum_{j=1}^{M}\E_{Q}\left[I_{i,j} \log p(r_{i,j} | r_{-(i,j)}, U_i, V_j, W_{1:m}, \alpha_i, \beta_j, \tau\right] \\
  & + \sum_{i=1}^{N} \E_{Q}\left[\log p(U_i | \mu_U, \Lambda_U) \right] 
   + \sum_{i=1}^{N} \E_{Q}\left[\log p(\alpha_i | a_U, b_U) \right] 
   + \sum_{i=1}^{N} \E_{Q}\left[\log p(\gamma_i | \mu_\gamma, \lambda_\gamma) \right]\\
  & + \sum_{j=1}^{M} \E_{Q}\left[\log p(V_j | \mu_V, \Lambda_V) \right] 
   + \sum_{j=1}^{M} \E_{Q}\left[\log p(\beta_j | a_V, b_V) \right]
   + \sum_{j=1}^{M} \E_{Q}\left[\log p(\eta_j | \mu_\eta, \lambda_\eta) \right]\\
  & + \sum_{k=1}^{M} \E_{Q}\left[\log p(W_k | \mu_W, \Lambda_W) \right] \\
  & + \E_{Q}\left[\log p(\tau | a_\tau, b_\tau)\right]\\
  & + \E_{Q}\left[\log p(\mu_U | \mu_0, \beta_0\Lambda_U )\right] + \E_{Q}\left[\log p(\Lambda_U | W_0, \nu_0)\right] \\
  & + \E_{Q}\left[\log p(\mu_V | \mu_0, \beta_0\Lambda_V )\right] + \E_{Q}\left[\log p(\Lambda_V | W_0, \nu_0)\right] \\
  & + \E_{Q}\left[\log p(\mu_W | \mu_0, \beta_0\Lambda_W )\right] + \E_{Q}\left[\log p(\Lambda_W | W_0, \nu_0)\right] \\
\end{aligned}
\end{equation}

Where again, the expectations are with respect to the unknown latent feature vectors, precision, and hyper-parameters.

The outline is as follows:

\begin{itemize}
\item In Section \ref{appendix:varLowerBound:completeLL:rating}, we derive the contribution from the conditional density for the rating, $\E_{Q}\left[I_{i,j} \log p(r_{i,j} | U_i, V_j, W_{1:m}, \alpha_i, \beta_j, \tau\right]$.  The final expression for this factor is Equation \eqref{eq:varLowerBound:completeLL:rating:exression};
\item In Section \ref{appendix:varLowerBound:completeLL:features}, we derive the contribution from the user feature vector, $\E_{Q}\left[\log p(U_i | \mu_U, \Lambda_U) \right]$.  The final expression for this factor is Equation \eqref{eq:varLowerBound:completeLL:userFeatures:expression};
\item In Section \ref{appendix:varLowerBound:completeLL:precision}, we derive the contribution from the user precision, $\E_{Q}\left[\log p(\alpha_i | a_U, b_U)\right]$.  The final expression for this factor is Equation \eqref{eq:varLowerBound:completeLL:userPrec:expression};
\item In Section \ref{appendix:varLowerBound:completeLL:userBias}, we derive the contribution from the user bias, $\E_{Q}\left[\log p(\gamma_i | \mu_\gamma, \lambda_\gamma)\right]$.  The final expression for this factor is Equation \eqref{eq:varLowerBound:completeLL:userBias:expression};
\item In Section \ref{appendix:varLowerBound:completeLL:hyperparameters}, we derive the contribution from the user feature hyper-parameters, \\$\E_{Q}\left[\log p(\mu_U | \mu_0, \beta_0\Lambda_U )\right] + \E_{Q}\left[\log p(\Lambda_U | W_0, \nu_0)\right]$.  The final expression is for this factor is Equation \eqref{eq:varLowerBound:completeLL:userHyper:expression}.
\end{itemize}

The item and side feature terms are analogous, and are not explicitly derived.

In what follows, we drop constants with respect to the parameters for simplicity.

\subsubsection{Conditional Density for the Rating}
\label{appendix:varLowerBound:completeLL:rating}

For the conditional density of the rating,

\begin{equation}
\begin{aligned}
\label{eq:varLowerBound:completeLL:rating}
& \E_{Q}\left[I_{i,j} \log P(r_{i,j} | r_{-(i,j)}, U_i, V_j, W_{1:m}, \alpha_i, \beta_j, \tau\right] \\
  =& \frac{I_{i,j}}{2}\E_{Q}\left[\log\alpha_i + \log\beta_j + \log\tau - \tau\alpha_i\beta_j(r_{i,j} - \hat{r}_{i,j})^2\right] \\
  =& \frac{I_{i,j}}{2}\bigg(\E_{Q}[\log\alpha_i] + \E_{Q}[\log\beta_j] + \E_{Q}[\log\tau] \\
   & - \E_{Q}[\tau]\E_{Q}[\alpha_i]\E_{Q}[\beta_j]\left[\E_{Q}[(r_{i,j} - \hat{r}_{i,j})^2\right]\bigg) \\
\end{aligned}
\end{equation}

Expanding the quadratic 

\begin{equation}
\begin{aligned}
\E_{Q}[(r_{i,j} - \hat{r}_{i,j})^2] 
  =& \E_{Q}[r_{i,j}^2 - 2r_{i,j}\hat{r}_{i,j} + \hat{r}_{i,j}^2] \\
  =& r_{i,j}^2 - 2r_{i,j}\E_{Q}[\hat{r}_{i,j}] + \E_{Q}[\hat{r}_{i,j}^2]
\end{aligned}
\end{equation}

Linearity of expectation and the independence assumption from the variational approximation gives a simple result for the second term,

\begin{equation}
\begin{aligned}
\label{eq:varLowerBound:completeLL:rating:Quad:firstOrder}
-2r_{i,j}\E_{Q}[\hat{r}_{i,j}] 
  =& -2r_{i,j}\E_{Q}\left[\left(\delta_{U} U_i + \delta_{W}\frac{\sum_{k=1}^{M}I_{i,k}W_k}{n_i}\right)^\top V_j\right] \\
  =& -2r_{i,j}\left(\delta_{U}\E_{Q}[U_i^\top V_j] + \frac{\delta_{W}}{n_i} \sum_{k=1}^{M}I_{i,k}\E_{Q}\left[W_k^\top V_j\right]\right) \\
  =& -2r_{i,j}\left(\delta_{U} \E_{Q}[U_i]^\top \E_{Q}[V_j] + \frac{\delta_{W}}{n_i} \sum_{k=1}^{M}I_{i,k} \E_{Q}[W_k]^\top\E_{Q}[V_j]\right) \\
  =& -2r_{i,j}\left(\delta_{U} \mu_{U_i}^\top \mu_{V_j} + \frac{\delta_{W}}{n_i} \sum_{k=1}^{M}I_{i,k} \mu_{W_k}^\top \mu_{V_j}\right)
\end{aligned}
\end{equation}

For the second moment, expanding the square leads to three additional terms

\begin{equation}
\begin{aligned}
\label{eq:varLowerBound:compelteLL:rating:Quad:square}
\E_{Q}[\hat{r}_{i,j}^2] 
  =& \delta_{U}\E_{Q}\left[V_j^\top U_iU_i^\top V_j\right] \\
  & + 2\frac{\delta_{U}\delta_{W}}{n_i}\sum_{k=1}^{M}I_{i,k}\E_{Q}[V_j^\top U_i W_k^\top V_j] \\
  & + \left(\frac{\delta_{W}}{n_i}\right)^2\E_{Q}\left[\left(V_j^\top \sum_{k=1}^{M}I_{i,k}W_k\right)\left(\sum_{\ell=1}^{M}I_{i,\ell}W_{\ell}^\top V_j\right)\right]
\end{aligned}
\end{equation}

For the first involving only user and items,

\begin{equation}
\begin{aligned}
\label{eq:varLowerBound:completeLL:rating:Quad:square:userItem}
& \delta_{U}\E_{Q}\left[V_j^\top U_iU_i^\top V_j\right] \\
  =& \delta_{U} \tr{\E_{Q}\left[V_j V_j^\top [U_iU_i^\top ]\right]} \\
  =& \delta_{U} \tr{\E_{Q}[V_jV_j^\top]\E_{Q}[U_iU_i^\top]} \\
  =& \delta_{U} \tr{(\Var_{Q}[V_j] + \E_{Q}[V_j]\E_{Q}[V_j]^\top)(\Var_{Q}[U_i] + \E_{Q}[U_i]\E_{Q}[U_i]^\top)} \\
  =& \delta_{U} \tr{(\Lambda^{-1}_{V_j} + \mu_{V_j}\mu_{V_j}^\top)(\Lambda^{-1}_{U_i} + \mu_{U_i}\mu_{U_i}^\top)} \\
\end{aligned}
\end{equation}

For the term involving user, item and side features,

\begin{equation}
\begin{aligned}
\label{eq:varLowerBound:completeLL:rating:Quad:square:userItemSide}
& 2\frac{\delta_{U}\delta_{W}}{n_i}\sum_{k=1}^{M}I_{i,k}\E_{Q}[V_j^\top U_i W_k^\top V_j] \\
  =& 2\frac{\delta_{U}\delta_{W}}{n_i}\sum_{k=1}^{M}I_{i,k}\tr{ \E_{Q}[V_j V_j^\top U_i W_k^\top] } \\
  =& 2\frac{\delta_{U}\delta_{W}}{n_i}\sum_{k=1}^{M}I_{i,k}\tr{ \E_{Q}[V_j V_j^\top] \E_{Q}[U_i] \E_{Q}[W_k]^\top] } \\
  =& 2\frac{\delta_{U}\delta_{W}}{n_i}\sum_{k=1}^{M}I_{i,k}\tr{ (\Lambda^{-1}_{V_j} + \mu_{V_j}\mu_{V_j}^\top) \mu_{U_i} \mu_{W_k}^\top } \\
\end{aligned}
\end{equation}

For the final term involving only item and side features,

\begin{equation}
\begin{aligned}
\label{eq:varLowerBound:completeLL:rating:Quad:square:itemSide}
& \left(\frac{\delta_{W}}{n_i}\right)^2\E_{Q}[(V_j^\top \sum_{k=1}^{M}I_{i,k}W_k)(\sum_{\ell=1}^{M}I_{i,\ell}W_{\ell}^\top V_j)] \\
  =& \left(\frac{\delta_{W}}{n_i}\right)^2\E_{Q}[(V_j^\top( \sum_{k=1}^{M}I_{i,k}W_kW_k^\top + \sum_{k\neq\ell} I_{i,k}I_{i,\ell}W_k W_\ell^\top ) V_j)] \\
  =& \left(\frac{\delta_{W}}{n_i}\right)^2\bigg(\sum_{k=1}^{M}I_{i,k} \E_{Q}[V_j^\top W_kW_k^\top V_j] + \sum_{k\neq\ell} I_{i,k}I_{i,\ell} \E_{Q}[V_j^\top W_k W_\ell^\top V_j] \bigg)\\
  =& \left(\frac{\delta_{W}}{n_i}\right)^2\bigg(\sum_{k=1}^{M}I_{i,k} \tr{\E_{Q}[V_jV_j^\top W_kW_k^\top]} + \sum_{k\neq\ell} I_{i,k}I_{i,\ell} \tr{\E_{Q}[V_jV_j^\top W_k W_\ell^\top]} \bigg)\\
  =& \left(\frac{\delta_{W}}{n_i}\right)^2\bigg(\sum_{k=1}^{M}I_{i,k} \tr{\E_{Q}[V_jV_j^\top] \E_{Q}[W_kW_k^\top]} \\
  & + \sum_{k\neq\ell} I_{i,k}I_{i,\ell} \tr{\E_{Q}[V_jV_j^\top] \E_{Q}[W_k]\E_{Q}[W_\ell^\top]} \bigg)\\
  =& \left(\frac{\delta_{W}}{n_i}\right)^2\bigg(\sum_{k=1}^{M}I_{i,k} \tr{(\Lambda^{-1}_{V_j} + \mu_{V_j}\mu_{V_j}^\top) (\Lambda^{-1}_{W_k} + \mu_{W_k}\mu_{W_k}^\top)} \\
  & + \sum_{k\neq\ell} I_{i,k}I_{i,\ell} \tr{(\Lambda^{-1}_{V_j} + \mu_{V_j}\mu_{V_j}^\top) \mu_{W_k}\mu_{W_\ell}^\top} \bigg)\\
\end{aligned}
\end{equation}

Combining Equations \eqref{eq:varLowerBound:completeLL:rating:Quad:square:itemSide}, \eqref{eq:varLowerBound:completeLL:rating:Quad:square:userItem}, \eqref{eq:varLowerBound:completeLL:rating:Quad:square:userItemSide} with the first order term in Equation \eqref{eq:varLowerBound:completeLL:rating:Quad:firstOrder} yields

\begin{equation}
\begin{aligned}
& \E_{Q}[(r_{i,j} - \hat{r}_{i,j})^2] \\
  =& r_{i,j}^2 \\
  & - 2r_{i,j}\left(\delta_{U} \mu_{U_i}^\top \mu_{V_j} + \frac{\delta_{W}}{n_i} \sum_{k=1}^{M}I_{i,k} \mu_{W_k}^\top \mu_{V_j}\right) \\
  & + \delta_{U} \tr{(\Lambda^{-1}_{V_j} + \mu_{V_j}\mu_{V_j}^\top)(\Lambda^{-1}_{U_i} + \mu_{U_i}\mu_{U_i}^\top)} \\
  & + 2\frac{\delta_{U}\delta_{W}}{n_i}\sum_{k=1}^{M}I_{i,k}\tr{ (\Lambda^{-1}_{V_j} + \mu_{V_j}\mu_{V_j}^\top) \mu_{U_i} \mu_{W_k}^\top } \\
  & + \left(\frac{\delta_{W}}{n_i}\right)^2\bigg[\sum_{k=1}^{M}I_{i,k} \tr{(\Lambda^{-1}_{V_j} + \mu_{V_j}\mu_{V_j}^\top) (\Lambda^{-1}_{W_k} + \mu_{W_k}\mu_{W_k}^\top)} \\
  & + \sum_{k\neq\ell} I_{i,k}I_{i,\ell} \tr{(\Lambda^{-1}_{V_j} + \mu_{V_j}\mu_{V_j}^\top) \mu_{W_k}\mu_{W_\ell}^\top} \bigg]\\
\end{aligned}
\end{equation}

Combining with the precision factors yields Equation \eqref{eq:varLowerBound:completeLL:rating}.

\begin{equation}
\begin{aligned}
\label{eq:varLowerBound:completeLL:rating:exression}
& \E_{Q}\left[I_{i,j} \log p(r_{i,j} | r_{-(i,j)}, U_i, V_j, W_{1:m}, \alpha_i, \beta_j, \tau)\right] \\
  =& \frac{1}{2}\sum_{i=1}^{N}\sum_{j=1}^{M} I_{i,j}\bigg(\left[-\log \tilde{b}_{U_i} + \psi(\tilde{a}_{U_i}) -\log \tilde{b}_{V_j} + \psi(\tilde{a}_{V_j}) -\log \tilde{b}_\tau + \psi(\tilde{a}_\tau)\right] \\
  & - \frac{\tilde{a}_\tau}{\tilde{b}_\tau}\frac{\tilde{a}_{U_i}}{\tilde{b}_{U_i}}\frac{\tilde{a}_{V_j}}{\tilde{b}_{V_j}}\bigg[r_{i,j}^2 
   - 2r_{i,j}\left(\delta_{U} \mu_{U_i}^\top \mu_{V_j} + \frac{\delta_{W}}{n_i} \sum_{k=1}^{M}I_{i,k} \mu_{W_k}^\top \mu_{V_j}\right) \\
  & + \delta_{U} \tr{(\Lambda^{-1}_{V_j} + \mu_{V_j}\mu_{V_j}^\top)(\Lambda^{-1}_{U_i} + \mu_{U_i}\mu_{U_i}^\top)} \\
  & + 2\frac{\delta_{U}\delta_{W}}{n_i}\sum_{k=1}^{M}I_{i,k}\tr{ (\Lambda^{-1}_{V_j} + \mu_{V_j}\mu_{V_j}^\top) \mu_{U_i} \mu_{W_k}^\top } \\
  & + \left(\frac{\delta_{W}}{n_i}\right)^2\bigg(\sum_{k=1}^{M}I_{i,k} \tr{(\Lambda^{-1}_{V_j} + \mu_{V_j}\mu_{V_j}^\top) (\Lambda^{-1}_{W_k} + \mu_{W_k}\mu_{W_k}^\top)} \\
  & + \sum_{k\neq\ell} I_{i,k}I_{i,\ell} \tr{(\Lambda^{-1}_{V_j} + \mu_{V_j}\mu_{V_j}^\top) \mu_{W_k}\mu_{W_\ell}^\top} \bigg)\bigg]\bigg)\\
\end{aligned}
\end{equation}

\subsubsection{User Latent Features}
\label{appendix:varLowerBound:completeLL:features}

For the conditional density of the user latent features

\begin{equation}
\begin{aligned}
& \E_{Q}\left[\log p(U_i | \mu_U, \Lambda_U)\right]\\
  =& \frac{1}{2}\E_{Q}[\log|\Lambda_U|] - \frac{1}{2}\E_{Q}[(U_i - \mu_U)^\top\Lambda_U(U_i - \mu_U)]
\end{aligned}
\end{equation}

For the quadratic form, we use conditional expectation as $(\mu_U, \Lambda_U)$ is jointly a Normal-Wishart under the variational approximation, hence not independent.

\begin{equation}
\begin{aligned}
\label{eq:varLowerBound:completeLL:userFeatures:Quad}
   & \E_{Q}[(U_i - \mu_U)^\top\Lambda_U(U_i - \mu_U)] \\
  =& \E_{Q}[U_i^\top\Lambda_U U_i - 2U_i^\top\Lambda_U\mu_U^\top + \mu_U^\top\Lambda_U\mu_U] \\
  =& \E_{Q}[U_i^\top\Lambda_U U_i] -2\E_{Q}[U_i]^\top\E_{Q}[\Lambda_U\mu_U^\top] + \E_{Q}[\mu_U^\top\Lambda_U\mu_U] \\
\end{aligned}
\end{equation}

Using the trace on the first term yields,

\begin{equation}
\begin{aligned}
\E_{Q}[U_i^\top\Lambda_U U_i] 
  =& \tr{\E_{Q}[\Lambda_U U_iU_i^\top]} \\
  =& \tr{\E_{Q}[\E_{Q}[\Lambda_U U_i U_i^\top|\Lambda_U]]} \\
  =& \tr{\E_{Q}[\Lambda_U \E_{Q}[U_i U_i^\top|\Lambda_U]]} \\
  =& \tr{\E_{Q}[\Lambda_U]\left(\Var_{Q}[U_i|\Lambda_U] + \E_{Q}[U_i]\E_{Q}[U_i]^\top\right)} \\
  =& \tr{\tilde{\nu}_U\tilde{W}_U\left(\tilde{\Lambda}_{U_i}^{-1} + \mu_{U_i}\mu_{U_i}^\top\right)}
\end{aligned}
\end{equation}

Iterated expectation on the second gives

\begin{equation}
\begin{aligned}
\E_{Q}[U_i]^\top\E_{Q}[\Lambda_U\mu_U^\top]
  =& \E_{Q}[U_i]^\top\E_{Q}[\E_{Q}[\Lambda_U\mu_U^\top|\Lambda_U]] \\
  =& \E_{Q}[U_i]^\top\E_{Q}[\Lambda_U]\E_{Q}[\mu_U|\Lambda_U]^\top \\
  =& \mu_{U_i}^\top \tilde{\nu}_0\tilde{W}_U \tilde{\mu}_U
\end{aligned}
\end{equation}

While both techniques applied to the third yields,

\begin{equation}
\begin{aligned}
\E_{Q}[\mu_U^\top\Lambda_U\mu_U] 
  =& \tr{\E_{Q}[\Lambda_U\mu_U\mu_U^\top]} \\
  =& \tr{\E_{Q}[\E_{Q}[\Lambda_U\mu_U\mu_U^\top|\Lambda_U]]} \\
  =& \tr{\E_{Q}[\Lambda_U]\E_{Q}[\mu_U\mu_U^\top|\Lambda_U]} \\
  =& \tr{\E_{Q}[\Lambda_U]\left(\Var_{Q}[\mu_U|\Lambda_U] + \E_{Q}[\mu_U|\Lambda_U]\E_{Q}[\mu_U|\Lambda_U]^\top\right)} \\
  =& \tr{\tilde{\nu}_U\tilde{W}_U\left(\tilde{\Lambda}_U + \tilde{\mu}_U\tilde{\mu}_U^\top\right)}
\end{aligned}
\end{equation}

Which simplifies to

\begin{equation}
\begin{aligned}
   & \E_{Q}[(U_i - \mu_U)^\top\Lambda_U(U_i - \mu_U)] \\
  =& \tilde{\nu}_U\left[(\mu_{U_i} - \tilde{\mu}_U)^\top\tilde{W}_U(\mu_{U_i} - \tilde{\mu}_U)+ \tilde{\beta_0}^{-1}\tr{\tilde{W}_U\tilde{\Lambda}_{U_i}^{-1}}\right]
\end{aligned}
\end{equation}

The log-precision expectation gives

\begin{equation}
\begin{aligned}
\label{eq:varLowerBound:completeLL:userFeatures:logPrec}
\E_{Q}[\log|\Lambda_U|] =& \sum_{i=1}^{d}\psi\left(\frac{\tilde{\nu}_U + 1 - i}{2}\right) + d\log2 + \log|\tilde{W}_U|
\end{aligned}
\end{equation}

Combining Equation \eqref{eq:varLowerBound:completeLL:userFeatures:Quad} - \eqref{eq:varLowerBound:completeLL:userFeatures:logPrec} and dividing by two gives the contribution to the variational lower bound from the user features,

\begin{equation}
\begin{aligned}
\label{eq:varLowerBound:completeLL:userFeatures:expression}
& \E_{Q}\left[\log p(U_i | \mu_U, \Lambda_U)\right]\\ 
  =& \frac{1}{2}\bigg[\sum_{i=1}^{d}\psi\left(\frac{\tilde{\nu}_U + 1 - i}{2}\right) + d\log2 + \log|\tilde{W}_U| \\
  & - \tilde{\nu}_U\left[(\mu_{U_i} - \tilde{\mu}_U)^\top\tilde{W}_U(\mu_{U_i} - \tilde{\mu}_U)+ \tilde{\beta_0}^{-1}\tr{\tilde{W}_U\tilde{\Lambda}_{U_i}^{-1}}\right]
\end{aligned}
\end{equation}

\subsubsection{User Precision}
\label{appendix:varLowerBound:completeLL:precision}

For the conditional density of the user precision

\begin{equation}
\begin{aligned}
\label{eq:varLowerBound:completeLL:userPrec:expression}
& \E_{Q}\left[\log p(\alpha_i | a_U, b_U) \right] \\  
  =& \E_{Q}\left[a_U\log b_U - \log \Gamma(a_U) + (a_U-1)\log\alpha_i - b_U\alpha_i\right] \\
  =& C + (a_U - 1) \E_{Q}\left[\log\alpha_i\right] - b_U\E\left[\alpha_i\right] \\
  =& C + (a_U - 1) (-\log \tilde{b}_{U_i} + \psi(\tilde{a}_{U_i})) - b_U \frac{\tilde{a}_{U_i}}{\tilde{b}_{U_i}} \\
  =& C + (a_U - 1) (-\log \tilde{b}_{U_i} + \psi(\tilde{a}_{U_i})) - b_U \frac{\tilde{a}_{U_i}}{\tilde{b}_{U_i}}\\
\end{aligned}
\end{equation}
Where $\psi(\cdot)$ is the Digamma function, $\psi(\cdot) = \frac{d}{d\cdot}\log\Gamma(\cdot)$.

\subsubsection{User Bias}
\label{appendix:varLowerBound:completeLL:userBias}

For the user bias $\gamma_i$, the contribution to the variational lower bound is

\begin{equation}
\begin{aligned}
\label{eq:varLowerBound:completeLL:userBias:expression}
   & \E_{Q}[\log p(\gamma_i)] \\
  =& \frac{1}{2}\E_{Q}[\log \lambda_{\gamma}] - \frac{\lambda_{\gamma}}{2}\E_{Q}[(\gamma_i - \mu_{\gamma})^2] \\
  =& \frac{1}{2}\log \lambda_{\gamma} - \frac{\lambda_{\gamma}}{2}\left[\Var_{Q}[\gamma_i] + \left(\E_{Q}[\gamma_i] - \mu_\gamma\right)^2\right] \\
  =& \frac{1}{2}\log \lambda_{\gamma} - \frac{\lambda_{\gamma}}{2}\left[\lambda_{\gamma_i} + \left(\mu_{\gamma_i} - \mu_\gamma\right)^2\right] \\
\end{aligned}
\end{equation}

The item bias contributions are analogous.

\subsubsection{User Hyper-parameters}
\label{appendix:varLowerBound:completeLL:hyperparameters}

For the conditional density of the user hyper-parameters $(\mu_U, \Lambda_U)$, the contribution to the variational lower bound is

\begin{equation}
\begin{aligned}
\label{eq:varLowerBound:completeLL:userHyper}
   & \E_{Q}[\log p(\mu_U, \Lambda_U)] \\
  =& \E_{Q}[\log p(\mu_U | \mu_0, \beta_0\Lambda_U)] + \E_{Q}[\log p(\Lambda_U | \nu_0, W_0)]
\end{aligned}
\end{equation}

The first term contains a factor of $\log|\Lambda_U|$, derived in Equation \eqref{eq:varLowerBound:completeLL:userFeatures:logPrec}, and the quadratic with respect to $\mu_U$.

For the quadratic term, we rearrange under the trace to obtain

\begin{equation}
\begin{aligned}
\label{eq:varLowerBound:completeLL:userHyper:Quad}
& \E_{Q}[(\mu_U - \mu_0)^\top\beta_0\Lambda_U(\mu_U - \mu_0)] \\
  =& \beta_0\E_{Q}[\tr{ \Lambda_U(\mu_U - \mu_0)(\mu_U - \mu_0)^\top }] \\
  =& \beta_0\tr{ \E_{Q}[\Lambda_U]\E_{Q}[(\mu_U - \mu_0)(\mu_U - \mu_0)^\top] } \\
  =& \beta_0\tr{ \E_{Q}[\Lambda_U] \cdot \{\Var_{Q}[\mu_U - \mu_0] + \E_{Q}[\mu_U - \mu_0]\E_{Q}[\mu_U - \mu_0]^\top\} } \\
  =& \beta_0\tr{ \E_{Q}[\Lambda_U] \cdot \{\Var_{Q}[\mu_U] + (\E_{Q}[\mu_U] - \mu_0)(\E_{Q}[\mu_U] - \mu_0)^\top\} } \\
  =& \beta_0\tr{ \tilde{\nu}_U\tilde{W}_U \cdot \{ \tilde{\Lambda}^{-1}_{U} + (\tilde{\mu}_U -\mu_0)(\tilde{\mu}_U -\mu_0)^\top\} } \\
  =& \tilde{\nu}_U\beta_0 \left(\tr{ \tilde{W}_U \tilde{\Lambda}^{-1}_{U} } + (\tilde{\mu}_U -\mu_0)^\top \tilde{W}_U(\tilde{\mu}_U -\mu_0)\right)
\end{aligned}
\end{equation}

Subtracting Equation \eqref{eq:varLowerBound:completeLL:userHyper:Quad} from Equation \eqref{eq:varLowerBound:completeLL:userFeatures:logPrec} and dividing by two gives the contribution to the lower bound from the conditional distribution for the user latent feature mean, the first term in Equation \eqref{eq:varLowerBound:completeLL:userHyper}.

\begin{equation}
\begin{aligned}
\label{eq:varLowerBound:completeLL:userHyper:userMean}
& \E_{Q}\left[ \log p(\mu_U | \mu_0, \beta_0\Lambda_U ) \right] \\
  =& \frac{1}{2}\bigg[\sum_{i=1}^{d}\psi\left(\frac{\tilde{\nu}_U + 1 - i}{2}\right) + d\log2 + \log|\tilde{W}_U| \\
  & - \tilde{\nu}_U\beta_0 \left(\tr{ \tilde{W}_U \tilde{\Lambda}^{-1}_{U} } + (\tilde{\mu}_U -\mu_0)^\top \tilde{W}_U(\tilde{\mu}_U -\mu_0)\right) \bigg]
\end{aligned}
\end{equation}

For the second term in Equation \eqref{eq:varLowerBound:completeLL:userHyper}, the Wishart on the user precision matrix, we have

\begin{equation}
\begin{aligned}
\label{eq:varLowerBound:completeLL:userHyper:userPrecision}
& \E_{Q}\left[\log p(\Lambda_U | W_0, \nu_0)\right] \\
  =& \frac{\nu_0 - d - 1}{2}\E_{Q}\left[\log |\Lambda_U|\right] - \frac{1}{2}\left(\tr{\E_{Q}[W_0^{-1}\Lambda_U]}\right) \\
  =& \frac{\nu_0 - d - 1}{2}\E_{Q}\left[\log |\Lambda_U|\right] - \frac{1}{2}\mathrm{tr}\left(W_0^{-1}\E_{Q}[\Lambda_U]\right) \\
  =& \frac{\nu_0 - d - 1}{2}\left[\sum_{i=1}^{d}\psi\left(\frac{\tilde{\nu}_U + 1 - i}{2}\right) + p\log2 + \log|\tilde{W}_U|\right] \\
  & - \frac{\tilde{\nu}_{U}}{2}\tr{W_0^{-1}\tilde{W}_U}
\end{aligned}
\end{equation}

Combining Equations \eqref{eq:varLowerBound:completeLL:userHyper:userMean} and \eqref{eq:varLowerBound:completeLL:userHyper:userPrecision} yield the contribution of interest, Equation \eqref{eq:varLowerBound:completeLL:userHyper}, as

\begin{equation}
\begin{aligned}
\label{eq:varLowerBound:completeLL:userHyper:expression}
   & \E_{Q}[\log(p(\mu_U, \Lambda_U)] \\
  =& \E_{Q}[\log p(\mu_U | \mu_0, \beta_0\Lambda_U)] + \E_{Q}[\log p(\Lambda_U | \nu_0, W_0)] \\
  =& \frac{1}{2}\bigg[\sum_{i=1}^{d}\psi\left(\frac{\tilde{\nu}_U + 1 - i}{2}\right) + d\log2 + \log|\tilde{W}_U| \\
  & - \tilde{\nu}_U\beta_0 \left(\tr{ \tilde{W}_U \tilde{\Lambda}^{-1}_{U} } + (\tilde{\mu}_U -\mu_0)^\top \tilde{W}_U(\tilde{\mu}_U -\mu_0)\right) \bigg] \\
  & \frac{\nu_0 - d - 1}{2}\left[\sum_{i=1}^{d}\psi\left(\frac{\tilde{\nu}_U + 1 - i}{2}\right) + p\log2 + \log|\tilde{W}_U|\right] \\
  & - \frac{\tilde{\nu}_{U}}{2}\tr{W_0^{-1}\tilde{W}_U}
\end{aligned}
\end{equation}

\subsection{Entropy}
\label{appendix:varLowerBound:entropy}

From Equation \eqref{eq:meanFieldApprox}, the entropy term takes the form

\begin{equation}
\begin{aligned}
\label{eq:varLowerBound:entropy}
H[Q] =& -\E_{Q}[\log Q(\tau, \alpha_{1:N}, \beta_{1:M}, U_{1:N}, V_{1:M}, W_{1:M}, \mu_U, \Lambda_U, \mu_V, \lambda_V, \mu_W, \Lambda_W) ] \\
  =& -\E_{Q}\left[\log Q(\tau)\right] \\
    & - \sum_{i=1}^{N} \E_{Q}\left[\log Q(U_i)\right] - \sum_{i=1}^{N} \E_{Q}\left[Q(\alpha_i)\right] \\
    & - \sum_{j=1}^{M} \E_{Q}\left[ \log Q(V_j)\right] - \sum_{i=1}^{M}\E_{Q}\left[\log Q(\beta_j) \right] \\
    & - \sum_{k=1}^{M} \E_{Q}\left[\log Q(W_k) \right]\\
    & -\E_{Q}\left[ \log Q(\mu_U, \Lambda_U)\right]
     - \E_{Q}\left[ \log Q(\mu_V, \Lambda_V)\right]
     - \E_{Q}\left[ \log Q(\mu_W, \Lambda_W)\right]
\end{aligned}
\end{equation}

As we did for the expected complete log likelihood, we derive each factor separately.

\begin{itemize}
\item In Section \ref{appendix:varLowerBound:entropy:features}, we derive the entropy of the user feature vector, $\E_{Q}\left[\log Q(U_i)\right]$.  The final expression for this factor is Equation \eqref{eq:varLowerBound:entropy:features:expression};
\item In Section \ref{appendix:varLowerBound:entropy:globalPrec}, we derive the entropy of the global precision, $\E_{Q}\left[\log Q(\tau)\right]$.  The final expression for this factor is Equation \eqref{eq:varLowerBound:entropy:globalPrec:expression};
\item In Section \ref{appendix:varLowerBound:entropy:userBias}, we derive the entropy of the user bias, $\E_{Q}\left[\log Q(\gamma_i)\right]$.  The final expression for this factor is Equation \eqref{eq:varLowerBound:entropy:userBias:expression};
\item In Section \ref{appendix:varLowerBound:entropy:hyperparameters}, we derive the entropy of the user feature hyper-parameters, $\E_{Q}\left[\log Q(\mu_U, \Lambda_U)\right]$  The final expression is for this factor is Equation \eqref{eq:varLowerBound:entropy:hyperparameters:expression}.
\end{itemize}

\subsubsection{Feature Vectors}
\label{appendix:varLowerBound:entropy:features}

We derive the contribution from a single user feature.

\begin{equation}
\begin{aligned}
\label{eq:varLowerBound:entropy:features:expression}
   & \E_{Q}[\log Q(U_i)] \\
  =& \frac{1}{2}\E_{Q}[\log|\tilde{\Lambda}_{U_i}] - \frac{1}{2}\E_{Q}[(U_i - \mu_{U_i})^\top\tilde{\Lambda}_{U_i}(U_i - \mu_{U_i})] \\
  =& \frac{1}{2}\log|\tilde{\Lambda}_{U_i}|
\end{aligned}
\end{equation}

The first is parameter, hence constant, while the second term is zero as it is an expectation of a quadratic form centered by the mean and scaled by the precision, see Section \ref{sec:expQuadForm}.

\subsubsection{Precision Terms}
\label{appendix:varLowerBound:entropy:globalPrec}

We derive the contribution from the global precision factor.  The other precisions follow analogously.

\begin{equation}
\begin{aligned}
\label{eq:varLowerBound:entropy:globalPrec:expression}
   & \E_{Q}[ \log Q(\tau) ] \\
  =& \tilde{a}_\tau\log\tilde{b}_\tau + \log\Gamma(\tilde{a}_\tau) + (\tilde{a}_\tau -1)\E_{Q}[\log \tau] - \tilde{b}_\tau\E_{E}[\tau] \\
  =& \tilde{a}_\tau\log\tilde{b}_\tau + \log\Gamma(\tilde{a}_\tau) + (\tilde{a}_\tau -1)(-\log\tilde{b}_\tau + \psi(\tilde{a}_\tau)) - \tilde{b}_\tau\frac{\tilde{a}_\tau}{\tilde{b}_\tau} \\
  =& \tilde{a}_\tau\log\tilde{b}_\tau + \log\Gamma(\tilde{a}_\tau) + (\tilde{a}_\tau -1)(-\log\tilde{b}_\tau + \psi(\tilde{a}_\tau)) - \tilde{a}_\tau\\
  =& -\tilde{a}_\tau - \log \tilde{b}_\tau - \log \Gamma(\tilde{a}_\tau) - (\tilde{a}_\tau-1)\psi(\tilde{a}_\tau)
\end{aligned}
\end{equation}

\subsubsection{User Bias}
\label{appendix:varLowerBound:entropy:userBias}

For the user bias $\gamma_i$, the contribution to the entropy is

\begin{equation}
\begin{aligned}
\label{eq:varLowerBound:entropy:userBias:expression}
   & \E_{Q}[\log Q(\gamma_i)] \\
  =& \frac{1}{2}\E_{Q}[\log \lambda_{\gamma_i}] - \frac{\lambda_{\gamma_i}}{2}\E_{Q}[(\gamma_i - \mu_{\gamma_i})^2] \\
  =& \frac{1}{2}\log \lambda_{\gamma_i} - \frac{\lambda_{\gamma_i}}{2}\Var_{Q}[\gamma_i] \\
  =& \frac{1}{2}\log \lambda_{\gamma_i} - \frac{\lambda_{\gamma_i}}{2}\frac{1}{\lambda_{\gamma_i}} \\
  =& \frac{1}{2}\log \lambda_{\gamma_i} - \frac{1}{2}
\end{aligned}
\end{equation}

The item bias contributions are analogous.

\subsubsection{Hyper-parameters}
\label{appendix:varLowerBound:entropy:hyperparameters}

We derive the contribution from the user hyper-parameters $(\mu_U, \Lambda_U)$,

\begin{equation}
\begin{aligned}
   & \E_{Q}\left[\log Q(\mu_U, \Lambda_U)\right] \\
  =& \E_{Q}[\log Q(\mu_U | \Lambda_U) ] + \E_{Q}[\log Q(\Lambda_U)] \\
  =& \E_{Q}\left[\frac{\tilde{\beta}_0}{2}\log|\Lambda_U| - \frac{\tilde{\beta}_0}{2}(\mu_U - \tilde{\mu}_U)^\top \Lambda_U (\mu_U - \tilde{\mu}_U)\right] \\ 
   & + \E_{Q}\left[-\frac{\tilde{\nu}_U}{2}\log|\tilde{W}_U| + \frac{\tilde{\nu}_U - d - 1}{2}\log|\Lambda_U| - \frac{1}{2}\tr{\tilde{W}_U^{-1}\Lambda_U}\right] \\
  =& \E_{Q}\left[\frac{\beta_0}{2}\log|\Lambda_U|\right] - \E_{Q}\left[\frac{\tilde{\beta}_0}{2}(\mu_U - \tilde{\mu}_U)^\top \Lambda_U (\mu_U - \tilde{\mu}_U)\right] \\ 
   & + \E_{Q}\left[-\frac{\tilde{\nu}_U}{2}\log|\tilde{W}_U|\right] + \E_{Q}\left[\frac{\tilde{\nu}_U - d - 1}{2}\log|\Lambda_U|\right] - \E{Q}\left[\frac{1}{2}\tr{\tilde{W}_U^{-1}\Lambda_U}\right]
\end{aligned}
\end{equation}

The second expectation involving the quadratic form is zero, as before, while the third is a constant.  The remaining terms contribute,

\begin{equation}
\begin{aligned}
\label{eq:varLowerBound:entropy:hyperparameters:expression}
   & \E_{Q}\left[\log Q(\mu_U, \Lambda_U)\right] \\
  =& \frac{\tilde{\beta}_0}{2}\left[\sum_{i=1}^{d}\psi\left(\frac{\tilde{\nu}_U + 1 - i}{2} + d\log 2\right) + \log|\tilde{W}_U|\right] - 0 \\
   & - \frac{\tilde{\nu}_U}{2}\log|\tilde{W}_U| + \frac{\tilde{\nu}_U-d-1}{2}\left[\sum_{i=1}^{d}\psi\left(\frac{\tilde{\nu}_U + 1 - i}{2}\right) + d\log 2 + \log|\tilde{W}_U|\right] - \frac{1}{2}\tr{\tilde{W}_U^{-1} \tilde{\nu}_U \tilde{W}_U} \\
  =& \frac{\tilde{\nu}_U-d}{2}\left[\sum_{i=1}^{d}\psi\left(\frac{\tilde{\nu}_U + 1 - i}{2}\right) + d\log 2 \right] - \frac{d}{2}\log|\tilde{W}_U^{-1}| - \frac{\tilde{\nu}_U d}{2}
\end{aligned}
\end{equation}

\end{document}